\documentclass{article}

\usepackage{microtype}
\usepackage{nicefrac}
\usepackage{graphicx}
\usepackage{bbm}
\usepackage{tcolorbox}
\usepackage{amssymb,amsmath,amsthm,enumitem}
\usepackage{subfigure}
\usepackage{booktabs} % for professional tables

\usepackage{float}
\usepackage{bigints}
\usepackage{mathabx}
\usepackage{scalerel}
\usepackage{natbib}
\usepackage{float}

\usepackage{hyperref}

\usepackage[accepted]{icml2021}

\icmltitlerunning{Statistical Guarantees for Fairness Aware Plug-In Algorithms}

\begin{document}

\twocolumn[
\icmltitle{Statistical Guarantees for Fairness Aware Plug-In Algorithms}




\begin{icmlauthorlist}
\icmlauthor{Drona Khurana}{to}
\icmlauthor{Srinivasan Ravichandran}{to}
\icmlauthor{Sparsh Jain}{to}
\icmlauthor{Narayanan Unny Edakunni}{to}
\end{icmlauthorlist}

\icmlaffiliation{to}{AI Labs, American Express, India}

\icmlcorrespondingauthor{Drona Khurana}{drona.khurana@aexp.com}

\icmlkeywords{Machine Learning, ICML}
\vskip 0.3in
]



\printAffiliationsAndNotice{}  % leave blank if no need to mention equal contribution

\begin{abstract}
A plug-in algorithm to estimate Bayes Optimal Classifiers for fairness-aware binary classification has been proposed in \cite{menon2018cost}. However, the statistical efficacy of their approach has not been established. We prove that the plug-in algorithm is statistically consistent. We also derive finite sample guarantees associated with learning the Bayes Optimal Classifiers via the plug-in algorithm. Finally, we propose a protocol that modifies the plug-in approach, so as to simultaneously guarantee fairness and differential privacy with respect to a binary feature deemed sensitive.
\end{abstract}

\section{Introduction and Related Work}
Bayes Optimal Classifiers (BOCs) \cite{devroye1996probabilistic} are of significant importance, since they achieve the least average error possible for any classification task. However, BOCs are generally specified in terms of unknown distributional quantities. Constructing sound estimators for BOCs, provided access to only a finite training sample, is thus of utmost practical relevance. One approach to estimating the BOC is through constructing 'plug-in' estimators. The plug-in principle applied to a broad class of problems, including that of binary classification, is well studied in the statistics literature \cite{audibert2007fast, denis2017confidence, yang1999minimax}. Indeed, the existence of a plug-in classifier that is optimal in the minimax sense is established in \cite{audibert2007fast, yang1999minimax}. 
In their work, \cite{menon2018cost} propose a plug-in algorithm to estimate the BOCs corresponding to fairness-aware learning (FAL) tasks. However, \cite{menon2018cost} do not provide guarantees on the statistical efficacy of their algorithm. In this paper, we plug this gap by {\em \textbf{proving that the plug-in algorithm of} \cite{menon2018cost}, \textbf{is indeed statistically consistent}}. We also \textbf{{\em characterise the sample complexity associated with the task of learning a low regret classifier via the plug-in algorithm}}. Closest to our work is that of \cite{chzhen2019leveraging}, wherein an asymptotic study (for a different fairness aware plug-in classifier) is carried out. The work of \cite{chzhen2019leveraging} however, focuses on settings wherein perfect fairness constraints are imposed. It is well established that due to inherent fairness-accuracy trade-offs, ensuring perfect fairness without considerable loss in accuracy is generally not possible \cite{menon2018cost, zhao2019inherent, chen2018my}. We thus focus on approximate notions of two fairness metrics, Demographic Parity (DPar) and Equality of Opportunity (EO). Further, the approach of \cite{chzhen2019leveraging} requires access to the sensitive variable (denoted $\overline{Y}$ hereon) at test time which is often not permitted. The plug-in approach of \cite{menon2018cost} however does not necessitate test-time access to $\overline{Y}$. Indeed, real-world settings may impose even more stringent requirements on $\overline{Y}$ . For example, we may be required to ensure that our model does not leak information about the sensitive attribute, $\overline{Y}$, corresponding to any individual. In such cases, a possible solution is to protect individuals via Differential Privacy (DP) \cite{dwork2006calibrating}. The literature combining fairness and privacy \cite{jagielski2019differentially, cummings2019compatibility, mozannar2020fair}, is emerging and limited. Such settings motivate us to \textbf{{\em propose an easy to deploy, modified version of the plug-in algorithm, referred to as DP Plug-in}}. The framework ensures that $\overline{Y}$ is protected via DP. Using publicly available data sets, we demonstrate empirically, that the \textbf{{\em DP Plug-in algorithm achieves strong privacy-fairness-accuracy guarantees}}, as it outperforms the private, fair approach of \cite{jagielski2019differentially} across 3 out of 4 experimental sets ups considered. 

\section{Background and Notation}
For brevity, we only introduce the main features from \cite{menon2018cost} pertinent to our study in this section. We present other useful definitions and results from \cite{menon2018cost} in section A of the supplement. Additionally, our focus in the main thesis of this paper will be on the approximate EO criterion for the case when $\overline{Y}$ is unavailable during test-time. Analogous analyses for 1) the case when $\overline{Y}$ is available at test time, and for 2) the approximate DPar criterion are presented in sections B and C of the supplement. 
\newpage

Access to a finite training sample, \begin{small}{$S = \left\{x_{i}, y_{i}, \overline{y}_{i}\right\}_{i=1}^{n}$ }\end{small}drawn i.i.d from some unknown distribution $\mathbbm{P}$ is assumed in \cite{menon2018cost}. $\smash{\forall i \in [n]}$ the triplet $\left(x_{i}, y_{i}, \overline{y}_{i}\right)$ is a realisation of the random variable triplet \begin{small}{$\left(X, Y, \overline{Y}\right)$}\end{small} comprising of the feature, label and sensitive attribute respectively, where $X$ is supported on measurable domain $\chi$ and $Y, \overline{Y} \in \{-1, 1\}$. Let \begin{small}{$\smash{\pi = \mathbbm{P}(Y = 1), \overline{\pi} = \mathbbm{P}(\overline{Y} = 1)}$}\end{small} and \begin{small}{$\smash{\beta = \mathbbm{P}(\overline{Y} = 1 \vert Y = 1)}$}\end{small}. Assume that $\smash{\pi, \overline{\pi}, \beta > 0}$. Let \begin{small}$\smash{(X, Y) \sim \mathcal{D}}$\end{small}, \begin{small}{$ \hspace{0.3em}\smash{\left(X, \overline{Y} \vert Y = 1\right) \sim \overline{\mathcal{D}}_{EO}}.$ Let $\smash{f:\chi \rightarrow\{-1, 1\}}$}\end{small} denote a binary classifier. Regression functions w.r.t. \begin{small}$\mathcal{D}, \mathcal{\overline{D}}_{EO}$\end{small} are given by \begin{small}{$\smash{\eta(x) = \mathbbm{P}(Y = 1 \vert X = x)}$}\end{small}, \begin{small}{$\smash{\overline{\eta}_{EO}(x, y) = \mathbbm{P}(\overline{Y} = 1 \vert X=x, Y=y)}$}\end{small}  respectively.\\\\
A central object of interest in \cite{menon2018cost}, is the notion of cost-sensitive risks (CSR). We denote false positive and negative rates of $f$ w.r.t $\mathcal{D}$, by \begin{small}{$FPR_{\mathcal{D}}(f)$}\end{small} and  \begin{small}{$FNR_{\mathcal{D}}(f)$}\end{small} respectively.

{\bf Definition 2.1} {\hspace{0.1em} \it The cost sensitive risk of a binary classifier $\smash{f: \chi \rightarrow \{-1, 1\}}$, with respect to a distribution $\mathcal{D}$, parameterised by $c\in\left(0, 1\right)$ is given by:\\
\setlength\abovedisplayskip{5.5pt}
\begin{small}
$ CS(f; \mathcal{D}, c) := c\left(1-\pi\right)FPR\left(f;\mathcal{D}\right) + \pi(1-c)FNR\left(f;\mathcal{D}\right)$
\end{small}
}

We now define our fairness metric of interest, i.e., Equality of Opportunity (EO). \\\\
{\bf Definition 2.2} {\hspace{0.1em} \it A binary classifier $\smash{f: \chi \rightarrow \{-1, 1\}}$, admits Equality of Opportunity (EO) if:
$\smash{\mathbbm{P}(f = 1 \vert Y = 1, \overline{Y} = -1) = \mathbbm{P}(f = 1 \vert Y = 1, \overline{Y} = 1)}$
}
\\\\
Thus, EO requires parity in the TPRs between groups as explicated in Definition 2.2. Obtaining perfect fairness while retaining non-trivial accuracy is generally not possible, and so \cite{menon2018cost} introduce approximate measures of fairness which require the additive or multiplicative disparity between prediction rates to be small. A key lemma in \cite{menon2018cost} draws an equivalence between the super-level sets of approximate fairness measures and CSRs. This in turn leads to a reduction of the FAL problem to a problem with constraints on cost-sensitive risks (The reader may refer to section A of the supplement for a more thorough presentation of these results). The FAL problem in \cite{menon2018cost} is thus posed in terms of CSRs as follows: 
\\\\
{\bf Problem 2.3} { (\textbf{Cost-sensitive FAL}) \it For trade-off parameter $\lambda \in \mathbbm{R}$, and cost parameters $c, \overline{c} \in \left(0, 1\right)^{2}$, minimise the fairness-aware cost-sensitive risk:
\vspace*{-1.57mm}
\begin{small}{
\[
R_{FA}(f; \mathcal{D}, \overline{\mathcal{D}}_{EO}, c, \overline{c}, \lambda) = CS(f; \mathcal{D}, c) - \lambda CS(f; \overline{\mathcal{D}}_{EO}, \overline{c})
\]
}\end{small}}
Equipped with this soft constrained FAL problem formulated in terms of CSRs, \cite{menon2018cost} derive the BOCs corresponding to such problems. We present the BOC for approx. EO and the plug-in algorithm of \cite{menon2018cost} to estimate this BOC, in Theorem 2.4 and Algorithm 1 respectively. It is this (optimal) classifier and algorithm that will make for the key objects of our theoretical analysis in Section 3.

{\bf Theorem 2.4} {\hspace{0.1em} (\textbf{BOC for FAL}) \it Pick any costs $c, \overline{c} \in (0, 1)^{2}$ and trade-off parameter $\lambda \in \mathbbm{R}$. Then:
\begin{small}
\begin{equation*}
\begin{split}
Argmin_{f} \hspace{0.3em}R_{FA} (f; \mathcal{D}, \overline{\mathcal{D}}_{EO}) = f^{*} =  sign\circ s^{*} \\
where, \hspace{0.2em} s^{*}\left(x\right) = \left\{ 1 - \frac{\lambda}{\pi}\left(\overline{\eta}_{EO}(x, 1) - \overline{c}\right) \right\}\eta(x) - c
\end{split}
\end{equation*}
\end{small}
}
\vspace{-1.5em}
\begin{algorithm}
   \caption{\small{Plugin approach to FAL, EO setting}}
\begin{algorithmic}

   \STATE \small{{\bfseries Input:} Sample $S$ = $\{x_{i}, y_{i}, \overline{y}_{i}\}_{i=1}^{n}$ from distribution $\mathbbm{P}$; cost parameters $c, \overline{c}$; trade-off parameter $\lambda$}
   \vspace{2.5mm}
   \STATE {\bfseries Estimate:} \small{$\pi$ via \hspace{0.5 em}$\hat{\pi} = \frac{1}{n}\sum_{i=1}^{n}\mathbbm{I}\{y_{i} = 1\}$}
   \STATE {\bfseries Estimate:} \small{$\eta$:$\chi \rightarrow [0, 1]$ using appropriate CPE on $\{x_{i}, y_{i}\}_{i=1}^{n}$}
   \STATE {\bfseries Estimate:} \small{$\overline{\eta}_{EO}$: $(\chi, Y) \rightarrow [0, 1]$ using appropriate CPE on $S$}
   \STATE {\bfseries Compute:} \small{$\hat{s}\left(x\right) =\left\{ 1 - \frac{\lambda}{\hat{\pi}}(\hat{\overline{\eta}}_{EO}(x, 1) - \overline{c}) \right\}\hat{\eta}(x) - c$}
   \vspace{2.5mm}
   \STATE {\bfseries Return:} \small{$\hat{f}\left(x\right) = sign\circ \hat{s}(x)$}
\end{algorithmic}
\end{algorithm}
\section{Theory}

In this section, we analyse the asymptotic and non-asymptotic properties of the plug-in algorithm (Algorithm 1). Recall from Problem 2.3, that our goal is to minimise the fairness aware cost-sensitive risk, which for a given choice of cost parameters $c, \overline{c} \in [0, 1]^{2}$ and trade-off parameter $\lambda \in \mathbbm{R}$ is given by \begin{small}
$\smash{CS(f; \mathcal{D}, c) - \lambda CS(f; \overline{\mathcal{D}}_{EO}, \overline{c})}.$\end{small}

In the language of \cite{narasimhan2014statistical}, we introduce the notion of performance measure. A performance measure, defined w.r.t a distribution $\mathbbm{P}$, and performance metric $\Psi$, is a mapping from the space of measurable functions $\mathcal{F}$ to the reals, i.e., $\mathfrak{P}_{\mathbbm{P}}^{\Psi}: \mathcal{F} \rightarrow \mathbbm{R}$.  In our setting, the performance metric is simply given by the negative of the objective function of Problem 2.3. Unless stated otherwise, we will denote $\overline{\mathcal{D}}_{EO}$ by $\overline{\mathcal{D}}$ and $\overline{\eta}_{EO}$ by $\overline{\eta}$ from hereon. Our performance measure is given by:
\vspace*{-2mm}
\begin{small}
\begin{equation*}
\begin{split}
    \mathfrak{P}_{\mathbbm{P}}^{\Psi}(f) = \Psi[TPR_{\mathcal{D}}(f), TNR_{\mathcal{D}}(f), \pi, TPR_{\overline{\mathcal{D}}}(f), \\ TNR_{\overline{\mathcal{D}}}(f), \beta] = - \left\{CS\left(f; \mathcal{D}, c\right) \hspace{0.1em} - \hspace{0.1em}  \lambda CS\left(f; \overline{\mathcal{D}}, \overline{c}\right)\right\}
\end{split}
\end{equation*}
\end{small}
Thus, a classifier's performance measure explains its merit with regards to the combined, fairness-utility objective, appropriately balanced by cost and trade-off parameters. Performance metric $\Psi$, makes explicit that our performance measure is a function of the classifier's TPRs and TNRs with respect to $\mathcal{D}$ and $\overline{\mathcal{D}}_{EO}$, as well as distributional quantities $\pi$ and $\beta$. The CSRs, and thus the performance measure, are linear in TPRs, TNRs and class probabilities implying the performance measure is continuous in the arguments of $\Psi$. The regret of a classifier $f$, w.r.t performance measure $\mathfrak{P}_{\mathbbm{P}}^{\Psi}$ is defined as: \begin{small}$ regret_{\mathbbm{P}}^{\Psi}(f) = \mathfrak{P}_{\mathbbm{P}}^{\Psi, *}- \mathfrak{P}_{\mathbbm{P}}^{\Psi}(f)$\end{small}
where, $\mathfrak{P}_{\mathbbm{P}}^{\Psi, *} = \mathfrak{P}_{\mathbbm{P}}^{\Psi}(f^{*})$. In our case, $f^{*}$ is the BOC introduced in Theorem 2.4

\subsection{Asymptotic Analysis}
In this sub-section, we prove that the plug-in procedure yields an estimator $\hat{f}$ which is $\Psi$-consistent, implying that $ regret_{\mathbbm{P}}^{\Psi}(\hat{f}) \overset{p}{\to} 0$, where $\overset{p}{\to}$ denotes convergence in probability. We denote the estimators of $\eta, \overline{\eta}$ by $\hat{\eta}$ and $\hat{\overline{\eta}}$ respectively. In order to proceed we make the following assumptions: \\\\
{\bf Assumption 1} \textit{{\hspace{0.1em} \begin{small}$\mathbbm{P}_{X \vert Y=1}\left(\gamma(x) \leq c\right), \mathbbm{P}_{X \vert Y=-1}\left(\gamma(x) \leq c\right), \\ \mathbbm{P}_{X \vert Y=1,  \overline{Y}=1}\left(\gamma(x) \leq c\right)$ and $\mathbbm{P}_{X \vert Y=1, \overline{Y}=-1}\left(\gamma(x) \leq c\right)$\end{small} are continuous at c, where \begin{small}$\gamma(x) = (1 + \frac{\lambda\overline{c}}{\pi})\eta(x) - \frac{\lambda}{\pi}\overline{\eta}(x, 1)\eta(x)
$\end{small}, i.e, \begin{small}$\gamma(x)$\end{small} is \begin{small}$s^{*}(x)$\end{small} in Theorem 2.4 without the constant term \begin{small}$c$\end{small}}}
\\\\\\
{\bf Assumption 2} \textit{{\hspace{0.1em} Class probability estimators (CPEs) $\hat{\eta}, \hat{\overline{\eta}}$ are $L$-1 consistent, i.e., \begin{small}$
\mathbbm{E}_{X}\left[\vert \eta(x) - \hat{\eta}(x) \vert\right]  \overset{p}{\to} 0;\quad
\mathbbm{E}_{X, Y}\left[\vert \overline{\eta}(x, y) - \hat{\overline{\eta}}(x, y) \vert\right]  \overset{p}{\to} 0 $\end{small}
}}
\\\\
{\it Remark: As noted in \cite{narasimhan2014statistical, chzhen2019leveraging}, Assumption 2 is not a very strong one, as an appropriately regularized ERM yields an $L$-1 consistent class probability estimator for proper losses \cite{menon2013statistical, agarwal2013surrogate}. }
\\\\
{\bf Assumption 3} \textit{{\hspace{0.1em} Domain $\chi$ is compact and there exist constants \begin{small}$a, B \in \mathbbm{R}_{+}$\end{small}, such that the PDFs, \begin{small}$f_{X \vert Y = -1}, f_{X}, f_{X \vert Y = 1}$\end{small} satisfy \begin{small}$\forall x \in \chi, \hspace{0.25em}0 < a \leq f_{X \vert Y = -1}(x),f_{X}(x), f_{X \vert Y = 1}(x) \leq B$\end{small}
}}
\\\\
{\it Remark: We make this assumption for technical convenience. This is akin to the 'strong density assumption' defined in \cite{audibert2007fast}. This assumption is not necessary for the case when $\overline{Y}$ is available at test time, or for either case relating to the approximate Demographic Parity criterion}
\\\\
We now state our key lemma that facilitates the consistency result. Denoting the estimator derived via the plug-in procedure for $\gamma(x)$ by $\hat{\gamma}(x) = (1 + \frac{\lambda\overline{c}}{\hat{\pi}})\hat{\eta}(x) - \frac{\lambda}{\hat{\pi}}\hat{\overline{\eta}}(x, 1)\hat{\eta}(x)$, we have: 
\\\\
{\bf Lemma 3.1} \textit{{\hspace{0.1em} Provided Assumptions 2 and 3 hold, $\hat{\gamma}$ is $L$-1 consistent, i.e.,\hspace{0.25em}\begin{small} $\smash{ \mathbbm{E}_{X}\left[\vert \gamma(x) - \hat{\gamma}(x) \vert\right] \overset{p}{\to} 0}$\end{small}}}
\\\\
The validity of Lemma 3.1 allows us to leverage the proof template of \cite{narasimhan2014statistical} which in turn proves the plug-in algorithm's consistency. \\\\
{\bf Theorem 3.2 \hspace{0.1em} \it Provided Assumptions 1, 2 and 3 hold, the plug-in algorithm is $\Psi$-consistent, i.e., the algorithm yields \begin{small}$\hat{f} = sign \circ \left\{ \hat{\gamma} - c \right\}$,\end{small} s.t., \begin{small}$\mathfrak{P}_{\mathbbm{P}}^{\Psi}(\hat{f}) \overset{p}{\to} \mathfrak{P}_{\mathbbm{P}}^{\Psi, *}$\end{small}, i.e., \begin{small}$regret_{\mathbbm{P}}^{\Psi}(\hat{f}) \overset{p}{\to} 0$\end{small}
}
\\\\
\textit{Proof sketch}: Lemma 3.1 and Assumption 1 allow us to show that the plug-in yields an estimator $\hat{f}$ which is s.t.: 
\begin{small}
\[
TPR_{\mathcal{D}}(\hat{f}) \overset{p}{\to} TPR_{\mathcal{D}}(f^{*}); \hspace{0.25em} TNR_{\mathcal{D}}(\hat{f}) \overset{p}{\to} TNR_{\mathcal{D}}(f^{*})
\]
\[
TPR_{\overline{\mathcal{D}}}(\hat{f}) \overset{p}{\to} TPR_{\overline{\mathcal{D}}}(f^{*}); \hspace{0.25em} TNR_{\overline{\mathcal{D}}}(\hat{f}) \overset{p}{\to} TNR_{\overline{\mathcal{D}}}(f^{*})
\]
\end{small}
The result then follows by the Continuous Mapping Theorem \cite{mann1943stochastic}, since $\Psi$ is continuous in its arguments. Complete proofs and detailed discussion for the results presented in this section can be found in section B of the supplement. 
\subsection{Non-Asymptotic Analysis}
In this section, our objective is to characterise the sample complexity requirements associated with learning a classifier that yields small regret, via the plug-in algorithm of \cite{menon2018cost}. In our problem formulation, the performance measure of a classifier, is a linear function of its true positive and true negative rates. This implies that the performance measure is non-decomposable, since it cannot be expressed as a summation/ expectation over individual instances. This is contrary to the case associated with most standard loss functions that feature in the ML literature, and thus the finite-sample analysis for our performance measure is non-standard. We provide a strategy that allows us to precisely relate the sample complexity of this task to the sample complexity associated with learning the regression functions, $\eta$ and $\overline{\eta}$, as well as other distributional quantities. We defer the detailed derivation of this strategy to section C of the supplement. We assume in this section that, $\pi = \mathbbm{P}(Y = 1)$ is known. While we can remove this assumption and modify our analysis to obtain equivalent results, we found doing so makes the underlying algebra/ geometry much more convoluted, without adding significant insight. Thus, for simplicity, we proceed by assuming $\pi$ is known. 
\\\\
Recall, by Assumption 2, that we are working in a setting wherein the class probability estimators (CPEs), $\hat{\eta}, \hat{\overline{\eta}}$ are $L$-1 consistent. Convergence in the $L$-1 norm implies convergence in probability, so we can meaningfully define the sample complexity associated with learning the regression function $\eta$ via the CPE $\hat{\eta}$:

{\bf Definition 3.4} {\hspace{0.1em} \it The sample complexity of learning $\eta$, is a mapping $m_{\eta}: (0,1)^{3} \rightarrow \mathbbm{N}$, where $m_{\eta}((\epsilon, \delta^{'}), \delta) $ is the minimal (integer) number of training samples required to ensure that, with probability $\geq (1 - \delta)$: \begin{small}$\mathbbm{P}_{X}(\vert \eta(x) - \hat{\eta}(x) \vert \geq \epsilon) \leq \delta^{'}$\end{small}}

Note, we show in Lemma B.1 of the supplement that  $\hat{\overline{\eta}}(\cdot, 1)$ is also $L$-1 consistent. The sample complexity of learning $\overline{\eta}(\cdot, 1)$ is thus analogously defined, and we denote this by $m_{\overline{\eta}}$. Our non-asymptotic result is derived via a geometric argument based in the plane of regression functions, i.e., the $(\overline{\eta}(\cdot, 1), \eta)$-plane. We define some key objects pertaining to our derivation. Consider in the $(\overline{\eta}(\cdot, 1), \eta)$-plane, the hyperbola \begin{small}$H(\lambda, \pi, c, \overline{c}) := \{(1 + \frac{\lambda\overline{c}}{\pi})\eta - \frac{\lambda}{\pi}\overline{\eta}(\cdot, 1)\eta - c = 0$\}\end{small}. Also, for $\epsilon \in (0, \frac{1}{2})$, let \begin{small}\textit{$X_{M} := \{x \in \chi:$ the square of length $2\epsilon$ centred at $(\overline{\eta}(x, 1), \eta(x))$ intersects the hyperbola $H(\lambda, \pi, c, \overline{c})$ in the $(\overline{\eta}(\cdot, 1), \eta)$-plane\}}\end{small}. Having defined \begin{small}$H(\lambda, \pi, c, \overline{c})$\end{small} and \begin{small}$X_{M}$\end{small}, we now state our non-asymptotic result: 

\begin{small}
{\bf Theorem 3.5 \hspace{0.1em} \it Let $\delta, \delta^{'}, \epsilon \in (0, \frac{1}{2})$. Pick any $t > Q = 4G \{max\{c(1-\pi), (1-c)\pi, \vert\lambda\vert\overline{c}(1-\beta), \vert\lambda\vert(1-\overline{c})\beta\}\}$, where $G = max\{\frac{B}{1-\pi}, \frac{B}{\pi \beta}, \frac{B}{\pi (1-\beta)}\}$, and $B = \delta^{'} + \mathbbm{P}_{X}(X_{M})$. Provided access to $n \geq max\{m_{\eta}((\epsilon, \frac{\delta^{'}}{2}), \frac{\delta}{8}), m_{\overline{\eta}}((\epsilon, \frac{\delta^{'}}{2}), \frac{\delta}{8})\}$ training samples drawn i.i.d. from $\mathbbm{P}$, the plug-in algorithm yields an estimator $\hat{f}$, such that, with probability at least $(1-\delta) \hspace{0.3em}: \hspace{0.3em} regret_{\mathbbm{P}}^{\psi}(\hat{f}) \leq t$
}
\end{small}

\textit{Proof sketch:} Our proof entails showing that, for appropriate $t$, \begin{small}
$\mathbbm{P}_{S\sim \mathbbm{P}^{n}}\left[regret_{\mathbbm{P}}^{\Psi} > t\right] {\leq} \delta$ \end{small}
holds, so long as we can meaningfully upper bound \begin{small}$\mathbbm{P}_{X}[f^{*}(x)\neq \hat{f}(x)]$\end{small} with probability \begin{small}$\geq (1-\frac{\delta}{4})$\end{small}. Denoting the upper bound by $B$, we characterise its form via a geometric argument. Roughly, \begin{small}$B \propto \mathbbm{P}_{X}(X_{M})$\end{small}, where \begin{small}$X_{M}$\end{small} is a specific region enclosing the hyperbola, \begin{small}$H(\lambda, \pi, c, \overline{c})$\end{small} in the \begin{small}$(\overline{\eta}(\cdot, 1), \eta)$\end{small}-plane. Then setting $t$ as described, the result follows. A visual simulation of the underlying geometry can be found in \textit{Figure 1}. 
\\\\
Theorem 3.5 tells us that the regret can be made to decrease arbitrarily, provided a sufficient increase in the number of training samples. The precise rate of decay, depends on 1) the sample complexities associated with learning the regression functions and 2) the rate at which the probability measure , i.e., $\mathbbm{P}_{X}$, decays around the hyperbola $H(\lambda, \pi, c, \overline{c})$ (in the $(\overline{\eta}(\cdot, 1), \eta)$-plane) upon shrinking the region of consideration around it (i.e., the region akin to the 'Projected $X_{M}$' region in \textit{Figure 1}). Refer to section C of supplement for a detailed proof and discussion.
\begin{figure}[H]
\includegraphics[width = 8cm]{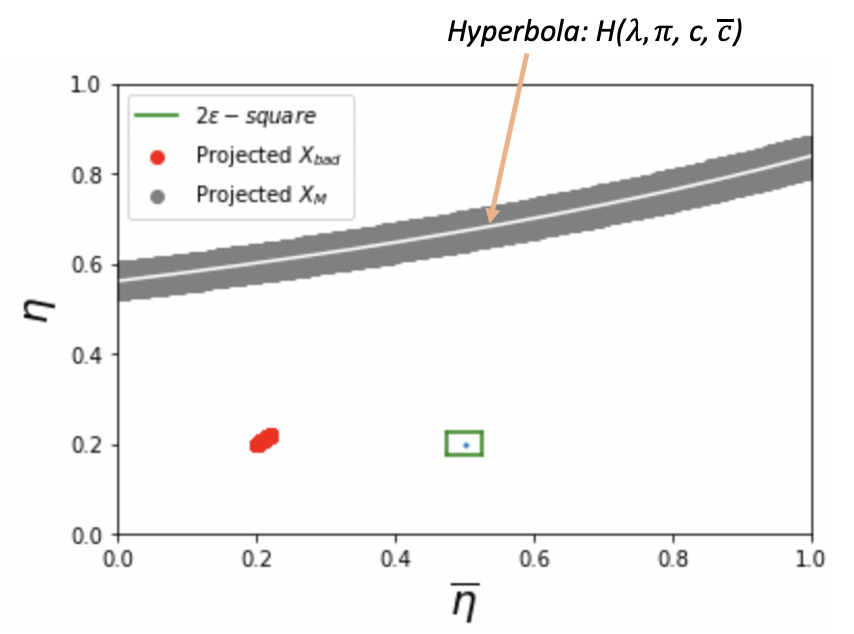}
\begin{tiny}
\caption{\textit{{With probability $\geq (1-\frac{\delta}{4}):$ For any point $x \in \chi:$ \hspace{0.1em} corresponding projected coordinates in the $(\overline{\eta}(\cdot, 1), \eta)$-plane, i.e., $(\overline{\eta}(x, 1), \eta(x))$ lie outside of \{projected $X_{M} \bigcup$ projected $X_{bad}$\}; we can be certain that $f^{*}(x)= \hat{f}(x)$ - the point centred within the green square of length $2\epsilon$} is one such point.}}
\end{tiny}
\end{figure}

\section{Fairness under Differential Privacy}

\begin{figure}[h!]
\includegraphics[width = 8cm]{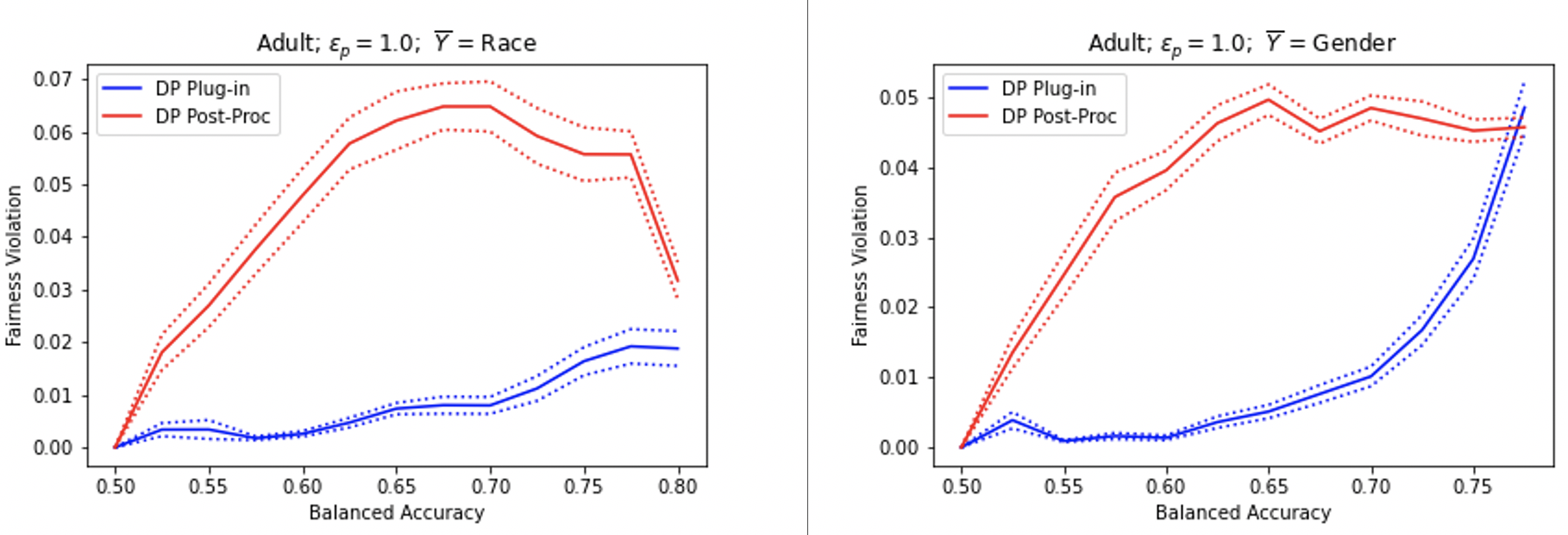}
\begin{tiny}
\caption{\textit{The solid blue curve represents model performance in the 'balanced accuracy-fairness violation plane' corresponding to our method, i.e., the DP Plug-in approach. Whereas the solid red curve, corresponds to model performance for the DP Post-Proc approach. The dotted curves represent the $\pm 0.2$ standard-deviations in fairness violations corresponding to each segment of balanced accuracy considered. Privacy parameter $\epsilon_{p} = 1.0$}}
\end{tiny}
\end{figure}

\begin{algorithm}
   \caption{\small{DP Plugin approach to FAL, EO setting}}
\begin{algorithmic}
   \STATE \small{{\bfseries Input:} Sample $S$ = $\{x_{i}, y_{i}, \overline{y}_{i}\}_{i=1}^{n}$ from distribution $\mathbbm{P}$; cost parameters $c, \overline{c}$; trade-off parameter $\lambda$}; privacy parameter $\epsilon_{p}$
   \vspace{2.5mm}
   \STATE {\bfseries Estimate:} \small{$\pi$ via \hspace{0.5 em}$\hat{\pi} = \frac{1}{n}\sum_{i=1}^{n}\mathbbm{I}\{y_{i} = 1\}$}
   \STATE {\bfseries Estimate:} \small{$\eta$:$\chi \rightarrow [0, 1]$ using appropriate CPE on $\{x_{i}, y_{i}\}_{i=1}^{n}$}
   \STATE {\bfseries Estimate:} \small{$\overline{\eta}_{EO}$: $(\chi, Y) \rightarrow [0, 1]$ using appropriate CPE on $S$}
   \STATE {\bfseries Privatise:} \small{$\hat{\overline{\eta}}_{EO}$ via appropriate privacy preserving protocol yielding, $\epsilon_{p}$-DP protected $\hat{\overline{\eta}}_{EO}^{priv}$}
   \STATE {\bfseries Compute:} {$\hat{s}^{priv}\left(x\right) =$  \begin{scriptsize}$\smash{\left\{ 1 - \frac{\lambda}{\hat{\pi}}(\hat{\overline{\eta}}_{EO}^{priv}(x, 1) - \overline{c}) \right\}\hat{\eta}(x) - c}$\end{scriptsize}}
   \vspace{2.5mm}
   \STATE {\bfseries Return:} \small{$\hat{f}^{priv}\left(x\right) = sign \circ \hat{s}^{priv}\left(x\right)$}
\end{algorithmic}
\end{algorithm}

In this section, we work in a setting wherein there is an additional requirement for our modelling pipeline to mitigate information leakage about the sensitive attribute, $\overline{Y}$. To meet such a requirement we make use of a notion of privacy known as differential privacy \cite{dwork2006calibrating}. We detail the DP Plugin protocol in Algorithm 2. We compare DP Plug-in's performance against that of the private, fair, post-processing approach (DP Post-Proc) of \cite{jagielski2019differentially}. We found that our method outperforms the DP Post-Proc approach in three, out of four experimental set ups considered. We present our evaluations on the Adult data set \cite{dheeru2017uci} in \textit{Figure 2}. Complete details for the DP Plug-in algorithm, its non-asymptotic analysis, and for our experimental set up and methodology, can be found in section D of the supplement.

\section{Conclusion and Future Work}
Our main contributions in this paper included $(1)$ proving the plug-in algorithm of \cite{menon2018cost} is consistent, $(2)$ characterising the sample complexity of learning fairness-aware BOCs via the plug-in algorithm, and $(3)$ proposing an easy to deploy, privacy-preserving protocol for the plug-in algorithm. As future directions, we believe it would be valuable to extend our analysis to the case where the sensitive attribute in non-binary; the case where multiple attributes are deemed sensitive. It would also be beneficial to study the statistical properties of learning algorithms across other settings, such as those demanding individual fairness, model explainability, or intersections between such areas of ethical and practical importance.

\clearpage
\bibliography{main.bib}
\bibliographystyle{icml2021}

\onecolumn
\icmltitle{Supplement for Statistical Guarantees for Fairness Aware Plug-In Algorithms}
\icmltitlerunning{Supplement for Statistical Guarantees for Fairness Aware Plug-In Algorithms}
\appendix
\section{Background}
In this section we summarise some key results from \cite{menon2018cost}. In particular, in the first subsection, we introduce our fairness metrics and measures of interest and we define the notion of cost-sensitive risk (CSR). We present the relationship between approximate notions of fairness and CSRs, as outlined in \cite{menon2018cost}. This will make clear the justification behind the use of cost-sensitive risks in formalising our fairness aware learning problem (Problem 2.3 in the main text). In subsequent subsections, we introduce the Bayes Optimal Classifiers and plug-in algorithms corresponding to four settings of interest, which are : 1) Approximate Equality of Opportunity when the sensitive attribute is unavailable at test time, 2) Approximate Equality of Opportunity when the sensitive attribute is available at test time, 3) Approximate Demographic Parity when the sensitive attribute is unavailable at test time and 4) Approximate Demographic Parity when the sensitive attribute is available at test-time.

\subsection{Cost Sensitive Risks and Fairness Aware Learning}
Access to a finite training sample, \small{$S = \left\{x_{i}, y_{i}, \overline{y}_{i}\right\}_{i=1}^{n}$ }drawn i.i.d from some unknown distribution $\mathbbm{P}$ is assumed in \cite{menon2018cost}. $\smash{\forall i \in [n]}$ the triplet $\left(x_{i}, y_{i}, \overline{y}_{i}\right)$ is a realisation of the random variable triplet \small{$\left(X, Y, \overline{Y}\right)$} comprising of the feature, label and sensitive attribute respectively. The features, $X$ are supported on some measurable domain $\chi$, whereas the labels, $Y$, and sensitive attributes, $\overline{Y}$, are binary and take values in $\{-1, 1\}$. The general convention is to assume $\overline{Y} = -1$ corresponds to the traditionally underprivileged or minority subgroup, whereas $\overline{Y} = 1$ corresponds to the traditionally privileged or majority subgroup of the population. Similarly, $Y = -1$ denotes the unfavourable or negative outcome, whereas $Y = 1$ denotes the favourable or positive outcome. We adopt these conventions throughout our work. \\\\
Let $ \small{\smash{\pi = \mathbbm{P}(Y = 1), \hspace{0.3em} \beta = \mathbbm{P}(\overline{Y} = 1)}}$ and, $\small{\smash{\beta = \mathbbm{P}(\overline{Y} = 1 \vert Y = 1)}}$. We assume throughout that $\pi, \beta$ and $\beta$ are non-zero. Further, let \small{$\smash{\left(X, Y\right) \sim \mathcal{D}}$}, \small{$\smash{\left(X, \overline{Y}\right) \sim \mathcal{D}_{DPar}}$}  and \small{$\smash{\left(X, \overline{Y} \vert Y = 1\right) \sim \overline{\mathcal{D}}_{EO}}$}. $\overline{\mathcal{D}}$ may be used to denote either one of $\mathcal{D}_{DPar}, \mathcal{D}_{EO}$. Now, let $\smash{f:\chi \rightarrow[0, 1]}$ denote a binary classifier on measurable domain $\chi. \,\, f$ yields predictions via \small{$\smash{(\hat{Y} \vert X = x)\ \sim Bernoulli(f(x))}$}. \\

Let \small{$\smash{\eta(x) = \mathbbm{P}(Y = 1 \vert X = x)}$} denote the regression function. Further, let \small{$\smash{\overline{\eta}_{DPar}(x) = \mathbbm{P}(\overline{Y} = 1 \vert X=x)}$} and, \small{$\smash{\overline{\eta}_{EO}(x, y) = \mathbbm{P}(\overline{Y} = 1 \vert X=x, Y=y)}$}. Again, $\overline{\eta}$ may be used to denote either $\overline{\eta}_{DPar}$ or $\overline{\eta}_{EO}$. Unless specified otherwise, estimators for each of these quantities will be denoted by the quantity along with the 'hat' notation on top. For example: $\hat{\eta}_{DPar}$ is the estimator corresponding to $\eta_{DPar}$. The false positive and false negative rates of a classifier $f$ with respect to a distribution $\mathcal{D}$ are defined as follows:
\[
FNR_{\mathcal{D}}(f) := \mathbbm{E}_{X \vert Y = 1}[1-f(x)] \hspace{0.3em}; \hspace{0.3em} FPR_{\mathcal{D}}(f) := \mathbbm{E}_{X \vert Y = -1}[f(x)]
\]
A central object in \cite{menon2018cost}, is the cost-sensitive risk of a classifier, defined below: 
\\\\
{\bf Definition A.1} {\hspace{0.1em} \it The cost sensitive risk of a classifier $f$:$\chi \rightarrow [0, 1]$ with respect to a distribution $\mathcal{D}$, parameterised by $c\in\left(0, 1\right)$ is given by:
\setlength\abovedisplayskip{5.5pt}
\begin{small}
\[
CS(f; \mathcal{D}, c) := c\left(1-\pi\right)FPR\left(f;\mathcal{D}\right) + \pi(1-c)FNR\left(f;\mathcal{D}\right)
\]
\end{small}
}
The cost sensitive risk of a classifier is thus a convex combination of its errors and serves as a measure of (weighted/ balanced) accuracy. The cost sensitive risk allows us to custom weigh the misclassification errors. Indeed, in certain domains, the cost associated with a false negative may be significantly higher than that associated with a false positive (eg: disease diagnosis), or vice-versa.  We also define the balanced cost sensitive risk of a classifier, which is a convex combination of its false positive and false negative rates. 
\\\\
{\bf Definition A.2} {\hspace{0.1em} \it The balanced cost sensitive risk of a classifier $f$:$\chi \rightarrow [0, 1]$ with respect to a distribution $\mathcal{D}$, parameterised by $c\in\left(0, 1\right)$ is given by:
\setlength\abovedisplayskip{5.5pt}
%\begin{small}
\[
CS_{bal}(f; \mathcal{D}, c) := cFPR\left(f;\mathcal{D}\right) + (1-c)FNR\left(f;\mathcal{D}\right)
\]
%\end{small}
}
Let us now define our fairness metrics of interest, namely Equality of Opportunity (EO) and Demographic Parity (DPar). 
\\\\
{\bf Definition A.3} {\hspace{0.1em} \it A binary classifier $f$, with corresponding predictor $\widehat{Y}$ admits Equality of Opportunity if:
\[
\mathbbm{P}(\hat{Y} = 1 \vert Y = 1, \overline{Y} = -1) = \mathbbm{P}(\hat{Y} = 1 \vert Y = 1, \overline{Y} = 1)
\]
}
{\bf Definition A.4} {\hspace{0.1em} \it A binary classifier $f$, with corresponding predictor $\widehat{Y}$ admits Demographic Parity if:
\[
\mathbbm{P}(\hat{Y} = 1 \vert \overline{Y} = -1) = \mathbbm{P}(\hat{Y} = 1 \vert \overline{Y} = 1)
\]
}
EO and DPar are commonly used metrics across the algorithmic fairness literature. EO requires parity in true positive rates (TPRs) between the two groups, i.e., $\overline{Y} = -1$ and $\overline{Y} = 1$. Whereas DPar requires parity in overall positive rates between the two groups. It is well established however, that attempting to enforce parity (i.e., perfect fairness) typically leads to unacceptable degradation in measures of utility (such as accuracy or the F1 score). It is therefore useful to define approximate notions of fairness via fairness measures. The goal is then to train models to achieve reasonably high accuracy while ensuring they admit approximate fairness (specified typically via constraints on these fairness measures). \cite{menon2018cost} make use of two fairness measures of interest, namely, Disparate Impact (DI) and Mean Difference (MD): \\\\
{\bf Definition A.5} {\hspace{0.1em} \it The Disparate Impact of a classifier $f$ with corresponding predictor $\hat{Y}$, is given by
\[
\frac{\mathbbm{P}(\hat{Y}=1 \vert \overline{Y} = -1)}{\mathbbm{P}(\hat{Y}=1 \vert \overline{Y} = 1)}
\]
}
{\bf Definition A.6} {\hspace{0.1em} \it The Mean Difference of a classifier $f$ with corresponding predictor $\hat{Y}$, is given by
\[
\mathbbm{P}(\hat{Y}=1 \vert \overline{Y} = -1) - \mathbbm{P}(\hat{Y}=1 \vert \overline{Y} = 1)
\]
}
In the notation of \cite{menon2018cost}, $R_{fair}$ may be used to refer to either fairness measure, DI or MD. Notice that DI is a multiplicative measure of disparity between rates, whereas MD is an additive measure of disparity between rates. By rates, we are referring of course to the overall positive rates, and thus we DI and MD as defined, serve as fairness measures for the DPar metric. However, the same measures extend analogously to the EO metric, by simply replacing overall positive rates with true positive rates. Typically, to ensure a classifier is fair, we would require imposing that its DI be somewhere close to 1. Since the numerator in DI corresponds to the positive rate of the underprivileged group, we may impose a constraint on the DI to be larger than 0.8 (this corresponds to the rule of four-fifths, used widely in the hiring domain \cite{eeoc1979uniform}). However, there is often a danger with imposing such a constraint, since DI is unbounded above, and this can lead to learning anti-classifiers that yield poor fairness scores via trivial transformations of the classifier. To avoid this, \cite{menon2018cost}, introduce a symmetrised notion of DI, that requires the ratio of positive rates and the ratio of negative rates to simultaneously be larger than some desired threshold. Similarly, a symmetrised notion for MD is also introduced in \cite{menon2018cost}. We refer the reader to \cite{menon2018cost} for a thorough discussion on this technicality. Now that we have key definitions in place, we compile key results from \cite{menon2018cost} that allow us to reduce fairness aware learning (i.e., learning with constraints on fairness measures) to learning with constraints on cost-sensitive risks. Note, the following results are equally applicable to the EO and DPar metrics, thus our use of the fairness measures DI and MD should be seen as metric agnostic from hereon. We will refer to the paradigm of learning with constraints imposed on fairness measures as that of approximate fairness.\\

In the language of \cite{menon2018cost}, we denote the Disparate Impact of a classifier $f$, by $DI(f)$, and the ratio of overall negative rates as $DI(1-f)$. $MD(f)$ and $MD(1-f)$ are analogously defined. We now recall a key result from \cite{menon2018cost}, which establishes an equivalence between the super-level sets of the disparate impact measure and the balanced cost sensitive risk for any classifier. 
\vspace{1.5mm}

{\bf Lemma A.7} {\hspace{0.1em} \it Pick any randomised classifier f. Then for any $\tau \in \left(0, \infty \right)$, if $\kappa = \frac{\tau}{1+\tau} \in \left(0, 1\right)$, 
\[
DI\left(f\right) \geq \tau \iff CS_{bal}\left(f; \overline{\mathcal{D}}, 1-\kappa \right) \geq \kappa
\]
}
\cite{menon2018cost} also present a similar, in fact, stronger that establishes an equivalence between the super-level sets of the mean difference measure and the balanced cost sensitive risk for any classifier. \\\\
{\bf Lemma A.8} {\hspace{0.1em} \it Pick any randomised classifier f. Then for any $\tau \in \left(-1, 1 \right)$, if $\kappa = \frac{1+\tau}{2} \in \left(0, 1\right)$, 
\[
MD(f) \geq \tau \iff CS_{bal}(f; \mathcal{D}, \frac{1}{2}) \geq \kappa
\]
}
Equipped with Lemmas A.7 and/ or A.8, \cite{menon2018cost} show the fairness-aware learning problem can be reduced to a problem with constraints on cost sensitive risks. \\\\
{\bf Lemma A.9} {\hspace{0.1em} \it Pick any distributions $\mathcal{D}, \overline{\mathcal{D}}$, and fairness measure $R_{fair} \in \{DI, MD\}$. Pick any $c, \overline{\tau} \in (0, 1)$. Then $\exists \lambda \in \mathbbm{R}, \overline{c} \in (0, 1)$ with 
\[
min_{f} CS(f; \mathcal{D}, c): min(R_{fair}(f; \overline{D}), R_{fair}(1-f; \overline{D})) \geq \tau \equiv min_{f} CS(f; \mathcal{D}, c) - \lambda CS(f; \overline{\mathcal{D}}, \overline{c})
\]

}

\textit{Remark: We may use the balanced cost sensitive risk, i.e., $CS_{bal}(f; \overline{\mathcal{D}}, \overline{c})$ in place of the standard cost sensitive risk, i.e., $CS(f; \overline{\mathcal{D}}, \overline{c})$ for the second term in the soft-constrained objective of Lemma A.9. However, the (asymptotic and non-asymptotic) analyses derived for the formulation using the standard cost sensitive risk applies directly to the formulation wherein we would instead use the balanced cost sensitive risk. Doing so would simply require dropping a constant term in the objective, and then proceeding as per usual. Indeed, the balanced cost sensitive risk formulation can be seen as a special case of the standard cost-sensitive risk formulation. Much of the analysis in \cite{menon2018cost} also works with the more general standard cost-sensitive risk (it is implicitly assumed that their analysis would extend to the case of balanced cost sensitive risk). We thus focus on the notion of standard cost sensitive risk throughout our paper.}
\\\\
We refer the reader to \cite{menon2018cost} for a more detailed discussion and derivations of the results compiled in this section. Having done so, we have thus established the relationship between FAL and cost-sensitive risks. We therefore obtain Problem 2.3 in the main text. 
\subsection{Approximate Equality of Opportunity when the sensitive attribute is unavailable at test time}
 Denoting $\overline{\eta}$ as $\overline{\eta}_{EO}$ in this subsection. For trade-off parameter $\lambda \in \mathbbm{R}$, and cost parameters $c, \overline{c} \in \left(0, 1\right)^{2}$, and $\forall x \in \chi$, the Bayes Optimal Classifier in this setting \cite{menon2018cost} is given by: 

\[f^{*}(x) = H_{\alpha}\circ s^{*}(x), \hspace{0.25 em} for \hspace{0.25em} any \hspace{0.25em} \alpha \in [0, 1]\] 
\[s^{*}(x) = \left\{ 1 - \frac{\lambda}{\pi}\left(\overline{\eta}(x, 1) - \overline{c}\right) \right\}\eta(x) - c,
\hspace{0.25em}and,\]
\[ H_{\alpha}(z) = \mathbbm{I}\{z > 0\} + \alpha\mathbbm{I}\{z = 0\} \hspace{0.25em} is\hspace{0.25em}  the\hspace{0.25em}  Heaviside \hspace{0.25em} function\] 
\\\\
The plug-in algorithm to estimate the BOC in this setting \cite{menon2018cost} is given below:

\begin{algorithm}
   \caption{\small{Plugin approach to FAL, EO setting, $\overline{Y}$ unavailable at test time}}
\begin{algorithmic}

   \STATE \small{{\bfseries Input:} Sample $S$ = $\{x_{i}, y_{i}, \overline{y}_{i}\}_{i=1}^{n}$ from distribution $\mathbbm{P}$; cost parameters $c, \overline{c}$; trade-off parameter $\lambda$}
   \vspace{2.5mm}
   \STATE {\bfseries Estimate:} \small{$\pi$ via \hspace{0.5 em}$\hat{\pi} = \frac{1}{n}\sum_{i=1}^{n}\mathbbm{I}\{y_{i} = 1\}$}
   \STATE {\bfseries Estimate:} \small{$\eta$:$\chi \rightarrow [0, 1]$ using appropriate CPE on $\{x_{i}, y_{i}\}_{i=1}^{n}$}
   \STATE {\bfseries Estimate:} \small{$\overline{\eta}$: $(\chi, Y) \rightarrow [0, 1]$ using appropriate CPE on $S$}
   \STATE {\bfseries Compute:} \small{$\hat{s}\left(x\right) =\left\{ 1 - \frac{\lambda}{\hat{\pi}}(\hat{\overline{\eta}}(x, 1) - \overline{c}) \right\}\hat{\eta}(x) - c$}
   \vspace{2.5mm}
   \STATE {\bfseries Return:} \small{$\hat{f}\left(x\right) = H_{\alpha}\left(\hat{s}\left(x\right)\right)$ for any $\alpha \in [0, 1]$}
\end{algorithmic}
\end{algorithm}
\textit{Remark: Since $\alpha$ can be arbitrarily chosen, we simply set $\alpha = 0$ in the main text. This yields a deterministic BOC of the form presented in Theorem 2.4 of the main text. Similarly the plug-in estimator corresponding to $\alpha = 0$ is presented in Algorithm 1. Throughout the main text, as well as in sections B, C and D of the supplement, we will set $\alpha = 0$, yielding a deterministic binary BOC and a deterministic plug-in estimator both of which will take values in $\{-1, 1\}$. There is nothing special about our decision to set $\alpha = 0$, as when $\alpha \neq 0$, the BOC is still deterministic almost everywhere. We make this remark explicit \textbf{only} to avoid confusion with the notion of randomised classifiers (in the language of \cite{menon2018cost}) used in this section, as opposed to that of deterministic binary classifiers presumed elsewhere. Indeed, the BOC, while deterministic when $\alpha = 0$, is an element of the set of randomised classifiers. }

\subsection{Approximate Equality of Opportunity when the sensitive attribute is available at test time}
Denoting $\overline{\eta}$ as $\overline{\eta}_{EO}$ in this subsection. For trade-off parameter $\lambda \in \mathbbm{R}$, and cost parameters $c, \overline{c} \in \left(0, 1\right)^{2}$, $\forall x \in \chi$ and for any $\overline{y} \in \{-1, 1\}$, the Bayes Optimal Classifier in this setting \cite{menon2018cost} is given by:

\[f^{*}(x, \overline{y}) = H_{\alpha}\circ s^{*}(x, \overline{y}), \hspace{0.25 em} for \hspace{0.25em} any \hspace{0.25em} \alpha \in [0, 1]\] 
\[s^{*}(x, -1) = \left\{ 1 + \frac{\lambda}{\pi}\overline{c}\right \}\eta(x, -1) -c \hspace{0.35em};\hspace{0.35em} 
s^{*}(x, 1) = \left\{ 1 - \frac{\lambda}{\pi}(1-\overline{c})\right \}\eta(x, 1) -c \quad and, 
\]
\[ H_{\alpha}(z) = \mathbbm{I}\{z > 0\} + \alpha\mathbbm{I}\{z = 0\} \hspace{0.25em} is\hspace{0.25em}  the\hspace{0.25em}  Heaviside \hspace{0.25em} function\] 
The plug-in algorithm to estimate the BOC in this setting \cite{menon2018cost} is given below:
\begin{algorithm}[H]
   \caption{\small{Plugin approach to FAL, EO setting, $\overline{Y}$ available at test time}}
\begin{algorithmic}

   \STATE \small{{\bfseries Input:} Sample $S$ = $\{x_{i}, y_{i}, \overline{y}_{i}\}_{i=1}^{n}$ from distribution $\mathbbm{P}$; cost parameters $c, \overline{c}$; trade-off parameter $\lambda$}
   \vspace{2.5mm}
   \STATE {\bfseries Estimate:} \small{$\pi$ via \hspace{0.5 em}$\hat{\pi} = \frac{1}{n}\sum_{i=1}^{n}\mathbbm{I}\{y_{i} = 1\}$}
   \STATE {\bfseries Estimate:} \small{$\eta$:$(\chi, \overline{Y}) \rightarrow [0, 1]$ using appropriate CPE on $S$}
   \STATE {\bfseries Compute:} \small{$\hat{s}\left(x, \overline{y}\right) =$ $\hat{\eta}(x, \overline{y})\left[1 + \frac{\lambda\overline{c}}{\hat{\pi}} - \frac{\lambda\mathbbm{I}\{\overline{y} = 1\}}{\hat{\pi}}\right] - c$ }
   \vspace{2.5mm}
   \STATE {\bfseries Return:} \small{$\hat{f}\left(x, \overline{y}\right) = H_{\alpha}\left(\hat{s}\left(x, \overline{y}\right)\right)$ for any $\alpha \in [0, 1]$}
\end{algorithmic}
\end{algorithm}

\subsection{Approximate Demographic Parity when the sensitive attribute is unavailable at test time}
Denoting $\overline{\eta}$ as $\overline{\eta}_{DPar}$ in this subsection. For trade-off parameter $\lambda \in \mathbbm{R}$, and cost parameters $c, \overline{c} \in \left(0, 1\right)^{2}$, and $\forall x \in \chi$, the Bayes Optimal Classifier in this setting \cite{menon2018cost} is given by:

\[f^{*}(x) = H_{\alpha}\circ s^{*}(x), \hspace{0.25 em} for \hspace{0.25em} any \hspace{0.25em} \alpha \in [0, 1]\] 
\[s^{*}(x) = \eta(x) - \{c + \lambda(\overline{\eta}(x) - \overline{c})\}\quad and, 
\]
\[ H_{\alpha}(z) = \mathbbm{I}\{z > 0\} + \alpha\mathbbm{I}\{z = 0\} \hspace{0.25em} is\hspace{0.25em}  the\hspace{0.25em}  Heaviside \hspace{0.25em} function\] 
The plug-in algorithm to estimate the BOC in this setting \cite{menon2018cost} is given below:

\begin{algorithm}
   \caption{\small{Plugin approach to FAL, DPar setting, $\overline{Y}$ unavailable at test time}}
\begin{algorithmic}

   \STATE \small{{\bfseries Input:} Sample $S$ = $\{x_{i}, y_{i}, \overline{y}_{i}\}_{i=1}^{n}$ from distribution $\mathbbm{P}$; cost parameters $c, \overline{c}$; trade-off parameter $\lambda$}
   \vspace{2.5mm}
   \STATE {\bfseries Estimate:} \small{$\eta$:$\chi \rightarrow [0, 1]$ using appropriate CPE on $\{x_{i}, y_{i}\}_{i=1}^{n}$}
   \STATE {\bfseries Estimate:} \small{$\overline{\eta}$: $\chi \rightarrow [0, 1]$ using appropriate CPE on $\{x_{i}, \overline{y}_{i}\}_{i=1}^{n}$}
   \STATE {\bfseries Compute:} \small{$\hat{s}\left(x\right) = \hat{\eta}(x) - \{c + \lambda(\hat{\overline{\eta}}(x) - \overline{c})\}$}
   \vspace{2.5mm}
   \STATE {\bfseries Return:} \small{$\hat{f}\left(x\right) = H_{\alpha}\left(\hat{s}\left(x\right)\right)$ for any $\alpha \in [0, 1]$}
\end{algorithmic}
\end{algorithm}

\subsection{Approximate Demographic Parity when the sensitive attribute is available at test time}
Denoting $\overline{\eta}$ as $\overline{\eta}_{DPar}$ in this subsection. For trade-off parameter $\lambda \in \mathbbm{R}$, and cost parameters $c, \overline{c} \in \left(0, 1\right)^{2}$, $\forall x \in \chi$ and for any $\overline{y} \in \{-1, 1\}$, the Bayes Optimal Classifier in this setting \cite{menon2018cost} is given by:

\[f^{*}(x, \overline{y}) = H_{\alpha}\circ s^{*}(x, \overline{y}), \hspace{0.25 em} for \hspace{0.25em} any \hspace{0.25em} \alpha \in [0, 1]\] 
\[s^{*}(x, -1) = \eta(x, -1) - c + \lambda\overline{c} \quad and, 
s^{*}(x, 1) = \eta(x, 1) - c + \lambda\overline{c} - \lambda
\]
\[ H_{\alpha}(z) = \mathbbm{I}\{z > 0\} + \alpha\mathbbm{I}\{z = 0\} \hspace{0.25em} is\hspace{0.25em}  the\hspace{0.25em}  Heaviside \hspace{0.25em} function\] 

The plug-in algorithm to estimate the BOC in this setting \cite{menon2018cost} is given below:

\begin{algorithm}
   \caption{\small{Plugin approach to FAL, DPar setting, $\overline{Y}$ available at test time}}
\begin{algorithmic}

   \STATE \small{{\bfseries Input:} Sample $S$ = $\{x_{i}, y_{i}, \overline{y}_{i}\}_{i=1}^{n}$ from distribution $\mathbbm{P}$; cost parameters $c, \overline{c}$; trade-off parameter $\lambda$}
   \vspace{2.5mm}
   \STATE {\bfseries Estimate:} \small{$\eta$:$(\chi, \overline{Y})\rightarrow [0, 1]$ using appropriate CPE on $\{x_{i}, y_{i}\}_{i=1}^{n}$}
   \STATE {\bfseries Compute:} \small{$\hat{s}\left(x, \overline{y}\right) = \hat{\eta}(x, \overline{y}) - c + \lambda\overline{c} -\lambda\mathbbm{I}\{\overline{y} = 1\}$}
   \vspace{2.5mm}
   \STATE {\bfseries Return:} \small{$\hat{f}\left(x, \overline{y}\right) = H_{\alpha}\left(\hat{s}\left(x, \overline{y}\right)\right)$ for any $\alpha \in [0, 1]$}
\end{algorithmic}
\end{algorithm}
We recall from the main text, that the fairness-aware learning (FAL) problem may be specified as follows:
\\\\
{ (\textbf{Cost-sensitive FAL}) \it For trade-off parameter $\lambda \in \mathbbm{R}$, and cost parameters $c, \overline{c} \in \left(0, 1\right)^{2}$, minimise the fairness-aware cost-sensitive risk:
\vspace*{-1.57mm}
\[
R_{FA}(f; \mathcal{D}, \overline{\mathcal{D}}_{EO}, c, \overline{c}, \lambda) = CS(f; \mathcal{D}, c) - \lambda CS(f; \overline{\mathcal{D}}, \overline{c})
\]
}
\\
In the language of \cite{narasimhan2014statistical}, we introduce the notion of performance measure. A performance measure w.r.t a distribution $\mathbbm{P}$ and performance metric $\Psi$ is a mapping from the space of measurable functions $\mathcal{F}$ to the reals, i.e., $\mathfrak{P}_{\mathbbm{P}}^{\Psi}: \mathcal{F} \rightarrow \mathbbm{R}$.  In our case, the performance metric is simply given by the negative of the objective function of Problem 2.3. Thus, a classifier's performance measure explains its merit with regards to the combined, utility plus fairness objective, appropriately balanced by cost and trade-off parameters. Performance metric $\Psi$ makes explicit the fact that our performance measure is a function of the classifier's TPRs and TNRs with respect to $\mathcal{D}$ and $\overline{\mathcal{D}}$, as well as distributional quantities $\pi$ and $\beta$. Our performance measure is given by:
\vspace*{-1mm}
\[
\mathfrak{P}_{\mathbbm{P}}^{\Psi}(f) = \Psi\left[TPR_{\mathcal{D}}(f), TNR_{\mathcal{D}}(f), \pi, TPR_{\overline{\mathcal{D}}}(f), TNR_{\overline{\mathcal{D}}}(f), \beta\right] 
\]
\[
= - \left\{CS\left(f; \mathcal{D}, c\right) \hspace{0.1em} - \hspace{0.1em}  \lambda CS\left(f; \overline{\mathcal{D}}, \overline{c}\right)\right\}
\]
\\\\
The objective of the FAL problem is therefore equivalent to maximising the aforementioned performance measure. In subsequent sections of the supplement, we will analyse the plug-in algorithm's efficacy (in the asymptotic and finite sample settings) with respect to our objective. As pointed out in the previous remark, across all our settings of interest, the BOC and corresponding plug-in estimators are deterministic when $\alpha = 0$ and are deterministic almost everywhere when $\alpha \neq 0$. Since $\alpha$ can be arbitrarily chosen in $[0, 1]$, i.e., the entire range of values yield BOCs, we will set $\alpha = 0$ in the remainder of this paper and work in the realm of deterministic binary classifiers taking values in $\{-1, 1\}$, rather than randomised classifiers. Please turn over to the next page. 
\newpage
\section{Asymptotic Analysis for the Plug-In Algorithm}
In this section, we will focus on the asymptotic analysis of the plug-in algorithm with respect to the performance measure of interest. The subsequent subsections will cover four settings of interest, namely, 1) Approximate Equality of Opportunity when the sensitive attribute is unavailable at test time, 2) Approximate Equality of Opportunity when the sensitive attribute is available at test time, 3) Approximate Demographic Parity when the sensitive attribute is unavailable at test time and 4) Approximate Demographic Parity when the sensitive attribute is available at test-time. 

\subsection{Approximate Equality of Opportunity when the sensitive attribute is unavailable at test time}
In this subsection $\overline{\mathcal{D}}$ will be used to refer to $\overline{\mathcal{D}}_{EO}$. Recall, $\pi = \mathbbm{P}(Y=1)$ and $\beta = \mathbbm{P}(\overline{Y}=1\vert Y=1)$. Assuming $\pi, \beta > 0$, our performance measure is given by: 
\[
\mathfrak{P}_{\mathbbm{P}}^{\Psi}(f) = \Psi\left[TPR_{\mathcal{D}}(f), TNR_{\mathcal{D}}(f), \pi, TPR_{\overline{\mathcal{D}}}(f), TNR_{\overline{\mathcal{D}}}(f), \beta\right] 
= - \left\{CS\left(f; \mathcal{D}, c\right) \hspace{0.1em} - \hspace{0.1em}  \lambda CS\left(f; \overline{\mathcal{D}}, \overline{c}\right)\right\}
\]
\[
= -\{c\left(1-\pi\right)FPR_{\mathcal{D}}(f) + \pi(1-c)FNR_{\mathcal{D}}(f)\} + \lambda\{\overline{c}\left(1-\beta\right)FPR_{\overline{\mathcal{D}}}(f) + \beta(1-\overline{c})FNR_{\overline{\mathcal{D}}}(f)\}
\]
Where we make use of the definition of cost sensitive risk as laid out in the section A.1 of the supplement. Thus, the performance measure is linear in the arguments of $\Psi$, and therefore continuous in these arguments. The regret of a classifier $f$, w.r.t performance measure $\mathfrak{P}_{\mathbbm{P}}^{\Psi}$ is defined as: 
\[
regret_{\mathbbm{P}}^{\Psi}(f) = \mathfrak{P}_{\mathbbm{P}}^{\Psi, *}- \mathfrak{P}_{\mathbbm{P}}^{\Psi}(f)
\]
where, $\mathfrak{P}_{\mathbbm{P}}^{\Psi, *} = \mathfrak{P}_{\mathbbm{P}}^{\Psi}(f^{*})$. Here $f^{*}$ is the Bayes Optimal Classifier corresponding to the Equal Opportunity case. Denoting $\overline{\eta}$ as $\overline{\eta}_{EO}$ in this subsection, we have that $\forall x \in \chi$:

\[f^{*}(x) = sign\circ s^{*}(x), \hspace{0.25 em} for \hspace{0.25em} any \hspace{0.25em} \alpha \in [0, 1]\] 
\[where, \hspace{0.3em} s^{*}(x) = \left\{ 1 - \frac{\lambda}{\pi}\left(\overline{\eta}(x, 1) - \overline{c}\right) \right\}\eta(x) - c\] 

We now prove that the plug-in procedure yields an estimator $\hat{f}$ which is $\Psi$-consistent, implying that $ regret_{\mathbbm{P}}^{\Psi}(\hat{f}) \overset{p}{\to} 0$, where $\overset{p}{\to}$ denotes convergence in probability. We denote the estimators of $\eta, \overline{\eta}$ by $\hat{\eta}$ and $\hat{\overline{\eta}}$ respectively. In order to proceed we make the following assumptions: \\\\
{\bf Assumption 1} \textit{{\hspace{0.1em} $\mathbbm{P}_{X \vert Y=1}\left(\gamma(x) \leq k\right), \mathbbm{P}_{X \vert Y=-1}\left(\gamma(x) \leq k\right), \\ \mathbbm{P}_{X \vert Y=1,  \overline{Y}=1}\left(\gamma(x) \leq k\right)$ and $\mathbbm{P}_{X \vert Y=1, \overline{Y}=-1}\left(\gamma(x) \leq k\right)$ are continuous at $k =c$, where $\gamma(x) = (1 + \frac{\lambda\overline{c}}{\pi})\eta(x) - \frac{\lambda}{\pi}\overline{\eta}(x, 1)\eta(x)
$, i.e, $\gamma(x)$ is $s^{*}(x)$ without the constant term $-c$}}
\\\\
{\bf Assumption 2} \textit{{\hspace{0.1em} Class probability estimators (CPEs) $\hat{\eta}, \hat{\overline{\eta}}$ are $L1$-consistent, i.e., 
\[
\mathbbm{E}_{X}\left[\vert \eta(x) - \hat{\eta}(x) \vert\right]  \overset{p}{\to} 0;\quad
\mathbbm{E}_{X, Y}\left[\vert \overline{\eta}(x, y) - \hat{\overline{\eta}}(x, y) \vert\right]  \overset{p}{\to} 0 \]
}}
\vspace*{-1mm}
{\it Remark: As noted in \cite{narasimhan2014statistical, chzhen2019leveraging}, Assumption 2 is not a very strong one, as an appropriately regularized ERM yields an $L$-1 consistent class probability estimator for proper losses \cite{menon2013statistical, agarwal2013surrogate}.}
\\\\
{\bf Assumption 3} \textit{{\hspace{0.1em} Domain $\chi$ is compact and there exist constants $a, B \in \mathbbm{R}_{+}$, such that the PDFs, $f_{X \vert Y = -1}, f_{X}, f_{X \vert Y = 1}$ satisfy $\forall x \in \chi, \hspace{0.25em}0 < a \leq f_{X \vert Y = -1}(x),f_{X}(x), f_{X \vert Y = 1}(x) \leq B$
}}
\\\\
{\it Remark: We make this assumption for technical convenience. This is akin to the 'strong density assumption' defined in \cite{audibert2007fast}. This assumption is not necessary for the case when $\overline{Y}$ is available at test time, or for either case relating to the approximate Demographic Parity criterion}
\\\\
We will denote the estimator derived via the plug-in procedure for $\gamma(x)$ by $\hat{\gamma}(x) = (1 + \frac{\lambda\overline{c}}{\hat{\pi}})\hat{\eta}(x) - \frac{\lambda}{\hat{\pi}}\hat{\overline{\eta}}(x, 1)\hat{\eta}(x)$. We now prove the helper results that allow us to prove Lemma 3.1 of the main section. 
\\\\
{\bf Lemma B.1} \textit{Provided Assumptions 2 and 3 hold, $\smash{\mathbbm{E}_{X}\left[\vert \overline{\eta}(x, 1) - \hat{\overline{\eta}}(x, 1) \vert\right]  \overset{p}{\to} 0}$}
\begin{proof}\[
\mathbbm{E}_{X, Y}\left[\vert \hat{\overline{\eta}}(x, y) - \overline{\eta}(x, y) \vert\right] = \bigintsss_{x \in \chi}\left\{ \sum_{y \in \{-1, 1\}}\vert \hat{\overline{\eta}}(x, y) - \overline{\eta}(x, y)  \vert f_{X, Y}(x, y) \right\} dx
\]
\[
= \pi \underbrace{\bigintsss_{x \in \chi} \vert \hat{\overline{\eta}}(x, 1) - \overline{\eta}(x, 1)  \vert f_{X \vert Y = 1}(x) dx}_\text{term 1} + (1-\pi) \underbrace{\bigintsss_{x \in \chi} \vert \hat{\overline{\eta}}(x, -1) - \overline{\eta}(x, -1)  \vert f_{X \vert Y = -1}(x) dx}_\text{term 2}
\]
Notice that by Assumption 2, $\mathbbm{E}_{X, Y}\left[\vert \hat{\overline{\eta}}(x, y) - \overline{\eta}(x, y) \vert\right] \overset{p}{\to} 0$ and that terms 1 and 2 are non negative, implying that they must converge to 0 in probability as well. Thus, we have that:
\[
\mathbbm{E}_{X \vert Y = 1}[\vert \hat{\overline{\eta}}(x, 1) - \overline{\eta}(x, 1)  \vert] = \bigintsss_{x \in \chi} \vert \hat{\overline{\eta}}(x, 1) - \overline{\eta}(x, 1)  \vert f_{X \vert Y = 1}(x) dx \overset{p}{\to} 0
\]
Now, 
\[
\mathbbm{E}_{X}[\vert \hat{\overline{\eta}}(x, 1) - \overline{\eta}(x, 1)  \vert] = \bigintsss_{x \in \chi} \vert \hat{\overline{\eta}}(x, 1) - \overline{\eta}(x, 1)  \vert f_{X}(x) dx 
\]
There are two mutually exclusive and exhaustive cases that arise. In case 1, $\{x \in \chi: f_{X}(x) \leq f_{X \vert Y = 1}(x)\}$, and in case 2, $\{x \in \chi: f_{X}(x) > f_{X \vert Y = 1}(x)\}$. By Assumption 3, we know that, there exist constants $a, B \in \mathbbm{R}_{+}$, such that the PDFs, $f_{X \vert Y = -1}, f_{X}, f_{X \vert Y = 1}$ satisfy $\forall x \in \chi, \hspace{0.25em}0 < a \leq f_{X \vert Y = -1}(x),f_{X}(x), f_{X \vert Y = 1}(x) \leq B$. In case 2, $\{x \in \chi: a \leq f_{X\vert Y=1}(x) < f_{X}(x) \leq B \} \implies \{x \in \chi: \frac{f_{X}(x)}{f_{X\vert Y=1}(x)} \leq \frac{B}{a}\} \implies \{x \in \chi: f_{X}(x) \leq \frac{f_{X\vert Y=1}(x)B}{a} \} $. However, $a<B$, implying that $\forall x \in \chi, f_{X}(x) \leq \frac{f_{X\vert Y=1}(x)B}{a}$. Thus we have that:
\[
0 \leq \bigintsss_{x \in \chi} \vert \hat{\overline{\eta}}(x, 1) - \overline{\eta}(x, 1)  \vert f_{X}(x) dx \leq \frac{B}{a} \bigintsss_{x \in \chi} \vert \hat{\overline{\eta}}(x, 1) - \overline{\eta}(x, 1)  \vert f_{X \vert Y=1}(x) dx \overset{p}{\to} 0
\]
where the right most term above is simply a constant times term 1 and thus the convergence to 0 in probability. It follows by an application of the Squeeze Theorem that:
\[
 \mathbbm{E}_{X}[\vert \hat{\overline{\eta}}(x, 1) - \overline{\eta}(x, 1)  \vert] = \bigintsss_{x \in \chi} \vert \hat{\overline{\eta}}(x, 1) - \overline{\eta}(x, 1)  \vert f_{X}(x) dx  \overset{p}{\to} 0
\]
\end{proof}
{\bf Lemma B.2} Assuming $\pi > 0$, \textit{$\frac{1}{\hat{\pi}} \overset{p}{\to} \frac{1}{\pi}$}
\begin{proof}{ We first note that our estimator, \hspace{0.25em}$\hat{\pi} = \frac{1}{n}\sum_{i=1}^{n}\mathbbm{I}\{y_{i} = 1\}$ converges to $\pi$ in probability, by the Weak Law of Large Numbers, i.e., $\hat{\pi} \overset{p}{\to} \pi$. We can not directly apply the Continuous Mapping Theorem, since there is a non-zero probability that $\hat{\pi} = 0$. To make clear that we are working with a 'sequence' of random variables, we will denote $\hat{\pi}$ estimated via n samples by $\hat{\pi}_{n}$. Now notice we have to show that, for any $\epsilon > 0$: 
\[
\mathbbm{P}(\vert \frac{1}{\hat{\pi}_{n}} - \frac{1}{\pi} \vert > \epsilon) \to 0 \hspace{0.25em} as \hspace{0.25em} n \to \infty
\]
Now, 
\[
\Big\{\vert \frac{1}{\hat{\pi}_{n}} - \frac{1}{\pi} \vert > \epsilon \Big\} \subseteq \Big\{\vert \hat{\pi}_{n} - \pi \vert > \epsilon\hat{\pi}_{n}\pi : \hat{\pi}_{n} \geq \frac{\pi}{2}\Big\} \bigcup \Big\{\vert \hat{\pi}_{n} - \pi \vert > \epsilon\hat{\pi}_{n}\pi : \hat{\pi}_{n} < \frac{\pi}{2}\Big\} \subseteq  \Big\{\vert \hat{\pi}_{n} - \pi \vert > \epsilon\frac{\pi^{2}}{2} \Big\} \bigcup \Big\{\hat{\pi}_{n} < \frac{\pi}{2} \Big\}
\]
Which means, via a union bound that,
\begin{equation}
   0 \leq \mathbbm{P}(\vert \frac{1}{\hat{\pi}_{n}} - \frac{1}{\pi} \vert > \epsilon) \hspace{0.3em}\leq\hspace{0.3em} \underbrace{\mathbbm{P}(\vert \hat{\pi}_{n} - \pi \vert > \epsilon\frac{\pi^{2}}{2})}_\text{term 1}\hspace{0.3em} +\hspace{0.3em} \underbrace{\mathbbm{P}(\hat{\pi}_{n} < \frac{\pi}{2})}_\text{term 2} 
\end{equation}
Clearly, term 1 converges to 0 because $\hat{\pi}_{n} \overset{p}{\to} \pi$. As for term 2, it is true that: 
\[
0 \leq \mathbbm{P}(\hat{\pi}_{n} < \frac{\pi}{2}) \hspace{0.3em} \leq \hspace{0.3em} \underbrace{\mathbbm{P}(\vert \hat{\pi}_{n} - \pi \vert > \frac{\pi}{2})}_\text{term 3}
\]
Again, due to $\hat{\pi}_{n} \overset{p}{\to} \pi$, term 3 converges to 0 $\implies$ term 2 converges to 0 by the Squeeze Theorem. Thus, applying Squeeze Theorem again to $(1)$, $\mathbbm{P}(\vert \frac{1}{\hat{\pi}_{n}} - \frac{1}{\pi} \vert > \epsilon) \to 0$. Since we chose $\epsilon > 0$ arbitrarily, the result follows. 
}\end{proof}

{\bf Lemma B.3} \textit{$\hat{\eta}\hat{\overline{\eta}}(\cdot, 1)$} is $L$-1 consistent, i.e., $\mathbbm{E}_{X}[\vert \hat{\eta}\hat{\overline{\eta}}(x, 1) - \eta(x)\overline{\eta}(x, 1) \vert] \overset{p}{\to} 0$
\begin{proof} We have by Assumption 2 and Lemma B.1 respectively that, $\hat{\eta} \overset{L^{1}}{\to} \eta$ and, $\hat{\overline{\eta}}(\cdot, 1)  \overset{L^{1}}{\to} \overline{\eta}(\cdot, 1) $. Convergence in the $L$-1 norm implies convergence in probability and thus, $\hat{\eta} \overset{p}{\to} \eta$ and, $\hat{\overline{\eta}}(\cdot, 1)  \overset{p}{\to} \overline{\eta}(\cdot, 1) \implies \hat{\eta}\hat{\overline{\eta}}(\cdot, 1) \overset{p}{\to} \eta\overline{\eta}(\cdot, 1)$. Now, note that $\vert \hat{\eta}\hat{\overline{\eta}}(\cdot, 1) \vert$ is uniformly bounded above by 1, regardless of the number of samples used to obtain $\vert \hat{\eta}\hat{\overline{\eta}}(\cdot, 1) \vert$. This implies that $\vert \hat{\eta}\hat{\overline{\eta}}(\cdot, 1) \vert$ is uniformly integrable in $L$-1. Now convergence in probability and uniform integrability in $L$-1 for a sequence of random variables, implies convergence in $L$-1 to the same limit (as the sequence converges to in probability). Thus,  $\hat{\eta}\hat{\overline{\eta}}(\cdot, 1) \overset{L^{1}}{\to} \eta\overline{\eta}(\cdot, 1)$
\end{proof}

We are now ready to prove Lemma 3.1 of the main text, restating it below as Lemma B.4 (here): 
\\\\
{\bf Lemma B.4} \textit{{\hspace{0.1em} Provided Assumptions 2 and 3 hold, $\hat{\gamma}$ is $L$-1 consistent, i.e.,\hspace{0.25em}$\smash{ \mathbbm{E}_{X}\left[\vert \gamma(x) - \hat{\gamma}(x) \vert\right] \overset{p}{\to} 0}$}}
\begin{proof} First note that, using the triangle inequality, $\forall x \in \chi$: 
\[
0 \leq \vert \gamma(x) - \hat{\gamma}(x) \vert = \vert \{ \eta(x) + \lambda\overline{c}\frac{\eta(x)}{\pi} - \lambda\frac{\eta(x)\overline{\eta}(x, 1)}{\pi}\}  -\{ \hat{\eta}(x) + \lambda\overline{c}\frac{\hat{\eta}(x)}{\hat{\pi}} - \lambda\frac{\hat{\eta}(x)\hat{\overline{\eta}}(x, 1)}{\hat{\pi}}\}\vert 
\]
\[
\leq \vert \eta(x) - \hat{\eta}(x)  \vert + \vert\lambda\overline{c}\vert \hspace{0.25em} \vert \frac{\eta(x)}{\pi} - \frac{\hat{\eta}(x)}{\hat{\pi}} \vert + \vert \lambda\vert \hspace{0.25em} \vert \frac{\eta(x)\overline{\eta}(x, 1)}{\pi} - \frac{\hat{\eta}(x)\hat{\overline{\eta}}(x, 1)}{\hat{\pi}}  \vert
\]
Therefore, taking expectations we get that: 
\begin{equation}
   0 \leq \mathbbm{E}_{X}[\vert \gamma(x) - \hat{\gamma}(x) \vert] \leq \underbrace{\mathbbm{E}_{X}[\vert \eta(x) - \hat{\eta}(x)  \vert]}_\text{term 1} + \vert\lambda\overline{c}\vert \hspace{0.25em}\underbrace{\mathbbm{E}_{X}[ \vert \frac{\eta(x)}{\pi} - \frac{\hat{\eta}(x)}{\hat{\pi}} \vert]}_\text{term 2} + \vert \lambda\vert \hspace{0.25em} \underbrace{\mathbbm{E}_{X}[ \vert \frac{\eta(x)\overline{\eta}(x, 1)}{\pi} - \frac{\hat{\eta}(x)\hat{\overline{\eta}}(x, 1)}{\hat{\pi}}  \vert]}_\text{term 3} 
\end{equation}

By Assumption 2, term 1 converges in probability to 0. We reiterate that the randomness in the value of $\hat{\pi}$ and the functional form of estimators $\hat{\eta}$ and $\hat{\overline{\eta}}$ stems from the random draw of the training sample. Let's focus on term 2: 
\[
0 \leq \mathbbm{E}_{X}[ \vert \frac{\eta(x)}{\pi} - \frac{\hat{\eta}(x)}{\hat{\pi}} \vert] = \frac{1}{\pi\hat{\pi}}\mathbbm{E}_{X}[\vert \hat{\pi}\eta(x) - \pi\hat{\eta}(x) \vert] = \frac{1}{\pi\hat{\pi}}\mathbbm{E}_{X}[\vert \left\{\hat{\pi}\eta(x) - \pi\eta(x)  \right\} + \left(\pi\eta(x) - \pi\hat{\eta}(x)\right) \vert] 
\]
\[
\overset{(a)}{\leq} \hspace{0.4em}\frac{1}{\pi\hat{\pi}}\left\{\vert \hat{\pi} - \pi \vert \mathbbm{E}_{X}[\eta(x)] + \pi\mathbbm{E}_{X}[\vert \eta(x) - \hat{\eta}(x) \vert]\right\} \hspace{0.4em} \overset{(b)}{=} \hspace{0.4em} \underbrace{\vert 1 - \frac{\pi}{\hat{\pi}} \vert}_\text{term 4} + \underbrace{\frac{1}{\hat{\pi}}\mathbbm{E}_{X}[\vert \eta(x) - \hat{\eta}(x) \vert]}_\text{term 5}
\]
\\\\
Where (a) follows from the triangle inequality and noting that the randomness of $\hat{\pi}$ is due to the random draw of the training data; (b) follows due to the fact that $\mathbbm{E}_{X}[\eta(x)] = \pi$. Now, by Lemma B.2 $\frac{1}{\hat{\pi}} \overset{p}{\to} \frac{1}{\pi}$, and thus (term 4 $\overset{p}{\to} 0$), by the Continuous Mapping Theorem; and (term 5 $\overset{p}{\to} 0$) due to Assumption 2 and the fact that $\hat{\pi} \overset{p}{\to} \pi > 0$. It follows by the Squeeze Theorem, that (term 2 $\overset{p}{\to} 0$). Similarly for term 3, we get that: 
\[
0 \leq \mathbbm{E}_{X}[ \vert \frac{\eta(x)\overline{\eta}(x, 1)}{\pi} - \frac{\hat{\eta}(x)\hat{\overline{\eta}}(x, 1)}{\hat{\pi}}  \vert] \overset{(c)}{\leq} \frac{1}{\pi\hat{\pi}} \left\{ \vert \hat{\pi} - \pi \vert \hspace{0.3em}\mathbbm{E}_{X}[\eta(x)\overline{\eta}(x, 1)] + \pi \mathbbm{E}_{X}[\vert \eta(x)\overline{\eta}(x, 1) -\hat{\eta}(x) \hat{\overline{\eta}}(x, 1) \vert]\right\} 
\]
\[
\overset{(d)}{\leq} \underbrace{\vert 1 - \frac{\pi}{\hat{\pi}} \vert}_\text{term 6} + \underbrace{\frac{1}{\hat{\pi}}\mathbbm{E}_{X}[\vert \eta(x)\overline{\eta}(x, 1) - \hat{\eta}(x)\hat{\overline{\eta}}(x, 1) \vert]}_\text{term 7}
\]
\\\\
where (c) follows from first adding and subtracting $\pi\eta(x)\overline{\eta}(x,1)$ and then applying the triangle inequality and (d) follows by noting that $\forall x \in \chi, \vert \eta(x)\overline{\eta}(x,1) \vert \leq 1$. Then, term 6 is the same as term 4, thus, (term 6 $\overset{p}{\to} 0$) and (term 7 $\overset{p}{\to} 0$) due to Lemma B.3 and the fact that $\hat{\pi} \overset{p}{\to} \pi > 0$. Thus, (term 3 $\overset{p}{\to} 0$) by the Squeeze Theorem. Returning to $(2)$, it follows by the Squeeze Theorem that, $\mathbbm{E}_{X}[\vert \gamma(x) - \hat{\gamma}(x) \vert] \overset{p}{\to} 0$, since terms 1, 2 and 3 converge to 0 in probability. \end{proof}

Consistency of the plug-in algorithm (w.r.t. Approximate EO) now follows by an application of Lemma 2 in \cite{narasimhan2014statistical}. For completeness, we restate and prove the consistency statement (Theorem 3.2 in the main text) below in Theorem B.5:
\\\\
{\bf Theorem B.5 \hspace{0.1em} \it Provided Assumptions 1, 2 and 3 hold, the plug-in algorithm is $\Psi$-consistent, i.e., the algorithm yields $\hat{f} = sign \circ \left\{ \hat{\gamma} - c \right\}$, s.t., $\mathfrak{P}_{\mathbbm{P}}^{\Psi}(\hat{f}) \overset{p}{\to} \mathfrak{P}_{\mathbbm{P}}^{\Psi, *}$, i.e., $ regret_{\mathbbm{P}}^{\Psi}(\hat{f}) \overset{p}{\to} 0$ 
}
\begin{proof}
We first show that $TPR_{\mathcal{D}}(\hat{f}) \overset{p}{\to} TPR_{\mathcal{D}}(f^{*})$
\[
TPR_{\mathcal{D}}(\hat{f}) \overset{p}{\to} TPR_{\mathcal{D}}(f^{*}) \iff \mathbbm{P}_{X\vert Y=1}[\hat{f}(x) = 1] \overset{p}{\to} \mathbbm{P}_{X\vert Y=1}[f^{*}(x) = 1]\]
\[\iff \mathbbm{P}_{X\vert Y=1}\left[sign\circ \left\{ \left\{ 1 - \frac{\lambda}{\hat{\pi}}\left(\hat{\overline{\eta}}(x, 1) - \overline{c}\right) \right\}\hat{\eta}(x) - c\right\} = 1\right] \overset{p}{\to} \mathbbm{P}_{X\vert Y=1}\left[sign\circ \left\{ \left\{ 1 - \frac{\lambda}{\pi}\left(\overline{\eta}(x, 1) - \overline{c}\right) \right\}\eta(x) - c\right\} = 1\right]
\]
\[
\iff \mathbbm{P}_{X\vert Y=1}\left[\hat{\gamma}(x) > c\right] \overset{p}{\to} \mathbbm{P}_{X\vert Y=1}\left[\gamma(x) > c\right] \iff \mathbbm{P}_{X\vert Y=1}\left[\hat{\gamma}(x) \leq c\right] \overset{p}{\to} \mathbbm{P}_{X\vert Y=1}\left[\gamma(x) \leq c\right]
\]
\\
Now, 
\[
\mathbbm{E}_{X}\left[\vert \gamma(x) - \hat{\gamma}(x) \vert\right] = \pi \underbrace{\mathbbm{E}_{X \vert Y = 1}\left[\vert \gamma(x) - \hat{\gamma}(x) \vert\right]}_\text{term 1} + (1-\pi) \underbrace{\mathbbm{E}_{X \vert Y = -1}\left[\vert \gamma(x) - \hat{\gamma}(x) \vert\right]}_\text{term 2}
\]
Since $\mathbbm{E}_{X}\left[\vert \gamma(x) - \hat{\gamma}(x) \vert\right] \overset{
p}{\to} 0$ (due to Lemma B.4), terms 1 and 2 also converge to 0 in probability, owing to their non-negativity. Now, for any $\epsilon_{1} > 0$, we have by Markov's Inequality
\[
0 \leq \mathbbm{P}_{X\vert Y=1}\left[\vert \hat{\gamma}(x) - \gamma(x) \vert > \epsilon_{1}\right] \leq \frac{\mathbbm{E}_{X \vert Y = 1}\left[\vert \gamma(x) - \hat{\gamma}(x) \vert\right] }{\epsilon_{1}} \overset{p}{\to} 0 \overset{(a)}{\implies} \underbrace{\mathbbm{P}_{X\vert Y=1}\left[\vert \hat{\gamma}(x) - \gamma(x) \vert > \epsilon_{1}\right]}_\text{term 3} \overset{p}{\to} 0
\]
where (a) follows by the Squeeze Theorem. We recall that the convergence in probability is w.r.t. a random draw of training sample $S$. For a fixed $S$ and a fixed $\epsilon_{2} > 0$, we have 
\[
\mathbbm{P}_{X \vert Y = 1}\left[\hat{\gamma}(x) \leq c\right] = \mathbbm{P}_{X \vert Y = 1}\left[\hat{\gamma}(x) \leq c, \gamma(x) \leq c + \epsilon_{2} \right] + \mathbbm{P}_{X \vert Y = 1}\left[\hat{\gamma}(x) \leq c, \gamma(x) > c + \epsilon_{2}\right] 
\]
\begin{equation}
   \leq \mathbbm{P}_{X \vert Y = 1}\left[\gamma(x) \leq c + \epsilon_{2}\right] + \mathbbm{P}_{X \vert Y = 1}\left[\vert \hat{\gamma}(x) - \gamma(x) \vert \geq \epsilon_{2}\right]  
\end{equation}

and, 
\[
\mathbbm{P}_{X \vert Y = 1}\left[\gamma(x) \leq c - \epsilon_{2}\right] = \mathbbm{P}_{X \vert Y = 1}\left[\hat{\gamma}(x) \leq c, \gamma(x) \leq c - \epsilon_{2}\right] + \mathbbm{P}_{X \vert Y = 1}\left[\hat{\gamma}(x) > c, \gamma(x) \leq c - \epsilon_{2}\right] 
\]
\begin{equation}
\leq \mathbbm{P}_{X \vert Y = 1}\left[\hat{\gamma}(x) \leq c\right] + \mathbbm{P}_{X \vert Y = 1}\left[\vert \hat{\gamma}(x) - \gamma(x)  \vert \geq \epsilon_{2}\right] 
\end{equation}
Consequently, from $(3), (4)$, we get that
\[
\mathbbm{P}_{X \vert Y = 1}\left[\gamma(x) \leq c - \epsilon_{2}\right] - \mathbbm{P}_{X \vert Y = 1}\left[\vert \hat{\gamma}(x) - \gamma(x)  \vert \geq \epsilon_{2}\right] \leq \mathbbm{P}_{X \vert Y = 1}\left[\hat{\gamma}(x) \leq c\right] 
\]
and,
\[
\mathbbm{P}_{X \vert Y = 1}\left[\hat{\gamma}(x) \leq c\right]  \leq \mathbbm{P}_{X \vert Y = 1}\left[\gamma(x) \leq c + \epsilon_{2}\right] + \mathbbm{P}_{X \vert Y = 1}\left[\vert \hat{\gamma}(x) - \gamma(x)  \vert \geq \epsilon_{2}\right]
\]
Subtracting the term $\mathbbm{P}_{X \vert Y = 1}\left[\gamma(x) \leq c\right]$ from both sides in each of the above inequalities and combining the resulting inequalities then gives us
\[
\vert \mathbbm{P}_{X \vert Y = 1}\left[\hat{\gamma}(x) \leq c\right] - \mathbbm{P}_{X \vert Y = 1}\left[\gamma(x) \leq c\right] \vert 
\]
\begin{equation}
\begin{split}
\leq max\{ \underbrace{\mathbbm{P}_{X \vert Y = 1}\left[\vert \hat{\gamma}(x) - \gamma(x)  \vert \geq \epsilon_{2}\right] + \mathbbm{P}_{X \vert Y = 1}\left[\gamma(x) \leq c + \epsilon_{2}\right] -  \mathbbm{P}_{X \vert Y = 1}\left[\gamma(x) \leq c\right]}_\text{term 4}, \\ \underbrace{\mathbbm{P}_{X \vert Y = 1}\left[\vert \hat{\gamma}(x) - \gamma(x)  \vert \geq \epsilon_{2}\right] - \mathbbm{P}_{X \vert Y = 1}\left[\gamma(x) \leq c - \epsilon_{2}\right] +  \mathbbm{P}_{X \vert Y = 1}\left[\gamma(x) \leq c\right]}_\text{term 5} \} 
\end{split}
\end{equation}
Keeping $S$ fixed, we now allow $\epsilon_{2} \to 0$. In particular, by Assumption 1 we have that, $\mathbbm{P}_{X \vert Y = 1}(\gamma(x) \leq k)$ is continuous at $k =c$ and so for the terms inside the above ‘max’, we have that $
lim_{\epsilon_{2} \to 0}$ (term 4)\hspace{0.4em} = \hspace{0.4em}  $lim_{\epsilon_{2} \to 0} \mathbbm{P}_{X \vert Y = 1}\left[\vert \hat{\gamma}(x) - \gamma(x) \vert \geq \epsilon_{2}\right]$ and $lim_{\epsilon_{2} \to 0}$ \hspace{0.4em}  (term 5) \hspace{0.4em}  = $lim_{\epsilon_{2} \to 0} \mathbbm{P}_{X \vert Y = 1}\left[\vert \hat{\gamma}(x) - \gamma(x) \vert \geq \epsilon_{2} \right]$. Thus, for a fixed $S$ the following holds from $(5)$: 
\[
0 \leq \hspace{0.4em} \vert \mathbbm{P}_{X \vert Y = 1}\left[\hat{\gamma}(x) \leq c\right]  - \mathbbm{P}_{X \vert Y = 1}\left[\gamma(x) \leq c\right]\vert \hspace{0.4em} \leq \hspace{0.4em} lim_{\epsilon_{2} \to 0} \mathbbm{P}_{X \vert Y = 1}\left[\vert \hat{\gamma}(x) - \gamma(x) \vert \geq \epsilon_{2} \right]
\]
Now since term 3 converges in probability to 0 for $\epsilon_{1}$, and since $\epsilon_{1}$ was arbitrarily chosen, we we obtain the following convergence in probability over a random draw of training sample $S$ from $\mathbbm{P}^{n}$: 
\[
\vert \mathbbm{P}_{X \vert Y = 1}\left[\hat{\gamma}(x) \leq c\right] - \mathbbm{P}_{X \vert Y = 1}\left[\gamma(x) \leq c\right] \vert \overset{p}{\to} 0
\]
which in turn, implies
\[
\mathbbm{P}_{X \vert Y = 1}\left[\hat{\gamma}(x) \leq c\right] \overset{p}{\to} \mathbbm{P}_{X \vert Y = 1}\left[\gamma(x) \leq c\right]
\]
and we have thus shown, $TPR_{\mathcal{D}}(\hat{f}) \overset{p}{\to} TPR_{\mathcal{D}}(f^{*})$. We can show similarly that $TNR_{\mathcal{D}}(\hat{f}) \overset{p}{\to} TNR_{\mathcal{D}}(f^{*})$ by simply replacing $Y = 1$ with $Y = -1$. We showed above, that term 1 converges to 0 in probability, i.e., $\mathbbm{E}_{X \vert Y = 1}\left[\vert \gamma(x) - \hat{\gamma}(x) \vert\right] \overset{p}{\to} 0$. The same proof progression then allows us to show that $TPR_{\overline{\mathcal{D}}}(\hat{f}) \overset{p}{\to} TPR_{\overline{\mathcal{D}}}(f^{*})$ and $TNR_{\overline{\mathcal{D}}}(\hat{f}) \overset{p}{\to} TNR_{\overline{\mathcal{D}}}(f^{*})$. The result follows by the Continuous Mapping Theorem, since the performance metric $\Psi$ (and thus the performance measure)  is continuous with respect to its arguments. 
\end{proof}

\subsection{Approximate Equality of Opportunity when the sensitive attribute is available at test time}
In this subsection $\overline{\mathcal{D}}$ will be used to refer to $\overline{\mathcal{D}}_{EO}$. Assuming $\pi, \beta >0$, our performance measure is given by: 
\[
\mathfrak{P}_{\mathbbm{P}}^{\Psi}(f) = \Psi\left[TPR_{\mathcal{D}}(f), TNR_{\mathcal{D}}(f), \pi, TPR_{\overline{\mathcal{D}}}(f), TNR_{\overline{\mathcal{D}}}(f), \beta\right] 
= - \left\{CS\left(f; \mathcal{D}, c\right) \hspace{0.1em} - \hspace{0.1em}  \lambda CS\left(f; \overline{\mathcal{D}}, \overline{c}\right)\right\}
\]
\[
= -\{c\left(1-\pi\right)FPR_{\mathcal{D}}(f) + \pi(1-c)FNR_{\mathcal{D}}(f)\} + \lambda\{\overline{c}\left(1-\beta\right)FPR_{\overline{\mathcal{D}}}(f) + \beta(1-\overline{c})FNR_{\overline{\mathcal{D}}}(f)\}
\]
Thus, the performance measure is linear in the arguments of $\Psi$, and therefore continuous in these arguments. The regret of a classifier $f$, w.r.t performance measure $\mathfrak{P}_{\mathbbm{P}}^{\Psi}$ is defined as: 
\[
regret_{\mathbbm{P}}^{\Psi}(f) = \mathfrak{P}_{\mathbbm{P}}^{\Psi, *}- \mathfrak{P}_{\mathbbm{P}}^{\Psi}(f)
\]
where, $\mathfrak{P}_{\mathbbm{P}}^{\Psi, *} = \mathfrak{P}_{\mathbbm{P}}^{\Psi}(f^{*})$. Here $f^{*}$ is the Bayes Optimal Classifier corresponding to the Equal Opportunity case. Denoting $\overline{\eta}$ as $\overline{\eta}_{EO}$ in this subsection, we have that $\forall x \in \chi$ and for any $\overline{y} \in \{-1, 1\}$:

\[f^{*}(x, \overline{y}) = sign\circ s^{*}(x, \overline{y}), \hspace{0.25 em} \] 
\[where, \hspace{0.3em} s^{*}(x, -1) = \left\{ 1 + \frac{\lambda}{\pi}\overline{c}\right \}\eta(x, -1) -c \hspace{0.35em};\hspace{0.35em} 
s^{*}(x, 1) = \left\{ 1 - \frac{\lambda}{\pi}(1-\overline{c})\right \}\eta(x, 1) -c \]

Notice, this time since the sensitive attribute, $\overline{Y}$ is available at test time as well, our estimate for $\eta$ is trained on the feature set as well as the sensitive attribute, with the final decision rule making use of both. We now prove that the plug-in procedure yields an estimator $\hat{f}$ which is $\Psi$-consistent, implying that $ regret_{\mathbbm{P}}^{\Psi}(\hat{f}) \overset{p}{\to} 0$, where $\overset{p}{\to}$ denotes convergence in probability. We denote the estimators of $\eta, \overline{\eta}$ by $\hat{\eta}$ and $\hat{\overline{\eta}}$ respectively. In order to proceed we require the following assumptions, analogous to Assumptions 1 and 2 of the main text to hold: 
\\\\
{\bf Assumption 4} \textit{{\hspace{0.1em} $\mathbbm{P}_{X \vert Y=1}\left(\gamma(x, \overline{y}) \leq k\right), \mathbbm{P}_{X \vert Y=-1}\left(\gamma(x, \overline{y}) \leq k\right), \\ \mathbbm{P}_{X \vert Y=1,  \overline{Y}=1}\left(\gamma(x, 1) \leq k\right)$ and $\mathbbm{P}_{X \vert Y=1, \overline{Y}=-1}\left(\gamma(x, -1) \leq k\right)$ are continuous at $k =c$, where $\gamma(x, \overline{y}) = \eta(x, \overline{y})\left[1 + \frac{\lambda\overline{c}}{\pi} - \frac{\lambda\mathbbm{I}\{\overline{y} = 1\}}{\pi}\right]
$, i.e, $\gamma(x, \overline{y})$ is $s^{*}(x, \overline{y})$ without the constant term $-c$}}
\\\\
{\bf Assumption 5} \textit{{\hspace{0.1em} The class probability estimator $\hat{\eta}$ is $L1$-consistent, i.e., 
\[
\mathbbm{E}_{X, \overline{Y}}\left[\vert \eta(x, \overline{y}) - \hat{\eta}(x, \overline{y}) \vert\right] \overset{p}{\to} 0  
\]
}}
{\it Remark: We now take expectation with respect to the joint distribution of $(X, \overline{Y})$ since the sensitive attribute is also used as a feature in this setting. This assumption is qualitatively no different from Assumption 2 in the main text.}
\\\\

Note that in this setting, we no longer require Assumption 3 from the main text to hold. We denote by $\hat{\gamma}(x, \overline{y}) = \hat{\eta}(x, \overline{y})\left[1 + \frac{\lambda\overline{c}}{\hat{\pi}} - \frac{\lambda\mathbbm{I}\{\overline{y} = 1\}}{\hat{\pi}}\right]
$ the plug-in estimator for $\gamma(x, \overline{y})$. We now state and prove the key lemma that in turn allows us to leverage the proof template of \cite{narasimhan2014statistical} to obtain our consistency result. 
\\\\
{\bf Lemma B.6} \textit{{\hspace{0.1em} Provided Assumption 5 holds, $\hat{\gamma}$ is $L$-1 consistent, i.e.,\hspace{0.25em}$\smash{ \mathbbm{E}_{X, \overline{Y}}\left[\vert \gamma(x, \overline{y}) - \hat{\gamma}(x, \overline{y}) \vert\right] \overset{p}{\to} 0}$}}
\begin{proof}
It follows by the triangle inequality, $\forall x \in \chi$, and for any $y \in \{-1, 1\}$ that,
\[
0 \leq \vert \gamma(x, \overline{y}) - \hat{\gamma}(x, \overline{y}) \leq \vert \eta(x, \overline{y}) - \hat{\eta}(x, \overline{y}) \vert + \vert \lambda\overline{c} \vert \hspace{0.35em} \vert \frac{\eta(x, \overline{y})}{\pi} - \frac{\hat{\eta}(x, \overline{y})}{\hat{\pi}} \vert + \vert\lambda\vert\hspace{0.35em}\vert \left(\frac{\eta(x, \overline{y})}{\pi} - \frac{\hat{\eta}(x, \overline{y})}{\hat{\pi}}\right) \mathbbm{I}\{\overline{y} = 1\} \vert
\]
\[
\leq \vert \eta(x, \overline{y}) - \hat{\eta}(x, \overline{y}) \vert + \left(\vert \lambda\overline{c} \vert + \vert \lambda \vert \right) \hspace{0.35em} \vert \frac{\eta(x, \overline{y})}{\pi} - \frac{\hat{\eta}(x, \overline{y})}{\hat{\pi}} \vert
\]
Now taking expectations on each side, we get that 
\begin{equation}
0 \leq \mathbbm{E}_{X, \overline{Y}}[\vert \gamma(x, \overline{y}) - \hat{\gamma}(x, \overline{y})]  \leq
 \underbrace{\mathbbm{E}_{X, \overline{Y}}[\vert \eta(x, \overline{y}) - \hat{\eta}(x, \overline{y}) \vert}_\text{term 1} + \left(\vert \lambda\overline{c} \vert + \vert \lambda \vert \right)] \hspace{0.35em} \underbrace{\mathbbm{E}_{X, \overline{Y}}[\vert \frac{\eta(x, \overline{y})}{\pi} - \frac{\hat{\eta}(x, \overline{y})}{\hat{\pi}} \vert]}_\text{term 2} 
\end{equation}
By Assumption 5, term 1 converges to 0 in probability. Let's focus on term 2: 
\[
\mathbbm{E}_{X, \overline{Y}}[\vert \frac{\eta(x, \overline{y})}{\pi} - \frac{\hat{\eta}(x, \overline{y})}{\hat{\pi}} \vert] = \frac{1}{\pi\hat{\pi}}\mathbbm{E}_{X, \overline{Y}}[\vert \{\hat{\pi}\eta(x, \overline{y}) - \pi\eta(x, \overline{y})\} + \{\pi\eta(x, \overline{y}) - \pi\hat{\eta}(x, \overline{y})\} \vert]
\]
\[
\overset{(a)}{\leq} \frac{1}{\pi\hat{\pi}}\left\{\vert \hat{\pi} - \pi \vert\mathbbm{E}_{X, \overline{Y}}[\eta(x, \overline{y})] + \pi\mathbbm{E}_{X, \overline{Y}}[\vert \eta(x, \overline{y}) - \hat{\eta}(x, \overline{y}) \vert]\right\}
= \underbrace{\vert 1 - \frac{\pi}{\hat{\pi}}}_\text{term 3} \vert + \underbrace{\frac{\mathbbm{E}_{X, \overline{Y}}[\vert \eta(x, \overline{y}) - \hat{\eta}(x, \overline{y}) \vert]}{\hat{\pi}}}_\text{term 4}
\]
By Lemma B.2 and the Continuous Mapping Theorem, (term 3) $\overset{p}{\to} 0$. Again by Lemma B.2 and noting that $\hat{\pi} \overset{p}{\to} \pi > 0$ coupled with Assumption 5, it follows that (term 4) $\overset{p}{\to} 0$. Now, returning to $(6)$, since terms 1 and 2 converge to 0 in probability, the result follows by the Squeeze Theorem. 
\end{proof}
We can now leverage the proof technique of \cite{narasimhan2014statistical} to obtain our consistency result. We do not present the proof in this case, since it follows a procedure analogous to that detailed in the proof for Theorem B.5
\\\\
{\bf Theorem B.7 \hspace{0.1em} \it Provided Assumptions 4 and 5 hold, the plug-in algorithm is $\Psi$-consistent, i.e., the algorithm yields $\hat{f} = sign \circ \left\{ \hat{\gamma} - c \right\}$, s.t., $\mathfrak{P}_{\mathbbm{P}}^{\Psi}(\hat{f}) \overset{p}{\to} \mathfrak{P}_{\mathbbm{P}}^{\Psi, *}$, i.e., $ regret_{\mathbbm{P}}^{\Psi}(\hat{f}) \overset{p}{\to} 0$ 
}
\subsection{Approximate Demographic Parity when the sensitive attribute is unavailable at test time}
In this subsection $\overline{\mathcal{D}}$ will be used to refer to $\overline{\mathcal{D}}_{DPar}$, where $(X, \overline{Y}) \sim \mathcal{D}_{DPar}$. Assuming $\pi, \beta >0$, our performance measure is given by: 
\[
\mathfrak{P}_{\mathbbm{P}}^{\Psi}(f) = \Psi\left[TPR_{\mathcal{D}}(f), TNR_{\mathcal{D}}(f), \pi, TPR_{\overline{\mathcal{D}}}(f), TNR_{\overline{\mathcal{D}}}(f)\right] 
= - \left\{CS\left(f; \mathcal{D}, c\right) \hspace{0.1em} - \hspace{0.1em}  \lambda CS_{bal}\left(f; \overline{\mathcal{D}}, \overline{c}\right)\right\}
\]
\[
= -\{c\left(1-\pi\right)FPR_{\mathcal{D}}(f) + \pi(1-c)FNR_{\mathcal{D}}(f)\} + \lambda\{\overline{c}FPR_{\overline{\mathcal{D}}}(f) + (1-\overline{c})FNR_{\overline{\mathcal{D}}}(f)\}
\]
Thus, the performance measure is linear in the arguments of $\Psi$, and therefore continuous in these arguments. The regret of a classifier $f$, w.r.t performance measure $\mathfrak{P}_{\mathbbm{P}}^{\Psi}$ is defined as: 
\[
regret_{\mathbbm{P}}^{\Psi}(f) = \mathfrak{P}_{\mathbbm{P}}^{\Psi, *}- \mathfrak{P}_{\mathbbm{P}}^{\Psi}(f)
\]
where, $\mathfrak{P}_{\mathbbm{P}}^{\Psi, *} = \mathfrak{P}_{\mathbbm{P}}^{\Psi}(f^{*})$. Here $f^{*}$ is the Bayes Optimal Classifier corresponding to the Equal Opportunity case. Denoting $\overline{\eta}$ as $\overline{\eta}_{EO}$ in this subsection, we have that $\forall x \in \chi$:

\[f^{*}(x) = sign \circ s^{*}(x)\] 
\[where, \hspace{0.3em} s^{*}(x) = \eta(x) - \{c + \lambda(\overline{\eta}(x) - \overline{c})\} 
\]

We now prove that the plug-in procedure yields an estimator $\hat{f}$ which is $\Psi$-consistent, implying that $ regret_{\mathbbm{P}}^{\Psi}(\hat{f}) \overset{p}{\to} 0$, where $\overset{p}{\to}$ denotes convergence in probability. We denote the estimators of $\eta, \overline{\eta}$ by $\hat{\eta}$ and $\hat{\overline{\eta}}$ respectively. In order to proceed we require the following assumptions, analogous to Assumptions 1 and 2 of the main text to hold: 
\\\\
{\bf Assumption 6} \textit{{\hspace{0.1em} $\mathbbm{P}_{X \vert Y=1}\left(\gamma(x) \leq k\right), \mathbbm{P}_{X \vert Y=-1}\left(\gamma(x) \leq k\right), \\ \mathbbm{P}_{X \vert \overline{Y}=1}\left(\gamma(x) \leq k\right)$ and $\mathbbm{P}_{X \vert \overline{Y}=-1}\left(\gamma(x) \leq k\right)$ are continuous at $k = (c - \lambda\overline{c})$, where $\gamma(x) = \eta(x) - \lambda\overline{\eta}(x)
$, i.e, $\gamma(x)$ is $s^{*}(x)$ without the constant term $-(c - \lambda\overline{c})$}}
\\\\
{\bf Assumption 7} \textit{{\hspace{0.1em} The class probability estimators $\hat{\eta}$ and $\hat{\overline{\eta}}$ are $L1$-consistent, i.e., 
\[
\mathbbm{E}_{X}\left[\vert \eta(x) - \hat{\eta}(x) \vert\right] \overset{p}{\to} 0 \hspace{0.3em} ; \hspace{0.3em} \mathbbm{E}_{X}\left[\vert \overline{\eta}(x) - \hat{\overline{\eta}}(x) \vert\right] \overset{p}{\to} 0
\]
}}
{\it Remark: This assumption is qualitatively no different from Assumption 2 in the main text}
\\\\
Note that in this setting, we no longer require Assumption 3 from the main text to hold. We denote the plug-in estimator for $\gamma$ by $\hat{\gamma}$, i.e., $\hat{\gamma}(x) = \hat{\eta}(x) - \lambda\hat{\overline{\eta}}(x)
$. We now state and prove the key lemma that in turn allows us to leverage the proof template of \cite{narasimhan2014statistical} to obtain our consistency result. 
\\\\
{\bf Lemma B.8} \textit{{\hspace{0.1em} Provided Assumption 7 holds, $\hat{\gamma}$ is $L$-1 consistent, i.e.,\hspace{0.25em}$\smash{ \mathbbm{E}_{X}\left[\vert \gamma(x) - \hat{\gamma}(x) \vert\right] \overset{p}{\to} 0}$}}
\begin{proof}
It follows by the triangle inequality that, $\forall x \in \chi$
\[
0 \leq \vert \gamma(x) - \hat{\gamma}(x) \vert \leq \vert \eta(x) -\hat{\eta}(x) \vert + \vert \lambda \vert \hspace{0.3em} \vert \overline{\eta}(x) - \hat{\overline{\eta}}(x) \vert \hspace{0.3em}
\]
Taking expectations we get,
\[
0 \leq\mathbbm{E}_{X}[\vert \gamma(x) - \hat{\gamma}(x) \vert \leq \underbrace{\mathbbm{E}_{X}[\vert \eta(x) -\hat{\eta}(x) \vert]}_\text{term 1} +  \vert \lambda \vert \hspace{0.3em} \underbrace{\mathbbm{E}_{X}[\vert \overline{\eta}(x) - \hat{\overline{\eta}}(x) \vert]}_\text{term 2} \hspace{0.3em}
\]
Terms 1 and 2 converge to 0 in probability due to Assumption 7. The result follows by the Sequeeze Theorem. 
\end{proof}
We can now leverage the proof technique of \cite{narasimhan2014statistical} to obtain our consistency result. We do not present the proof in this case, since it follows a procedure analogous to that detailed in the proof for Theorem B.5
\\\\
{\bf Theorem B.9 \hspace{0.1em} \it Provided Assumptions 6 and 7 hold, the plug-in algorithm is $\Psi$-consistent, i.e., the algorithm yields $\hat{f} = sign \circ \left\{ \hat{\gamma} - (c -\lambda\overline{c}) \right\}$, s.t., $\mathfrak{P}_{\mathbbm{P}}^{\Psi}(\hat{f}) \overset{p}{\to} \mathfrak{P}_{\mathbbm{P}}^{\Psi, *}$, i.e., $ regret_{\mathbbm{P}}^{\Psi}(\hat{f}) \overset{p}{\to} 0$ 
}
\subsection{Approximate Demographic Parity when the sensitive attribute is available at test time}
In this subsection $\overline{\mathcal{D}}$ will be used to refer to $\overline{\mathcal{D}}_{DPar}$, where $(X, \overline{Y}) \sim \mathcal{D}_{DPar}$. Assuming $\pi, \beta >0$, our performance measure is given by: 
\[
\mathfrak{P}_{\mathbbm{P}}^{\Psi}(f) = \Psi\left[TPR_{\mathcal{D}}(f), TNR_{\mathcal{D}}(f), \pi, TPR_{\overline{\mathcal{D}}}(f), TNR_{\overline{\mathcal{D}}}(f)\right] 
= - \left\{CS\left(f; \mathcal{D}, c\right) \hspace{0.1em} - \hspace{0.1em}  \lambda CS_{bal}\left(f; \overline{\mathcal{D}}, \overline{c}\right)\right\}
\]
\[
= -\{c\left(1-\pi\right)FPR_{\mathcal{D}}(f) + \pi(1-c)FNR_{\mathcal{D}}(f)\} + \lambda\{\overline{c}FPR_{\overline{\mathcal{D}}}(f) + (1-\overline{c})FNR_{\overline{\mathcal{D}}}(f)\}
\]
Thus, the performance measure is linear in the arguments of $\Psi$, and therefore continuous in these arguments. The regret of a classifier $f$, w.r.t performance measure $\mathfrak{P}_{\mathbbm{P}}^{\Psi}$ is defined as: 
\[
regret_{\mathbbm{P}}^{\Psi}(f) = \mathfrak{P}_{\mathbbm{P}}^{\Psi, *}- \mathfrak{P}_{\mathbbm{P}}^{\Psi}(f)
\]
where, $\mathfrak{P}_{\mathbbm{P}}^{\Psi, *} = \mathfrak{P}_{\mathbbm{P}}^{\Psi}(f^{*})$. Here $f^{*}$ is the Bayes Optimal Classifier corresponding to the Equal Opportunity case. Denoting $\overline{\eta}$ as $\overline{\eta}_{EO}$ in this subsection, we have that $\forall x \in \chi$:

\[f^{*}(x, \overline{y}) = H_{\alpha}\circ s^{*}(x, \overline{y}), \hspace{0.25 em} for \hspace{0.25em} any \hspace{0.25em} \alpha \in [0, 1]\] 
\[where, \hspace{0.3em} s^{*}(x, -1) = \eta(x, -1) - c + \lambda\overline{c} \quad and, 
s^{*}(x, 1) = \eta(x, 1) - c + \lambda\overline{c} - \lambda
\]

We now prove that the plug-in procedure yields an estimator $\hat{f}$ which is $\Psi$-consistent, implying that $ regret_{\mathbbm{P}}^{\Psi}(\hat{f}) \overset{p}{\to} 0$, where $\overset{p}{\to}$ denotes convergence in probability. We denote the estimators of $\eta, \overline{\eta}$ by $\hat{\eta}$ and $\hat{\overline{\eta}}$ respectively. In order to proceed we require the following assumptions, analogous to Assumptions 1 and 2 of the main text to hold: 
\\\\
{\bf Assumption 8} \textit{{\hspace{0.1em} $\mathbbm{P}_{X, \overline{Y} \vert Y=1}\left(\gamma(x, \overline{y}) \leq k\right), \mathbbm{P}_{X, \overline{Y} \vert Y=-1}\left(\gamma(x, \overline{y}) \leq k\right), \\ \mathbbm{P}_{X \vert \overline{Y}=1}\left(\gamma(x, 1) \leq k\right)$ and $\mathbbm{P}_{X \vert \overline{Y}=-1}\left(\gamma(x, -1) \leq k\right)$ are continuous at $k = (c - \lambda\overline{c})$, where $\gamma(x, \overline{y}) = \eta(x, \overline{y}) - \lambda\mathbbm{I}\{\overline{y} = 1\}
$, i.e, $\gamma(x, \overline{y})$ is $s^{*}(x, \overline{y})$ without the constant term $-(c - \lambda\overline{c})$}}
\\\\
{\bf Assumption 9} \textit{{\hspace{0.1em} The class probability estimator $\hat{\eta}$ is $L1$-consistent, i.e., 
\[
\mathbbm{E}_{X, \overline{Y}}\left[\vert \eta(x, \overline{y}) - \hat{\eta}(x, \overline{y}) \vert\right] \overset{p}{\to} 0 
\]
}}
{\it Remark: This assumption is qualitatively no different from Assumption 2 in the main text}
\\\\
Note that in this setting, we no longer require Assumption 3 from the main text to hold. We denote the plug-in estimator for $\gamma$ by $\hat{\gamma}$, i.e., $\hat{\gamma}(x, \overline{y}) = \hat{\eta}(x, \overline{y}) -\lambda\mathbbm{I}\{\overline{y} = 1\} $. We now state and prove the key lemma that in turn allows us to leverage the proof template of \cite{narasimhan2014statistical} to obtain our consistency result. 
\\\\
{\bf Lemma B.10} \textit{{\hspace{0.1em} Provided Assumption 8 holds, $\hat{\gamma}$ is $L$-1 consistent, i.e.,\hspace{0.25em}$\smash{ \mathbbm{E}_{X, \overline{Y}}\left[\vert \gamma(x, \overline{y}) - \hat{\gamma}(x, \overline{y}) \vert\right] \overset{p}{\to} 0}$}}
\begin{proof}
The result here is relatively straight-forward. Note that, $\forall x \in \chi$ and for any $\overline{y} \in \{-1, 1\}$
\[
\vert \gamma(x, \overline{y}) - \hat{\gamma}(x, \overline{y}) \vert = \vert \left\{\eta(x, \overline{y}) - \lambda\mathbbm{I}\{\overline{y} = 1\}\right\} - \left\{\hat{\eta}(x, \overline{y}) - \lambda\mathbbm{I}\{\overline{y} = 1\}\right\} \vert = \vert \eta(x, \overline{y}) - \hat{\eta}(x, \overline{y})\vert 
\]
Thus, taking expectations we get that, 
\[
\mathbbm{E}_{X, \overline{Y}}[\vert \gamma(x, \overline{y}) - \hat{\gamma}(x, \overline{y}) \vert] = \mathbbm{E}_{X, \overline{Y}}[\vert \eta(x, \overline{y}) - \hat{\eta}(x, \overline{y}) \vert]
\]
The result follows from Assumption 9.
\end{proof}
We can now leverage the proof technique of \cite{narasimhan2014statistical} to obtain our consistency result. We do not present the proof in this case, since it follows a procedure analogous to that detailed in the proof for Theorem B.5
\\\\
{\bf Theorem B.11 \hspace{0.1em} \it Provided Assumptions 8 and 9 hold, the plug-in algorithm is $\Psi$-consistent, i.e., the algorithm yields $\hat{f} = sign \circ \left\{ \hat{\gamma} - (c -\lambda\overline{c}) \right\}$, s.t., $\mathfrak{P}_{\mathbbm{P}}^{\Psi}(\hat{f}) \overset{p}{\to} \mathfrak{P}_{\mathbbm{P}}^{\Psi, *}$, i.e., $ regret_{\mathbbm{P}}^{\Psi}(\hat{f}) \overset{p}{\to} 0$ 
}
\section{Non-Asymptotic Analysis for the Plug-In Algorithm}
In this section, our objective is to characterise the sample complexity requirements associated with learning a classifier that yields 'small regret', via the plug-in algorithm of \cite{menon2018cost}. The meaning of 'small' will be made clear in the subsections that follow. By 'regret', we refer to the regret associated with the underlying distribution $\mathbbm{P}$ and performance metric, $\Psi$, i.e., $regret_{\mathbbm{P}}^{\Psi}$. As before, our performance measure is simply the negative of the objective of the FAL problem. Thus, in our context, the performance measure of a classifier, is a linear function of its true positive and true negative rates (w.r.t, feature-label distribution, $\mathcal{D}$, as well as feature-sensitive attribute distribution, $\overline{\mathcal{D}}$). This implies that the performance measure is non-decomposable, since it cannot be expressed as a summation/ expectation over individual instances. This is contrary to the case associated with most standard loss functions that feature in the ML literature, such as the 0-1 loss, the hinge loss, the exponential loss and the cross-entropy loss, all of which are decomposable. Therefore, the finite-sample analysis for our performance measure is non-standard. We thus provide a strategy that allows us to precisely relate the sample complexity of this task to the sample complexity associated with learning the regression functions, $\eta$ and $\overline{\eta}$, as well as other distributional quantities which we characterise using simple geometric arguments. \\\\
In the first subsection, we provide basic definitions. This serves as a precursor to the second subsection, wherein we derive the general template for our non-asymptotic analysis. We apply this template to four settings of interest, namely, 1) Approximate Equality of Opportunity when the sensitive attribute is unavailable at test time, 2) Approximate Equality of Opportunity when the sensitive attribute is available at test time, 3) Approximate Demographic Parity when the sensitive attribute is unavailable at test time and 4) Approximate Demographic Parity when the sensitive attribute is available at test-time. Similar to the main text, we will assume in the Approximate Equality of Opportunity setting, that the positive label probability, i.e., $\pi = \mathbbm{P}(Y = 1)$ is known. While we can remove this assumption and modify our analysis to obtain equivalent results, we found that doing so makes the underlying algebra/ geometry much more convoluted, without adding much overall insight. Thus, for simplicity, we will only detail out the analysis assuming $\pi$ is known. In the final subsection, we study the trade-off between fairness and accuracy under a finite sample. 
\subsection{Discussion}
Recall from previous sections, that obtaining our plug-in estimator consisted of estimating two regression functions, namely, $\smash{\eta(x) = \mathbbm{P}(Y = 1 \vert X = x)}$ and $\smash{\overline{\eta}(x) = \mathbbm{P}(\overline{Y} = 1 \vert X = x)}$ (or \hspace{0.3em} $\smash{\mathbbm{P}(\overline{Y}=1 \vert Y= 1, X = x)}$ for the case of Equality of Opportunity); here, $x \in \chi$.  Once we have estimators for these regression functions, i.e., $\smash{\hat{\eta}, \hat{\overline{\eta}}}$, we obtain our decision rule via a simple plug-in rule as laid out in \cite{menon2018cost}. The BOC is a function of the true regression functions. Similarly, the corresponding plug-in classifier, $\hat{f}$ is a function of their estimators, i.e., $\hat{\eta}$ and $\hat{\overline{\eta}}$. Intuitively, the better our approximation for the regression functions, the closer we will be with regards to our goal of achieving small regret. Our analysis makes more precise this relationship between the task of achieving low regret via the plug-in estimator, and the task(s) of learning the regression functions. Given this intuition, we start with the rather broad claim, that the sample complexity of learning a low-regret classifier via the plug-in procedure is directly proportional to the sample complexity of learning the regression functions. To define the sample complexity of learning the regression functions, we will require some notion of 'closeness' between regression functions. Two candidates for doing so are: 

\[
sup_{x\in X} \vert \eta(x) - \hat{\eta}(x) \vert
\]
and, 
\[\mathbbm{P}_{X}(\vert \eta(x) - \hat{\eta}(x) \vert \geq \epsilon)
\]
Analogous notions can be defined for $\overline{\eta}$. Note that $\epsilon$ here is a small positive constant. We will work with the second notion of closeness, as it is less strict. Our analysis applies to the first notion as well though, as it can be seen as a special case of the second notion in the context of our analysis. Notice that the quantity, $\mathbbm{P}_{X}(\vert \eta(x) - \hat{\eta}(x) \vert \geq \epsilon)$ depends on the estimation of $\eta$ via $\hat{\eta}$.The randomness in $\hat{\eta}$ thus stems from the random draw of the training sample, $S$. We now define the sample complexity of learning $\eta$ via class probability estimator $\hat{\eta}$ in Definition C.1: 

{\bf Definition C.1} {\hspace{0.1em} \it The sample complexity of learning $\eta$, is a mapping $m_{\eta}: (0,1)^{3} \rightarrow \mathbbm{N}$, where $m_{\eta}((\epsilon, \delta^{'}), \delta) $ is the minimal (integer) number of training samples required to ensure that, with probability $\geq (1 - \delta)$: $$\mathbbm{P}_{X}(\vert \eta(x) - \hat{\eta}(x) \vert \geq \epsilon) \leq \delta^{'}$$}

For any $\epsilon > 0$, it follows from Markov's Inequality and the $L$-1 consistency assumption on CPE $\hat{\eta} \hspace{0.3em}$(Assumption 2 of the main text) that $0 \leq \mathbbm{P}_{X}(\vert \eta(x) - \hat{\eta}(x) \vert \geq \epsilon) \leq \frac{\mathbb{E}_{X}(\vert\eta(x) - \hat{\eta}(x)\vert)}{\epsilon} \overset{p}{\to} 0$, and thus by the Squeeze Theorem, $\mathbbm{P}_{X}(\vert \eta(x) - \hat{\eta}(x) \vert \geq \epsilon) \overset{p}{\to} 0 \hspace{0.3em} \forall \epsilon > 0$. Thus the above definition is meaningful to work with. The rate of convergence (in probability) would of course, vary from setting to setting. We recall that the convergence in probability is w.r.t. a random draw of the training sample. In Lemma B.1, we also showed that $\hat{\overline{\eta}}(\cdot, 1)$ is $L$-1 consistent. We can thus analogously define and denote the sample complexity of learning $\overline{\eta}$ via $\hat{\overline{\eta}}(\cdot, 1)$ by  $m_{\overline{\eta}}$. Explicitly: \\\\
{\bf Definition C.2} {\hspace{0.1em} \it The sample complexity of learning $\overline{\eta}(\cdot, 1)$, is a mapping $m_{\overline{\eta}}: (0,1)^{3} \rightarrow \mathbbm{N}$, where $m_{\overline{\eta}}((\epsilon, \delta^{'}), \delta) $ is the minimal (integer) number of training samples required to ensure that, with probability $\geq (1 - \delta)$: $$ \mathbbm{P}_{X}(\vert \overline{\eta}(x, 1) - \hat{\overline{\eta}}(x, 1) \vert \geq \epsilon) \leq \delta^{'}$$}

Notice that in Definition C.2, we gave $\overline{\eta}$ two arguments, feature $x$ and $1$. This corresponds to $\overline{\eta} = \overline{\eta}_{EO}$. However, we can simply get rid of the second argument and obtain a similar definition for the sample complexity of $\overline{\eta}_{DPar}$. Even though we continue using two arguments for $\overline{\eta}$ in section C.2, this is not of consequence, the analysis can be replicated for $DPar$ by removing the second argument. 
\subsection{General Template}
Recall that, $\pi = \mathbbm{P}(Y=1)$ and $\beta = \mathbbm{P}(\overline{Y}=1 \vert Y = 1)$. We will assume, as we do throughout this paper that, $\pi, \beta > 0$. Suppose we have access to a training data set, $S = \{x_{i}, y_{i}, \overline{y}_{i}\}_{i=1}^{n}$, comprised of $n$ samples drawn i.i.d. from $\mathbbm{P}$, i.e., the joint distribution over \small{$\left(X, Y, \overline{Y}\right)$}.  Let $\smash{\delta, \delta^{'}, \epsilon \in (0, \frac{1}{2})}$. Let $n \geq max\{m_{\eta}((\epsilon, \frac{\delta^{'}}{2}), \frac{\delta}{8}), m_{\overline{\eta}}((\epsilon, \frac{\delta^{'}}{2}), \frac{\delta}{8})\}$. This implies by Definitions C.1, C.2 and an application of the union bound that with probability $\geq \left(1 - \frac{\delta}{4}\right)$, we can obtain $\hat{\eta}, \hat{\overline{\eta}}$, such that: 
\begin{equation}
    \mathbbm{P}_{X}(\vert \eta(x) - \hat{\eta}(x) \vert \geq \epsilon) \leq \frac{\delta^{'}}{2} \quad and, \quad \mathbbm{P}_{X}(\vert \overline{\eta}(x, 1) - \hat{\overline{\eta}}(x, 1) \vert \geq \epsilon) \leq \frac{\delta^{'}}{2} 
\end{equation}

Subsequently, we may refer to $(7)$ as $(7)$ itself or as the '$(1 - \frac{\delta}{4})$-event' to make the probability associated with the event clear. Now, provided $n \geq max\{m_{\eta}((\epsilon, \frac{\delta^{'}}{2}), \frac{\delta}{8}), m_{\overline{\eta}}((\epsilon, \frac{\delta^{'}}{2}), \frac{\delta}{8})\}$ training samples, what can be said about the plug-in algorithm's performance? More precisely, the plug-in algorithm will yield some classifier, $\hat{f}$, basis estimators, $\hat{\eta}$ and $\hat{\overline{\eta}}$. What can be said about $regret_{\mathbbm{P}}^{\Psi}(\hat{f})$ given the condition on the number of training samples, $n$? Let us analyse this regret. Suppose $t > 0$. 
\[
\mathbbm{P}_{S\sim \mathbbm{P}^{n}}\left[regret_{\mathbbm{P}}^{\Psi} > t\right] =\mathbbm{P}_{S\sim \mathbbm{P}^{n}}\left[\mid P_{\mathbbm{P}}^{\Psi, *}- P_{\mathbbm{P}}^{\Psi}(\hat{f}) \mid > t\right]
\]

\[
=\quad \mathbbm{P}_{S\sim \mathbbm{P}^{n}} \left[\;\middle|\; \{ CS\left(f^{*}; \mathcal{D}, c\right) - \lambda CS(f^{*}; \overline{\mathcal{D}}, \overline{c})\} - \{ CS(\hat{f}; \mathcal{D}, c) - \lambda CS(\hat{f}; \overline{\mathcal{D}}, \overline{c})
 \} \;\middle|\;  > t \right]
\]
\[
\overset{(a)}{\leq} \mathbbm{P}_{S\sim \mathbbm{P}^{n}} \left[ \;\middle|\; CS\left(f^{*}; \mathcal{D}, c\right) - CS\left(\hat{f}; \mathcal{D}, c\right) \;\middle|\; + \mid \lambda \mid \;\middle|\; CS(f^{*}; \mathcal{\overline{D}}, \overline{c}) - CS(\hat{f}; \mathcal{\overline{D}}, \overline{c}) \;\middle|\; > t \right]
\]
\[
\overset{(b)}{\leq} \mathbbm{P}_{S\sim \mathbbm{P}^{n}} \left[ \;\middle|\;  CS\left(f^{*}; \mathcal{D}, c\right) - CS\left(\hat{f}; \mathcal{D}, c\right) \;\middle|\; > \frac{t}{2} \right] + \mathbbm{P}_{S\sim \mathbbm{P}^{n}} \left[\mid \lambda \mid \;\middle|\; CS(f^{*}; \mathcal{\overline{D}}, \overline{c}) - CS(\hat{f}; \mathcal{\overline{D}}, \overline{c}) \;\middle|\; > \frac{t}{2} \right] 
\]
\begin{equation*}
\begin{split}
=  \mathbbm{P}_{S\sim \mathbbm{P}^{n}} \left[ \;\middle|\; \big\{ c(1-\pi)FPR_{\mathcal{D}}(f^{*}) + (1-c)\pi FNR_{\mathcal{D}}(f^{*})\big\} - \big\{c(1-\pi)FPR_{\mathcal{D}}(\hat{f}) + (1-c)\pi FNR_{\mathcal{D}}(\hat{f})\big\} \;\middle|\; > \frac{t}{2} \right] \\
+ \quad \mathbbm{P}_{S\sim \mathbbm{P}^{n}} \left[\mid \lambda \mid \;\middle|\; \big\{\overline{c}(1-\beta)FPR_{\overline{\mathcal{D}}}(f^{*}) + (1-\overline{c})\beta FNR_{\overline{\mathcal{D}}}(f^{*})\big\} - \big\{\overline{c}(1-\beta)FPR_{\overline{\mathcal{D}}}(\hat{f}) + (1-\overline{c})\beta FNR_{\overline{\mathcal{D}}}(\hat{f})\big\} \;\middle|\; > \frac{t}{2} \right] 
\end{split} 
\end{equation*}
\begin{equation}
\begin{split}
\overset{(c)}{\leq} \mathbbm{P}_{S\sim \mathbbm{P}^{n}} \Bigg[ \{c(1-\pi)\} \underbrace{\Big\vert FPR_{\mathcal{D}}(f^{*}) - FPR_{\mathcal{D}}(\hat{f})\Big\vert}_\text{term 1} > \frac{t}{4}\Bigg] +  \mathbbm{P}_{S\sim \mathbbm{P}^{n}} \Bigg[ \{(1-c)\pi\}   \underbrace{\Big\vert FNR_{\mathcal{D}}(f^{*}) - FNR_{\mathcal{D}}(\hat{f})\Big\vert}_\text{term 2} > \frac{t}{4}\Bigg] \\
+ \quad \mathbbm{P}_{S\sim \mathbbm{P}^{n}} \Bigg[ \{\vert \lambda \vert \overline{c}(1-\beta)\} \underbrace{\Big\vert FPR_{\overline{\mathcal{D}}}(f^{*}) - FPR_{\overline{\mathcal{D}}}(\hat{f})\Big\vert}_\text{term 3} > \frac{t}{4}\Bigg] +  \mathbbm{P}_{S\sim \mathbbm{P}^{n}} \Bigg[ \{\vert \lambda \vert(1-\overline{c})\hspace{0.2em}\beta\} \underbrace{\Big\vert FNR_{\overline{\mathcal{D}}}(f^{*}) - FNR_{\overline{\mathcal{D}}}(\hat{f})}_\text{term 4}\Big\vert > \frac{t}{4}\Bigg] 
\end{split} 
\end{equation}
where (a) follows from the triangle inequality and (b) follows from an application of the union bound. (c) again follows from the triangle inequality and an application of the union bound. Let us analyse term 1 in $(8)$. Noting that $TNR = 1 - FPR$, term 1 becomes: 
\[
\Big\vert TNR_{\mathcal{D}}(f^{*}) - TNR_{\mathcal{D}}(\hat{f})\Big\vert = \Bigg\vert \mathbbm{P}\Big[f^{*}(x)= -1 \vert Y = -1 \Big] - \mathbbm{P}\Big[\hat{f}(x) = -1 \vert Y = -1\Big]\Bigg\vert 
\]
\[
= \Big\vert \mathbbm{E}_{X \vert Y = -1}\Big[\mathbbm{I}\{f^{*}(x)= -1\}\Big] - \mathbbm{E}_{X \vert Y = -1}\Big[\mathbbm{I}\{\hat{f}(x) = -1\}\Big]\Big\vert 
\]
\[
\overset{(d)}{=} \Big\vert \mathbbm{E}_{X \vert Y = -1}\Big[\mathbbm{I}\{f^{*}(x)= -1\} - \mathbbm{I}\{\hat{f}(x) = -1\}\Big]\Big\vert
\]
\begin{equation}
\overset{(e)}{\leq}  \mathbbm{E}_{X \vert Y = -1}\Big[\hspace{0.3em}\Big\vert\mathbbm{I}\{f^{*}(x)= -1\} - \mathbbm{I}\{\hat{f}(x) = -1\}\Big\vert\hspace{0.3em}\Big] 
\end{equation}
Here, (d) follows by the linearity of expectations and (e) follows by an application of Jensen's Inequality. Note that the event within the expectation in $(9)$, i.e., $\Big\vert\mathbbm{I}\{f^{*}(x)= -1\} - \mathbbm{I}\{\hat{f}(x) = -1\}\Big\vert$ takes values in $\{0, 1\} \forall x \in \chi$. In fact, it takes the value $1$, exactly when $x: f^{*}(x)\neq \hat{f}(x)$. Therefore, $(9)$ becomes: 
\[
\mathbbm{E}_{X \vert Y = -1}\Big[\mathbbm{I}\left\{f^{*}(x)\neq \hat{f}(x)\right\} \Big] = \mathbbm{P}_{X \vert Y = -1}\Big[f^{*}(x)\neq \hat{f}(x)\Big]
\]
Repeating the analysis analogously for terms 2, 3 and 4 from $(8)$, we get: 
\begin{equation}
\Big\vert TNR_{\mathcal{D}}(f^{*}) - TNR_{\mathcal{D}}(\hat{f})\Big\vert \leq \mathbbm{P}_{X \vert Y = -1}\Big[f^{*}(x)\neq \hat{f}(x)\Big] 
\end{equation}
\begin{equation}
\Big\vert TPR_{\mathcal{D}}(f^{*}) - TPR_{\mathcal{D}}(\hat{f})\Big\vert \leq \mathbbm{P}_{X \vert Y = 1}\Big[f^{*}(x)\neq \hat{f}(x)\Big] 
\end{equation}
\begin{equation}
\Big\vert TNR_{\overline{\mathcal{D}}}(f^{*}) - TNR_{\overline{\mathcal{D}}}(\hat{f})\Big\vert \leq \mathbbm{P}_{\tiny{X \vert (Y = 1, \overline{Y} = -1)}}\Big[f^{*}(x)\neq \hat{f}(x)\Big] 
\end{equation}
\begin{equation}
\Big\vert TPR_{\overline{\mathcal{D}}}(f^{*}) - TPR_{\overline{\mathcal{D}}}(\hat{f})\Big\vert \leq \mathbbm{P}_{\tiny{X \vert (Y = 1, \overline{Y} = 1)}}\Big[f^{*}(x)\neq \hat{f}(x)\Big] 
\end{equation}
Returning to $(8)$, from equations $(10)-(13)$, we get that:
\begin{equation}
\begin{split}
(8) \quad {\leq} \quad \mathbbm{P}_{S\sim \mathbbm{P}^{n}} \Bigg[ \underbrace{\{c(1-\pi)\}\mathbbm{P}_{X \vert Y = -1}\Big[f^{*}(x)\neq \hat{f}(x)\Big]}_\text{term 5} > \frac{t}{4}\Bigg]  \\ + \quad  \mathbbm{P}_{S\sim \mathbbm{P}^{n}} \Bigg[ \underbrace{\{(1-c)\pi\}\mathbbm{P}_{X \vert Y = 1}\Big[f^{*}(x)\neq \hat{f}(x)\Big] }_\text{term 6} > \frac{t}{4}\Bigg] \\
+ \quad \mathbbm{P}_{S\sim \mathbbm{P}^{n}} \Bigg[ \{\underbrace{\vert\lambda\vert\overline{c}(1-\beta)\}{\mathbbm{P}_{\tiny{X \vert (Y = 1, \overline{Y} = -1)}}\Big[f^{*}(x)\neq \hat{f}(x)\Big]}}_\text{term 7} > \frac{t}{4}\Bigg] \\ + \quad \mathbbm{P}_{S\sim \mathbbm{P}^{n}} \Bigg[  \underbrace{\{\vert\lambda\vert(1-\overline{c})\hspace{0.2em}\beta\}\mathbbm{P}_{\tiny{X \vert (Y = 1, \overline{Y} = 1)}}\Big[f^{*}(x)\neq \hat{f}(x)\Big]}_\text{term 8}\Big\vert > \frac{t}{4}\Bigg] 
\end{split} 
\end{equation}
In order to proceed, we will show that, conditional on the $\left(1-\frac{\delta}{4}\right)$-event of $(7)$, the following term can be upper bounded: 
\begin{equation}
\mathbbm{P}_{X}\Big[f^{*}(x)\neq \hat{f}(x)\Big] 
\end{equation}

The form and characterisation of this upper bound, depends on the form of the BOC and its estimator being considered, and this varies across our four settings. We will delve into each setting in sub-subsections that follow, however first we need to carry out some more algebra. Let us for now use $B$ as a placeholder for this upper bound, i.e., conditional upon the $(1-\frac{\delta}{4})$-event, $\mathbbm{P}_{X}\Big[f^{*}(x)\neq \hat{f}(x)\Big] \leq B$. This implies that, conditional upon the $(1-\frac{\delta}{4})$-event of $(7)$:
\[
\pi \mathbbm{P}_{X \vert Y = 1}\Big[f^{*}(x)\neq \hat{f}(x)\Big] + (1-\pi) \mathbbm{P}_{X \vert Y = -1}\Big[f^{*}(x)\neq \hat{f}(x)\Big] \leq B
\]
\begin{equation}
\implies \mathbbm{P}_{X \vert Y = 1}\Big[f^{*}(x)\neq \hat{f}(x)\Big] \leq \frac{B}{\pi} \hspace{0.3em} and, 
\end{equation}
\begin{equation}
\hspace{0.3em} \mathbbm{P}_{X \vert Y = -1}\Big[f^{*}(x)\neq \hat{f}(x)\Big] \leq \frac{B}{1-\pi} 
\end{equation}
Recall, $\mathbbm{P}(\overline{Y} = 1 \vert Y = 1) = \beta$. Similarly, due to $(16)$, we also have that: 
\begin{equation}
\implies \mathbbm{P}_{X \vert (Y = 1, \overline{Y} = 1)}\Big[f^{*}(x)\neq \hat{f}(x)\Big] \leq \frac{B}{\pi \beta} \hspace{0.3em} and, 
\end{equation}
\begin{equation}
\hspace{0.3em} \mathbbm{P}_{X \vert (Y = 1, \overline{Y} = -1)}\Big[f^{*}(x)\neq \hat{f}(x)\Big] \leq \frac{B}{\pi (1-\beta}) 
\end{equation}
Now, let
\[G = max\{\frac{B}{\pi}, \frac{B}{1-\pi}, \frac{B}{\pi \beta}, \frac{B}{\pi (1-\beta)}\} = max\{\frac{B}{1-\pi}, \frac{B}{\pi \beta}, \frac{B}{\pi (1-\beta)}\}  
\]
Further, we set $t$, such that: 
\[
min\{\frac{t}{4c(1-\pi)}, \frac{t}{4(1-c)\pi}, \frac{t}{4\vert\lambda\vert\overline{c}(1-\beta)}, \frac{t}{4\vert\lambda\vert(1-\overline{c})\beta}\} > G 
\]
\[
\iff 
t > 4G \{max\{c(1-\pi), (1-c)\pi, \vert\lambda\vert\overline{c}(1-\beta), \vert\lambda\vert(1-\overline{c})\beta\}\} = Q
\]
Now, by equations $(16)-(19)$ and the assumption on $t$, terms 5 to 8 in $(14)$ are each smaller than $\frac{t}{4}$ with probability at least $(1-\frac{\delta}{4})$, i.e., $\mathbbm{P}_{S \sim \mathbbm{P}^{n}} [term \hspace{0.3em} j > \frac{t}{4}] \leq \frac{\delta}{4}$, for $j \in \{5, 6, 7, 8\}$. Collecting these failure probabilities and plugging them into $(14 )$, we have shown that, $\mathbbm{P}_{S\sim \mathbbm{P}^{n}}\left[regret_{\mathbbm{P}}^{\Psi} > t\right] \leq \delta$. 

Putting everything together, we get that, provided access to $n \geq max\{m_{\eta}((\epsilon, \frac{\delta^{'}}{2}), \frac{\delta}{8}), m_{\overline{\eta}}((\epsilon, \frac{\delta^{'}}{2}), \frac{\delta}{8})\}$ training samples drawn i.i.d. from $\mathbbm{P}$, for $t > Q = 4G \{max\{c(1-\pi), (1-c)\pi, \vert\lambda\vert\overline{c}(1-\beta), \vert\lambda\vert(1-\overline{c})\beta\}\}$, where $G = max\{\frac{B}{1-\pi}, \frac{B}{\pi \beta}, \frac{B}{\pi (1-\beta)}\}$, the plug-in algorithm yields an estimator $\hat{f}$, such that, with probability at least $(1-\delta) \hspace{0.3em}: \hspace{0.3em} regret_{\mathbbm{P}}^{\psi}(\hat{f}) \leq t$. $t$ here, depends on $G$ which is determined via the upper bound on $(15)$, i.e., $B$. We are thus left with characterising $B$. Let us study this upper bound across our settings of interest.

\subsubsection{Approximate Equality of Opportunity when the sensitive attribute is unavailable at test time}
Let $\overline{\eta}$ denote $\overline{\eta}_{EO}$ throughout this sub-subsection. Note that we assume $\pi$ is known. In this setting, $(15)$ takes form: 
\begin{equation}
 \mathbbm{P}_{X}\Big[sign\circ\{(1 + \frac{\lambda\overline{c}}{\pi})\eta(x) - \frac{\lambda}{\pi}\overline{\eta}(x, 1)\eta(x) -c\} \neq sign\circ\{(1 + \frac{\lambda\overline{c}}{\pi})\hat{\eta}(x) - \frac{\lambda}{\pi}\hat{\overline{\eta}}(x, 1)\hat{\eta}(x) -c\}\Big]    
\end{equation}
We now take a geometric view on the event within the aforementioned probability in $(20)$. In particular, note that $H(\lambda, \pi, c, \overline{c}) := \{(1 + \frac{\lambda\overline{c}}{\pi})\eta - \frac{\lambda}{\pi}\overline{\eta}(\cdot, 1)\eta - c = 0$\} defines a hyperbola in the $(\overline{\eta}(\cdot, 1), \eta)$-plane. 
\\\\
Further, since $\eta, \overline{\eta}(\cdot, 1)$ are probabilities, we restrict our view of the plane to the unit square with bottom-left vertex at the origin. Thus, for a given $x \in \chi$, $sign\circ\{(1 + \frac{\lambda\overline{c}}{\pi})\eta(x) - \frac{\lambda}{\pi}\overline{\eta}(x, 1)\eta(x) -c\} \neq sign\circ\{(1 + \frac{\lambda\overline{c}}{\pi})\hat{\eta}(x) - \frac{\lambda}{\pi}\hat{\overline{\eta}}(x, 1)\hat{\eta}(x) -c\} \iff x \in \Big\{x\in \chi: (\overline{\eta}(x, 1), \eta(x))$ \textit{and} $(\hat{\overline{\eta}}(x, 1), \hat{\eta}(x))$ \textit{lie on different sides of} $H(\lambda, \pi, c, \overline{c}) \Big\}$.\\\\ Now, conditioning on the $\smash{(1-\frac{\delta}{4})}$-event, we have by $(7)$ that: 
\[
\mathbbm{P}_{X}(\vert \eta(x) - \hat{\eta}(x) \vert \geq \epsilon) \leq \frac{\delta^{'}}{2} \quad and, \quad \mathbbm{P}_{X}(\vert\overline{\eta}(x, 1) - \hat{\overline{\eta}}(x, 1) \vert \geq \epsilon) \leq \frac{\delta^{'}}{2}
\]

Thus by the union bound, $\mathbbm{P}_{X}(\{\vert \eta(x) - \hat{\eta}(x) \vert \geq \epsilon\} \bigcup \{\vert\overline{\eta}(x, 1) - \hat{\overline{\eta}}(x, 1) \vert \geq \epsilon\}) \leq \delta^{'}$.  We define $X_{bad} := \{x \in \chi: \{\vert \eta(x) - \hat{\eta}(x) \vert \geq \epsilon\} \bigcup \{\vert\overline{\eta}(x, 1) - \hat{\overline{\eta}}(x, 1) \vert \geq \epsilon\}$, and as explained, $\mathbbm{P}_{X}(X_{bad}) \leq \delta^{'}$. This means, $\forall x \in \chi: x \notin X_{bad}$, estimators $(\hat{\overline{\eta}}(x, 1), \hat{\eta}(x))$, will be contained within the square of length $2\epsilon$ centred around $(\overline{\eta}(x, 1), \eta(x))$. \\

Now, there exists a margin like region engulfing the hyperbola $H(\lambda, \pi, c, \overline{c})$ in the $(\overline{\eta}(\cdot, 1), \eta)$-plane, for which $x:$ the $2\epsilon$ square centred at $(\overline{\eta}(x), \eta(x))$ will intersect the hyperbola $H(\lambda, \pi, c, \overline{c})$. This region is given by:
\textit{$X_{M}(\epsilon) := \{x \in \chi:$ the square of length $2\epsilon$ centred at $(\overline{\eta}(x, 1), \eta(x))$ intersects the hyperbola $H(\lambda, \pi, c, \overline{c})$ in the $(\overline{\eta}(\cdot, 1), \eta)$-plane\}}

From hereon we will usually drop the explicit dependence of $X_{M}$ on $\epsilon$ and refer to $X_{M}(\epsilon)$ simply as $X_{M}$. In other words, for $x \in X_{M}$, regardless of whether $x \in X_{bad}$ or not, there is a non-zero probability that, $sign\circ\{(1 + \frac{\lambda\overline{c}}{\pi})\eta(x) - \frac{\lambda}{\pi}\overline{\eta}(x, 1)\eta(x) -c\} \neq sign\circ\{(1 + \frac{\lambda\overline{c}}{\pi})\hat{\eta}(x) - \frac{\lambda}{\pi}\hat{\overline{\eta}}(x, 1)\hat{\eta}(x) -c\}$. Thus, due to $(7)$ and the definitions of $X_{bad}$ and $X_{M}$, if $x \notin \{X_{bad}\bigcup X_{M}\}$, then we can be certain that $sign\circ\{(1 + \frac{\lambda\overline{c}}{\pi})\eta(x) - \frac{\lambda}{\pi}\overline{\eta}(x, 1)\eta(x) -c\} = sign\circ\{(1 + \frac{\lambda\overline{c}}{\pi})\hat{\eta}(x) - \frac{\lambda}{\pi}\hat{\overline{\eta}}(x, 1)\hat{\eta}(x) -c\}$. In the worst case, $\{X_{bad} \bigcap X_{M}\} = \emptyset$. Therefore, we have shown that, conditional on the $(1-\frac{\delta}{4})-$event: 
\begin{equation}
    \mathbbm{P}_{X}\Big[sign\circ\{(1 + \frac{\lambda\overline{c}}{\pi})\eta(x) - \frac{\lambda}{\pi}\overline{\eta}(x, 1)\eta(x) -c\} \neq sign\circ\{(1 + \frac{\lambda\overline{c}}{\pi})\hat{\eta}(x) - \frac{\lambda}{\pi}\hat{\overline{\eta}}(x, 1)\hat{\eta}(x) -c\}\Big]  \leq \mathbbm{P}_{X}(X_{bad}) + \mathbbm{P}_{X}(X_{M}) \leq \delta^{'} + \mathbbm{P}_{X}(X_{M})
\end{equation}
\\
We have thus shown that, with probability at least $(1-\frac{\delta}{4})$, $\delta^{'} + \mathbbm{P}_{X}(X_{M})$ is an upper bound on $\mathbbm{P}_{X}\left[f^{*}(x)\neq \hat{f}(x)\right]$. This is precisely the upper bound $'B'$, we had wanted to characterise for $(15)$. Using (21), we can now state our main result for this setting below: 
\\\\
{\bf Theorem C.3 \hspace{0.1em} \it Let $\smash{\delta, \delta^{'}, \epsilon \in (0, \frac{1}{2})}$. Consider in the $(\overline{\eta}(\cdot, 1), \eta)$-plane, the hyperbola $H(\lambda, \pi, c, \overline{c}) := \{(1 + \frac{\lambda\overline{c}}{\pi})\eta - \frac{\lambda}{\pi}\overline{\eta}(\cdot, 1)\eta - c = 0$\}. Pick any $t > Q = 4G \{max\{c(1-\pi), (1-c)\pi, \vert\lambda\vert\overline{c}(1-\beta), \vert\lambda\vert(1-\overline{c})\beta\}\}$, where $G = max\{\frac{B}{1-\pi}, \frac{B}{\pi \beta}, \frac{B}{\pi (1-\beta)}\}$, and $B = \delta^{'} + \mathbbm{P}_{X}(X_{M})$. Here $X_{M} := \{x \in \chi:$ the square of length $2\epsilon$ centred at $(\overline{\eta}(x, 1), \eta(x))$ intersects the hyperbola $H(\lambda, \pi, c, \overline{c})$ in the $(\overline{\eta}(\cdot, 1), \eta)$-plane\}. Provided access to $n \geq max\{m_{\eta}((\epsilon, \frac{\delta^{'}}{2}), \frac{\delta}{8}), m_{\overline{\eta}}((\epsilon, \frac{\delta^{'}}{2}), \frac{\delta}{8})\}$ training samples drawn i.i.d. from $\mathbbm{P}$, the plug-in algorithm yields an estimator $\hat{f}$, such that, with probability at least $(1-\delta) \hspace{0.3em}: \hspace{0.3em} regret_{\mathbbm{P}}^{\psi}(\hat{f}) \leq t$
}
\\\\
\textbf{\underline{Subset Argument}}: For a given $(\delta, \delta^{'})$, in order to make the regret small, we would need to set $t$ to a smaller quantity. This can be done by making $Q$ smaller, which requires making $G$, and thus $B$ smaller. Making $B$ smaller in turn, entails making $P_{X}(X_M)$ small. By transitivity, our regret can be made arbitrarily small for a given choice of the pair, $(\delta, \delta^{'})$, if we can ensure $P_{X}(X_M)$ can be made sufficiently small. \\\\
Note that $P_{X}(X_{M})$ is a distributional quantity that can be hard to access or estimate without further knowledge of the underlying distribution. What we have shown in essence is that, for a given choice of the pair $(\delta, \delta^{'})$, provided a sufficiently large training sample, the regret can be made as small as desired. Let us formalise this further:

Choose a $(\delta, \delta^{'})$ and suppose for integers $n_{1}, n_{2}$, where $\smash{n_{1} < n_{2}}$, that,
\begin{equation*}
    \smash{n_1 \geq max\{m_{\eta}((\epsilon_{1}, \frac{\delta^{'}}{2}), \frac{\delta}{8}), m_{\overline{\eta}}((\epsilon_{1}, \frac{\delta^{'}}{2}), \frac{\delta}{8})\}}
\end{equation*} 
and,
\begin{equation*}
    \smash{n_2 \geq max\{m_{\eta}((\epsilon_{2}, \frac{\delta^{'}}{2}), \frac{\delta}{8}), m_{\overline{\eta}}((\epsilon_{2}, \frac{\delta^{'}}{2}), \frac{\delta}{8})\} \hspace{0.3em} where \hspace{0.3em} \epsilon_{1} > \epsilon_{2}}
\end{equation*}. Let
\begin{equation*}
X_{M}(\epsilon_{1}) := \{x \in \chi: \textit{\hspace{0.1em}the\hspace{0.1em} square\hspace{0.1em} of\hspace{0.1em} length\hspace{0.1em}} \hspace{0.1em}2\epsilon_{1} \textit{\hspace{0.1em} centred \hspace{0.1em} at \hspace{0.1em}} (\overline{\eta}(x, 1), \eta(x))\textit{ \hspace{0.1em} intersects \hspace{0.1em} the \hspace{0.1em} hyperbola \hspace{0.1em}} H(\lambda, \pi, c, \overline{c})\}
\end{equation*}
\begin{equation*}
    X_{M}(\epsilon_{2}) := \{x \in \chi: \textit{\hspace{0.1em}the\hspace{0.1em} square\hspace{0.1em} of\hspace{0.1em} length\hspace{0.1em}} \hspace{0.1em}2\epsilon_{2} \textit{\hspace{0.1em} centred \hspace{0.1em} at \hspace{0.1em}} (\overline{\eta}(x, 1), \eta(x))\textit{ \hspace{0.1em} intersects \hspace{0.1em} the \hspace{0.1em} hyperbola \hspace{0.1em}} H(\lambda, \pi, c, \overline{c})\}
\end{equation*}

Clearly, $X_{M}(\epsilon_{1}) \supset X_{M}(\epsilon_{2})$, since a larger square centred around any point will intersect the hyperbola if a smaller square centred around the same point does, though not necessarily vice-versa. And therefore, $\mathbbm{P}_{X}(X_{M}(\epsilon_{2})) < \mathbbm{P}_{X}(X_{M}(\epsilon_{1}))$. Therefore, by Theorem C.3, the regret can be made to decrease arbitrarily, provided a sufficient increase in the number of training samples. Since, this argument relied upon showing that $X_{M}(\epsilon_{1}) \supset X_{M}(\epsilon_{2})$, we will refer to this argument as the \textbf{Subset Argument}. The precise rate of decay depends on 1) the sample complexities associated with learning the regression functions and 2) the rate at which the probability measure ($\mathbbm{P}_{X}$) around the hyperbola $H(\lambda, \pi, c, \overline{c})$ decays (in the $(\overline{\eta}(\cdot, 1), \eta)$-plane) upon shrinking the region of consideration around it (i.e., the region akin to the 'Projected $X_{M}$' region in \textit{Figure 3}). A visual display of our analysis applied to two numerical instances can be found in \textit{Figure 3}.

\textit{Remark regarding Figure 3: The vertical asymptote for the hyperbola $H(\lambda, \pi, c, \overline{c})$ occurs at the x-coordinate $\{\overline{c} + \frac{\pi}{\lambda}\}$. Note that, depending on the location of the vertical asymptote of the hyperbola under consideration, there are two possible cases that arise. In case 1, the asymptote occurs outside of the 0-1 plane (see: \textit{Figure 3, Top panel}), when $\overline{c} + \frac{\pi}{\lambda} < 0$ or, when $\overline{c} + \frac{\pi}{\lambda} > 1$. In this case, the hyperbola in the 0-1 plane, i.e., the $(\overline{\eta}(\cdot, 1), \eta)$-plane, splits the plane into two. The region above the hyperbola, for which $sign\circ\{(1 + \frac{\lambda\overline{c}}{\pi})\eta - \frac{\lambda}{\pi}\overline{\eta}(\cdot, 1)\eta -c\}$ will be positive, and the region below the hyperbola, for which $sign\circ\{(1 + \frac{\lambda\overline{c}}{\pi})\eta - \frac{\lambda}{\pi}\overline{\eta}(\cdot, 1)\eta -c\}$ will be negative. \\\\
In case 2, the asymptote occurs inside of the 0-1 plane (see: \textit{Figure 3, Bottom panel}), when $0 \leq \overline{c} + \frac{\pi}{\lambda} \leq 1$. In this case, the hyperbola in the 0-1 plane, i.e., the $(\overline{\eta}(\cdot, 1), \eta)$-plane, splits the region into three, as the hyperbola features as two disjointed segments in the plane. Points falling in the region in between the two segments will have a sign opposite to the points falling in the regions on either side of the segments. We remark that the red regions (see: \textit{Figure 3, Right panel}) corresponding to the 'projected $X_{bad}$ region' are dummy regions simulated solely for demonstration purposes. In general, the area and location of these regions in the $(\overline{\eta}(\cdot, 1), \eta)$-plane is not known. Indeed such regions need not even be wholly contained within a local region in the plane, they may be the union of smaller disjointed regions scattered all across the plane.}

\begin{figure}[tbh]
\includegraphics[width = 17cm]{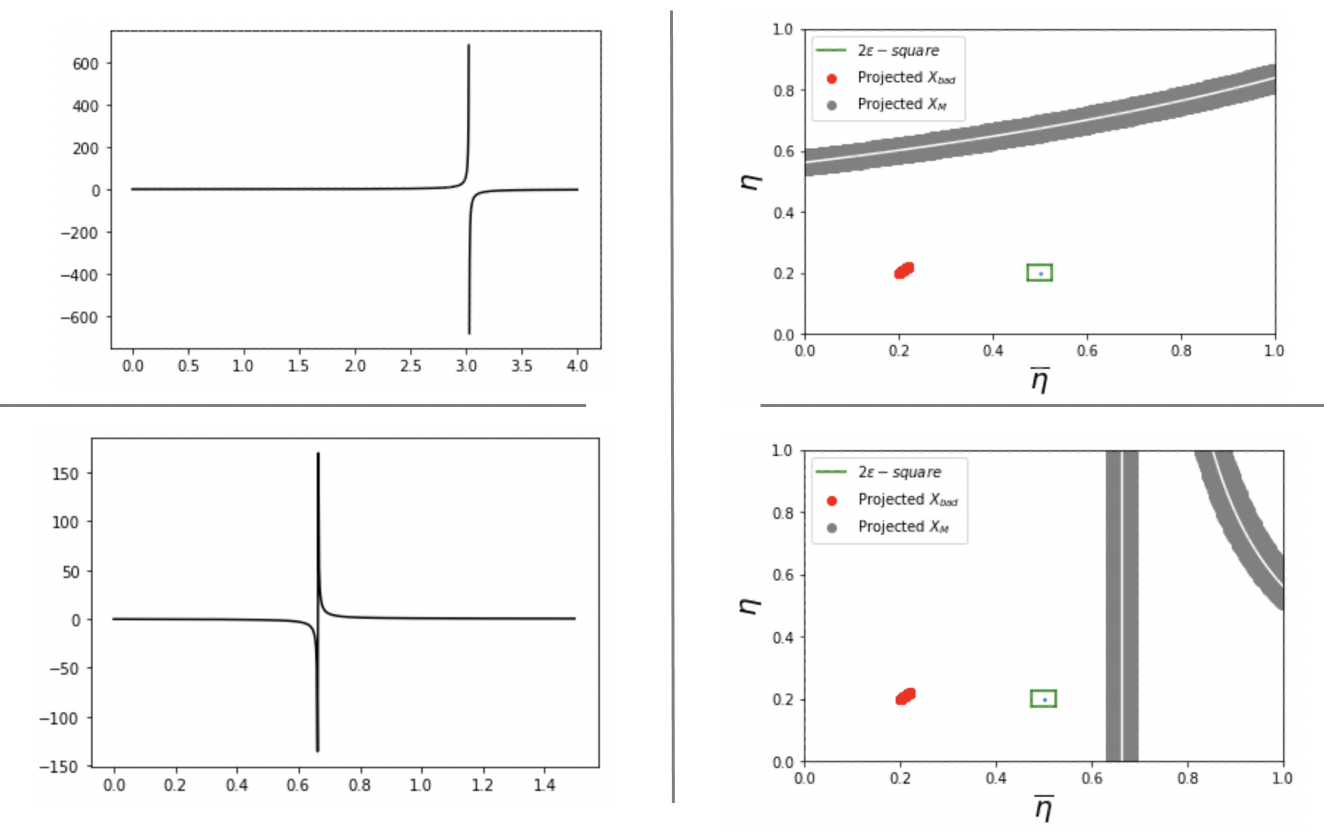}
\caption{\textit{{\textbf{Top Left}: The hyperbola $H(\lambda = 0.4, \pi = 0.85, c = 0.8, \overline{c} = 0.9)$ in the unrestricted plane. The vertical asymptote for this hyperbola occurs outside of the 0-1 plane when the x-coordinate = 3.02. \textbf{Top Right}: The hyperbola $H(\lambda = 0.4, \pi = 0.85, c = 0.8, \overline{c} = 0.9)$ in the restricted $(\overline{\eta}(\cdot, 1), \eta)$-plane, i.e., the 0-1 plane. The white stripe within the gray region is the hyperbola, $H$. The gray region is the set of coordinates of $X_{M}$ in the $(\overline{\eta}(\cdot, 1), \eta)$-plane. We will refer to this region as the 'projected $X_{M}$' region. The red region corresponds to the set of coordinates of $X_{bad}$ in the $(\overline{\eta}(\cdot, 1), \eta)$-plane. We will refer to this region as the 'projected $X_{bad}$' region. We will refer to this region as the 'projected $X_{bad}$' region. With probability $\geq (1-\frac{\delta}{4})$: For any point $x \in \chi:$ \hspace{0.3em} corresponding projected coordinates in the $(\overline{\eta}(\cdot, 1), \eta)$-plane, i.e., $(\overline{\eta}(x, 1), \eta(x))$ lie outside of \{projected $X_{M} \bigcup$ projected $X_{bad}$\}, we can be certain that $sign\circ\{(1 + \frac{\lambda\overline{c}}{\pi})\eta(x) - \frac{\lambda}{\pi}\overline{\eta}(x, 1)\eta(x) -c\} = sign\circ\{(1 + \frac{\lambda\overline{c}}{\pi})\hat{\eta}(x) - \frac{\lambda}{\pi}\hat{\overline{\eta}}(x, 1)\hat{\eta}(x) -c\}$ - an example of such a point is the point centred within the green square of length 2$\epsilon$. We set $\epsilon = 0.025$ in this simulation. \textbf{Bottom Left}: The hyperbola $H(\lambda = -3.6, \pi = 0.85, c = 0.8, \overline{c} = 0.9)$ in the unrestricted plane. The vertical asymptote for this hyperbola occurs outside of the 0-1 plane when the x-coordinate = 0.6638. \textbf{Bottom Right}: The hyperbola $H(\lambda = -3.6, \pi = 0.85, c = 0.8, \overline{c} = 0.9)$ in the restricted $(\overline{\eta}(\cdot, 1), \eta)$-plane, i.e., the 0-1 plane. The white stripe within the gray region is the hyperbola, $H$. The gray region is the set of coordinates of $X_{M}$ in the $(\overline{\eta}(\cdot, 1), \eta)$-plane. We will refer to this region as the 'projected $X_{M}$' region. The red region corresponds to the set of coordinates of $X_{bad}$ in the $(\overline{\eta}(\cdot, 1), \eta)$-plane. We will refer to this region as the 'projected $X_{bad}$' region. With probability $\geq (1-\frac{\delta}{4})$: For any point $x \in \chi:$ \hspace{0.3em} corresponding projected coordinates in the $(\overline{\eta}(\cdot, 1), \eta)$-plane, i.e., $(\overline{\eta}(x, 1), \eta(x))$ lie outside of \{projected $X_{M} \bigcup$ projected $X_{bad}$\}, we can be certain that $sign\circ\{(1 + \frac{\lambda\overline{c}}{\pi})\eta(x) - \frac{\lambda}{\pi}\overline{\eta}(x, 1)\eta(x) -c\} = sign\circ\{(1 + \frac{\lambda\overline{c}}{\pi})\hat{\eta}(x) - \frac{\lambda}{\pi}\hat{\overline{\eta}}(x, 1)\hat{\eta}(x) -c\}$ - an example of such a point is the point centred within the green square of length 2$\epsilon$. We set $\epsilon = 0.025$ in this simulation.}}}
\centering
\end{figure}
\clearpage
\subsubsection{Approximate Equality of Opportunity when the sensitive attribute is available at test time}
 Since the sensitive feature is available at test-time, we tweak our sample complexity definition (i.e., Definition C.1) appropriately.  
 
 {\bf Definition C.4} {\hspace{0.1em} \it The sample complexity of learning $\eta$, is a mapping $m_{\eta}: (0,1)^{3} \rightarrow \mathbbm{N}$, where $m_{\eta}((\epsilon, \delta^{'}), \delta) $ is the minimal (integer) number of training samples required to ensure that, with probability $\geq (1 - \delta)$:
\[
\mathbbm{P}_{X, \overline{Y}}(\vert \eta(x, \overline{y}) - \hat{\eta}(x, \overline{y}) \vert \geq \epsilon) \leq \delta^{'}
\]
}
Before proceeding, observe that: 
\[
\mathbbm{P}_{X, \overline{Y}}(\vert \eta(x, \overline{y}) - \hat{\eta}(x, \overline{y}) \vert \geq \epsilon) = \mathbbm{E}_{X, \overline{Y}}\left[\mathbbm{I}\{\vert \eta(x, \overline{y}) - \hat{\eta}(x, \overline{y}) \vert \geq \epsilon\}\right]
\]
\[
\mathbbm{E}_{X, \overline{Y}}\left[\mathbbm{I}\{\vert \hat{\eta}(x, \overline{y}) - \eta(x, \overline{y}) \vert \geq \epsilon\}\right] = \bigintsss_{x \in \chi}\left\{ \sum_{y \in \{-1, 1\}} \mathbbm{I}\{\vert \hat{\eta}(x, \overline{y}) - \eta(x, \overline{y})  \vert \geq \epsilon\} f_{X, \overline{Y}}(x, \overline{y}) \right\} dx
\]
\[
= \overline{\pi} \bigintsss_{x \in \chi} \mathbbm{I}\{\vert \hat{\eta}(x, 1) - \eta(x, 1)  \vert \geq \epsilon\} f_{X \vert \overline{Y} = 1}(x) dx + \hspace{0.4em} (1- \overline{\pi}) \bigintsss_{x \in \chi} \mathbbm{I}\{\vert \hat{\eta}(x, -1) - \overline{\eta}(x, -1)  \vert \geq \epsilon\} f_{X \vert \overline{Y} = -1}(x) dx
\]
\begin{equation}
= \overline{\pi}\hspace{0.3em}\mathbbm{P}_{X \vert \overline{Y} = 1}\left[\vert \hat{\eta}(x, 1) - \eta(x, 1) \vert \geq \epsilon \right] + (1-\overline{\pi})\hspace{0.3em}\mathbbm{P}_{X \vert \overline{Y} = -1}\left[\vert \hat{\eta}(x, -1) - \eta(x, -1) \vert \geq \epsilon\right]  
\end{equation}
For convenience, we also tweak our set-up to fit this setting.  Let $\smash{\delta, \delta^{''}, \epsilon \in (0, \frac{1}{2})}$. Suppose now that $n \geq m_{\eta}((\epsilon, \delta^{''}), \frac{\delta}{4})$. Thus by Definition C.4, we have that with probability $\geq \left(1 - \frac{\delta}{4}\right)$, we can obtain $\hat{\eta}$ such that: 
\[
\mathbbm{P}_{X, \overline{Y}}(\vert \eta(x, \overline{y}) - \hat{\eta}(x, \overline{y}) \vert \geq \epsilon) \leq \delta^{''}
\]
\[
\overset{by (22)}{\implies} \mathbbm{P}_{X \vert \overline{Y} = 1}\left[\vert \hat{\eta}(x, 1) - \eta(x, 1) \vert \geq \epsilon\right] \leq \frac{\delta^{''}}{\overline{\pi}} \hspace{0.4em} and \hspace{0.4em} \mathbbm{P}_{X \vert \overline{Y} = -1}\left[\vert \hat{\eta}(x, -1) - \eta(x, -1) \vert \geq \epsilon \right] \leq \frac{\delta^{''}}{1-\overline{\pi}}
\]
Set $\delta^{'} := \max\{\frac{\delta^{''}}{\overline{\pi}}, \frac{\delta^{''}}{1-\overline{\pi}}\}$. Then we have that, with probability at least $(1-\frac{\delta}{4})$: 
\begin{equation}
    \mathbbm{P}_{X \vert \overline{Y} = 1}\left[\vert \hat{\eta}(x, 1) - \eta(x, 1) \vert \geq \epsilon \right] \leq \delta^{'} 
\end{equation}
and, 
\begin{equation}
   \mathbbm{P}_{X \vert \overline{Y} = -1}\left[\vert \hat{\eta}(x, -1) - \eta(x, -1) \vert \geq \epsilon \right] \leq \delta^{'}  
\end{equation}
The template is directly applicable here, except the probabilities and expectations involved are now taken over feature, sensitive attribute pairs, rather that just the feature alone. For completeness we summarise the key steps from the template applied to this setting as well.  Now, for $t>0$, repeating steps from the template derivation presented in C.2, we would get in this setting that: 
\[
\mathbbm{P}_{S\sim \mathbbm{P}^{n}}\left[regret_{\mathbbm{P}}^{\Psi} > t\right] =\mathbbm{P}_{S\sim \mathbbm{P}^{n}}\left[\mid P_{\mathbbm{P}}^{\Psi, *}- P_{\mathbbm{P}}^{\Psi}(\hat{f}) \mid > t\right]
\]
\begin{equation}
\begin{split}
{\leq} \quad \mathbbm{P}_{S\sim \mathbbm{P}^{n}} \Bigg[ \underbrace{\{c(1-\pi)\}\mathbbm{P}_{X, \overline{Y} \vert Y = -1}\Big[f^{*}(x, \overline{y}) \neq \hat{f}(x, \overline{y})\Big]}_\text{term 1} > \frac{t}{4}\Bigg]  \\ + \quad  \mathbbm{P}_{S\sim \mathbbm{P}^{n}} \Bigg[\underbrace{\{(1-c)\pi\} \mathbbm{P}_{X, \overline{Y} \vert Y = 1}\Big[f^{*}(x, \overline{y}) \neq \hat{f}(x, \overline{y})\Big] }_\text{term 2} > \frac{t}{4}\Bigg] \\
+ \quad \mathbbm{P}_{S\sim \mathbbm{P}^{n}} \Bigg[ \underbrace{\{\vert\lambda\vert\overline{c}(1-\beta)\}{\mathbbm{P}_{\tiny{X \vert (Y = 1, \overline{Y} = -1)}}\Big[f^{*}(x, -1) \neq \hat{f}(x, -1)\Big]}}_\text{term 3} > \frac{t}{4}\Bigg] \\ + \quad \mathbbm{P}_{S\sim \mathbbm{P}^{n}} \Bigg[ \underbrace{ \{\vert\lambda\vert(1-\overline{c})\hspace{0.2em}\beta\}\mathbbm{P}_{\tiny{X \vert (Y = 1, \overline{Y} = 1)}}\Big[f^{*}(x, 1) \neq \hat{f}(x, 1)\Big]}_\text{term 4}\Big\vert > \frac{t}{4}\Bigg]
\end{split} 
\end{equation}
As before, we can proceed if we can upper bound $\mathbbm{P}_{X, \overline{Y}}\left[ f^{*}(x, \overline{y}) \neq f^{*}(x, \overline{y}) \right]$ conditional on the $(1-\frac{\delta}{4})$-event. For now, let $B$ be a placeholder for this upper bound. This implies that, conditional upon the $(1-\frac{\delta}{4})$-event:

\begin{equation}
\mathbbm{P}_{X, \overline{Y} \vert Y = 1}\Big[f^{*}(x, \overline{y}) \neq \hat{f}(x, \overline{y})\Big] \leq \frac{B}{\pi} \hspace{0.3em} and, 
\end{equation}
\begin{equation}
\hspace{0.3em} \mathbbm{P}_{X, \overline{Y} \vert Y = -1}\Big[f^{*}(x, \overline{y}) \neq \hat{f}(x, \overline{y})\Big] \leq \frac{B}{1-\pi} 
\end{equation}
Similarly, due to $(26)$, we also have that: 
\begin{equation}
\implies \mathbbm{P}_{X \vert (Y = 1, \overline{Y} = 1)}\Big[f^{*}(x, 1) \neq \hat{f}(x, 1)\Big] \leq \frac{B}{\pi \beta} \hspace{0.3em} and, 
\end{equation}
\begin{equation}
\hspace{0.3em} \mathbbm{P}_{X \vert (Y = 1, \overline{Y} = -1)}\Big[f^{*}(x, -1) \neq \hat{f}(x, -1)\Big] \leq \frac{B}{\pi (1-\beta)} 
\end{equation}
Now, let
\[G = max\{\frac{B}{\pi}, \frac{B}{1-\pi}, \frac{B}{\pi \beta}, \frac{B}{\pi (1-\beta)}\} = max\{\frac{B}{1-\pi}, \frac{B}{\pi \beta}, \frac{B}{\pi (1-\beta)}\}
\]
Further, we set $t$, such that: 
\[
min\{\frac{t}{4c(1-\pi)}, \frac{t}{4(1-c)\pi}, \frac{t}{4\vert\lambda\vert\overline{c}(1-\beta)}, \frac{t}{4\vert\lambda\vert(1-\overline{c})\beta}\} > G 
\]
\[
\iff 
t > 4G \{max\{c(1-\pi), (1-c)\pi, \vert\lambda\vert\overline{c}(1-\beta), \vert\lambda\vert(1-\overline{c})\beta\}\} = Q
\]

Now, by equations $(26)-(29)$ and due to the assumption on $t$, terms 1 to 4 in $(25)$ are each smaller than $\frac{t}{4}$ with probability at least $(1-\frac{\delta}{4})$, i.e., $\mathbbm{P}_{S \sim \mathbbm{P}^{n}} [term \hspace{0.3em} j > \frac{t}{4}] \leq \frac{\delta}{4}$, for $j \in \{1, 2, 3, 4\}$. Collecting these failure probabilities and plugging them into $(25)$, we have shown that, $\mathbbm{P}_{S\sim \mathbbm{P}^{n}}\left[regret_{\mathbbm{P}}^{\Psi} > t\right] \leq \delta$. 

Putting everything together, we get that, provided access to $n \geq m_{\eta}((\epsilon,\delta^{''}, \frac{\delta}{4})$ training samples drawn i.i.d. from $\mathbbm{P}$, for $t > Q = 4G \{max\{c(1-\pi), (1-c)\pi, \vert\lambda\vert\overline{c}(1-\beta), \vert\lambda\vert(1-\overline{c})\beta\}\}$, where $G = max\{\frac{B}{1-\pi}, \frac{B}{\pi \beta}, \frac{B}{\pi (1-\beta)}\}$, the plug-in algorithm yields an estimator $\hat{f}$, such that, with probability at least $(1-\delta) \hspace{0.3em}: \hspace{0.3em} regret_{\mathbbm{P}}^{\psi}(\hat{f}) \leq t$. $t$ here, depends on $G$ which is determined via the upper bound on $\mathbbm{P}_{X, \overline{Y}}\left[ f^{*}(x, \overline{y}) \neq f^{*}(x, \overline{y}) \right]$, i.e., $B$. We are thus left with characterising $B$. We now characterise this upper bound, just like we did for the case when $\overline{Y}$ was unavailable at test time. 
\\\\
The underlying geometry simplifies significantly in this case. Note that 
\[
\mathbbm{P}_{X, \overline{Y}}\left[ f^{*}(x, \overline{y}) \neq f^{*}(x, \overline{y}) \right]
\]
\begin{equation}
 \begin{split}
 = \quad (1-\overline{\pi})\underbrace{\mathbbm{P}_{X \vert \overline{Y} = -1 }\left[ sign\circ\{(1 + \frac{\lambda\overline{c}}{\pi})\eta(x, -1) - c\} \neq  sign\circ\{(1 + \frac{\lambda\overline{c}}{\pi})\hat{\eta}(x, -1) - c\} \right]}_\textit{term 5} \\ + \quad \overline{\pi}\underbrace{\mathbbm{P}_{X \vert \overline{Y} = 1 }\left[sign\circ\{(1 + \frac{\lambda(\overline{c}-1)}{\pi})\eta(x, 1) - c\} \neq sign\circ\{(1 + \frac{\lambda(\overline{c}-1)}{\pi})\hat{\eta}(x, 1) - c\} \right]}_\textit{term 6}
 \end{split}
\end{equation}
Let us focus on term 5 in $(30)$ first. Notice that, $sign\circ\{(1 + \frac{\lambda\overline{c}}{\pi})\eta(x, -1) - c\} \neq  sign\circ\{(1 + \frac{\lambda\overline{c}}{\pi})\hat{\eta}(x, -1) - c\}$ whenever $x: \eta(x, -1)$ and $\hat{\eta}(x, -1)$ lie on different sides of the threshold $T_{-1}(\lambda, \pi, c, \overline{c}) = \left(\frac{c}{1+\frac{\lambda\overline{c}}{\pi}}\right)$ on the $\eta(\cdot, -1)$-axis. 
\\\\
Conditioning on the $\smash{(1-\frac{\delta}{4})}$-event, we have by $(24)$ that:
\[\mathbbm{P}_{X \vert \overline{Y} = -1}\left[\vert \hat{\eta}(x, -1) - \eta(x, -1) \vert \geq \epsilon \right] \leq \delta^{'}\]

Denote $X_{-bad} := \{x \in \chi: \vert \hat{\eta}(x, -1) - \eta(x, -1) \vert \geq \epsilon\}$. Also, let $X_{-M}(\epsilon) := \{x \in \chi: \vert \eta(x, -1) -T_{-1}(\lambda, \pi, c, \overline{c}) \vert \leq \epsilon\}$. We drop the explicit dependence of $X_{-M}$ on $\epsilon$, and denote $X_{-M}(\epsilon)$ by $X_{-M}$ from hereon.  Thus, if $x \in X_{-M}$, then regardless of whether $x \in X_{-bad}$ or not, there is a non-zero probability, that $\hat{\eta}(x)$ lies the other side of the threshold relative to $\eta(x)$. Thus, due to $(24)$, and the definitions of $X_{-M}$ and $X_{-bad}$, we can be certain that $\eta(x, -1)$ and $\hat{\eta}(x, -1)$ lie on the same side of threshold $T_{-1}(\lambda, \pi, c, \overline{c})$ whenever $x \notin \{X_{-bad} \bigcup X_{-M}$\}. Let $\mathbbm{P}_{X\vert \overline{Y} = -1}(X_{-M}) = Q_{-M}$ and from $(24)$, we know that, $\mathbbm{P}_{X \vert Y = -1}(X_{-bad}) \leq \delta^{'}$. \\\\

Similarly, for term 6 in $(30)$, $sign\circ\{(1 + \frac{\lambda(\overline{c}-1)}{\pi})\eta(x, 1) - c\} \neq sign\circ\{(1 + \frac{\lambda(\overline{c}-1)}{\pi})\hat{\eta}(x, 1) - c\}$ whenever $x: \eta(x, 1)$ and $\hat{\eta}(x, 1)$ lie on different sides of the threshold $T_{1}(\lambda, \pi, c, \overline{c}) = \left(\frac{c}{1+\frac{\lambda(\overline{c}-1)}{\pi}}\right)$ on the $\eta(\cdot, 1)$-axis. 
\\\\
Conditioning on the $\smash{(1-\frac{\delta}{4})}$-event, we have by $(23)$ that:
\[\smash{\mathbbm{P}_{X \vert \overline{Y} = -1}\left[\vert \hat{\eta}(x, 1) - \eta(x, -1) \vert \geq \epsilon\right] \leq \delta^{'}}\]

Denote $\smash{X_{bad} := \{x \in \chi: \vert \hat{\eta}(x, 1) - \eta(x, 1) \vert \geq \epsilon\}}$. Also, let $\smash{X_{M}(\epsilon) := \{x \in \chi: \vert \eta(x, 1) - T_{1}(\lambda, \pi, c, \overline{c}) \vert \leq \epsilon\}}$. We drop the explicit dependence of $X_{M}$ on $\epsilon$, and denote $X_{M}(\epsilon)$ by $X_{M}$ from hereon. Thus, if $x \in X_{M}$, then regardless of whether $x \in X_{bad}$ or not, there is a non-zero probability, that $\hat{\eta}(x, 1)$ lies the other side of the threshold relative to $\eta(x, 1)$. Thus, due to $(23)$, and the definitions of $X_{M}$ and $X_{bad}$, we can be certain that $\eta(x, 1)$ and $\hat{\eta}(x, 1)$ lie on the same side of threshold $T_{1}(\lambda, \pi, c, \overline{c})$ whenever $x \notin \{X_{bad} \bigcup X_{M}$\}. Let $\mathbbm{P}_{X\vert \overline{Y} = 1}(X_{M}) = Q_{M}$, and from $(23)$, we know that, $\mathbbm{P}_{X \vert Y = 1}(X_{bad}) \leq \delta^{'}$.
\\\\
In the worst case, $\left\{X_{-M}\bigcap X_{-bad} \vert Y = -1\right\} = \left\{X_{M} \bigcap X_{bad} \vert Y = 1\right\} = \emptyset$. We have thus shown, that with probability at least $(1-\frac{\delta}{4})$ that term 5 in $(30)$ $\leq Q_{-M} + \delta^{'}$ and term 6 in $(30)$ $\leq Q_{-M} + \delta^{'}$. Setting $P_{M} = max\{Q_{-M}, Q_{M}\}$, we then get that: 
\[
\mathbbm{P}_{X, \overline{Y}}\left[ f^{*}(x, \overline{y}) \neq f^{*}(x, \overline{y}) \right] \leq (1-\overline{\pi})(P_{M} + \delta^{'}) + \overline{\pi}(P_{M} + \delta^{'}) = P_{M} + \delta^{'}
\]
We have thus shown that, with probability at least $(1-\frac{\delta}{4})$, $\delta^{'} + P_{M}$ is an upper bound on $\mathbbm{P}_{X}\left[f^{*}(x)\neq \hat{f}(x)\right]$. This is precisely the upper bound $'B'$, we had wanted to characterise. We can now state our main result for this setting below: 
\\\\
{\bf Theorem C.5 \hspace{0.1em} \it Let $\smash{\delta, \delta^{''}, \epsilon \in (0, \frac{1}{2})}$. Set \hspace{0.3em} $\smash{\delta^{'} = \max\{\frac{\delta^{''}}{\overline{\pi}}, \frac{\delta^{''}}{1-\overline{\pi}}\}}$. Let $T_{-1}(\lambda, \pi, c, \overline{c}) = \left(\frac{c}{1+\frac{\lambda\overline{c}}{\pi}}\right)$ and $T_{1}(\lambda, \pi, c, \overline{c}) = \left(\frac{c}{1+\frac{\lambda(\overline{c}-1)}{\pi}}\right)$ denote thresholds on the $\eta(\cdot, -1)$ and $\eta(\cdot, 1)$ axes respectively. Pick any $t > Q = 4G \{max\{c(1-\pi), (1-c)\pi, \vert\lambda\vert\overline{c}(1-\beta), \vert\lambda\vert(1-\overline{c})\beta\}\}$, where $G = max\{\frac{B}{1-\pi}, \frac{B}{\pi \beta}, \frac{B}{\pi (1-\beta)}\}$, and $B = \delta^{'} + P_{M}$. Here $P_{M} :=  max\{Q_{-M}, Q_{M}\}$, where $Q_{-M} = \mathbbm{P}_{X\vert \overline{Y} = -1}(X_{-M})$ and $Q_{M} = \mathbbm{P}_{X\vert \overline{Y} = -1}(X_{M})$, and $X_{-M} := \{x \in \chi: \vert \eta(x, -1) -T_{-1}(\lambda, \pi, c, \overline{c}) \vert \leq \epsilon\}$, while $X_{M} := \{x \in \chi: \vert \eta(x, 1) -T_{1}(\lambda, \pi, c, \overline{c}) \vert \leq \epsilon\}$ . Provided access to $n \geq m_{\eta}((\epsilon, \delta^{''}), \frac{\delta}{4})$ training samples drawn i.i.d. from $\mathbbm{P}$, the plug-in algorithm yields an estimator $\hat{f}$, such that, with probability at least $(1-\delta) \hspace{0.3em}: \hspace{0.3em} regret_{\mathbbm{P}}^{\psi}(\hat{f}) \leq t$
}
\\\\
Thus, by Theorem C.5, using an argument analogous to the \textbf{Subset Argument} presented following Theorem C.3, we have that the regret can be made to decrease arbitrarily, provided a sufficient increase in the number of training samples. The precise rate of decay depends on 1) the sample complexities associated with learning the regression function and 2) the rate at which the probability measure ($\mathbbm{P}_{X}$) around the thresholds $T_{-1}(\lambda, \pi, c, \overline{c}) = \left(\frac{c}{1+\frac{\lambda\overline{c}}{\pi}}\right)$ and $T_{1}(\lambda, \pi, c, \overline{c}) = \left(\frac{c}{1+\frac{\lambda(\overline{c}-1)}{\pi}}\right)$ decay (in the $\eta(\cdot, -1), \eta(\cdot, 1)$-axes respectively) upon shrinking the regions of consideration around them.

\subsubsection{Approximate Demographic Parity when the sensitive attribute is unavailable at test time}
Throughout this section, let $\overline{\eta}$ denote $\overline{\eta}_{DPar}$. Let $\smash{\delta, \delta^{'}, \epsilon \in (0, \frac{1}{2})}$.
\\\\
 Following precisely the template laid out in C.2, we get that, provided access to $\smash{n \geq max\{m_{\eta}((\epsilon, \frac{\delta^{'}}{2}), \frac{\delta}{8}), m_{\overline{\eta}}((\epsilon, \frac{\delta^{'}}{2}), \frac{\delta}{8})\}}$ training samples drawn i.i.d. from $\mathbbm{P}$,for $t > Q = 4G \{max\{c(1-\pi), (1-c)\pi, \vert\lambda\vert\overline{c}(1-\beta), \vert\lambda\vert(1-\overline{c})\beta\}\}$,\\
where $G = max\{\frac{B}{1-\pi}, \frac{B}{\pi \beta}, \frac{B}{\pi (1-\beta)}\}$, the plug-in algorithm yields an estimator $\hat{f}$, such that, with probability at least $(1-\delta) \hspace{0.3em}: \hspace{0.3em} regret_{\mathbbm{P}}^{\psi}(\hat{f}) \leq t$. $t$ here, depends on $G$ which is determined via the upper bound (which holds with probability $\geq (1-\frac{\delta}{4})$) on $\mathbbm{P}_{X}\left[\hat{f}(x)\neq f^{*}(x)\right]$, i.e., $B$. We are thus left with characterising $B$. Let us study this upper bound in this setting.
\\\\
Note that, 
\begin{equation}
    \mathbbm{P}_{X}\left[\hat{f}(x)\neq f^{*}(x)\right] = \mathbbm{P}_{X}\left[sign\circ \{\hat{\eta}(x) - \lambda(\hat{\overline{\eta}}(x) - \overline{c}) - c\}\neq sign\circ\{\eta(x) - \lambda(\overline{\eta}(x) - \overline{c}) - c\}\right]
\end{equation}
In this case, the underlying geometry is linear. Once again, consider the $(\overline{\eta}, \eta)$-plane restricted to the $(0,1)$-plane. For any $x \in \chi$, a sign disagreement occurs, i.e., $sign\circ \{\hat{\eta}(x) - \lambda(\hat{\overline{\eta}}(x) - \overline{c}) - c\}\neq sign\circ\{\eta(x) - \lambda(\overline{\eta}(x) - \overline{c}) - c\}$, whenever $x: (\hat{\overline{\eta}}(x), \hat{\eta}(x))$ and  $(\overline{\eta}(x), \eta(x))$ lie on different sides of the line, $L(\lambda, \pi, c, \overline{c}) = \{\eta -\lambda\overline{\eta} + \lambda\overline{c} - c\}$ in the $(\overline{\eta}, \eta)$-plane. Conditioning on the $(1-\frac{\delta}{4})$-event, analogous to $(7)$, we have that: 
\begin{equation}
    \mathbbm{P}_{X}(\vert \eta(x) - \hat{\eta}(x) \vert \geq \epsilon) \leq \frac{\delta^{'}}{2} \quad and, \quad \mathbbm{P}_{X}(\vert \overline{\eta}(x) - \hat{\overline{\eta}}(x) \vert \geq \epsilon) \leq \frac{\delta^{'}}{2}
\end{equation}
Thus by the union bound, $\mathbbm{P}_{X}(\{\vert \eta(x) - \hat{\eta}(x) \vert \geq \epsilon\} \bigcup \{\vert\overline{\eta}(x) - \hat{\overline{\eta}}(x) \vert \geq \epsilon\}) \leq \delta^{'}$.  We define $X_{bad} := \{x \in \chi: \{\vert \eta(x) - \hat{\eta}(x) \vert \geq \epsilon\} \bigcup \{\vert\overline{\eta}(x) - \hat{\overline{\eta}}(x) \vert \geq \epsilon\}$, and as explained, $\mathbbm{P}_{X}(X_{bad}) \leq \delta^{'}$. This means, $\forall x \in \chi: x \notin X_{bad}$, estimators $(\hat{\overline{\eta}}(x), \hat{\eta}(x))$, will be contained within the square of length $2\epsilon$ centred around $(\overline{\eta}(x), \eta(x))$. \\

Now, there exists a margin like region around the line, $L(\lambda, \pi, c, \overline{c})$, in the $(\overline{\eta}, \eta)$-plane, for which $x:$ the $2\epsilon$ square centred at $(\overline{\eta}(x), \eta(x))$ will intersect the line $L(\lambda, \pi, c, \overline{c})$. This region is given by:
\[X_{M}(\epsilon) := \{x \in \chi: \textit{\hspace{0.1em}the\hspace{0.1em} square\hspace{0.1em} of\hspace{0.1em} length\hspace{0.1em}} \hspace{0.1em}2\epsilon \textit{\hspace{0.1em} centred \hspace{0.1em} at \hspace{0.1em}} (\overline{\eta}(x), \eta(x))\textit{ \hspace{0.1em} intersects \hspace{0.1em} the \hspace{0.1em} line \hspace{0.1em}} L(\lambda, \pi, c, \overline{c})\}\]
We drop the explicit dependence of $X_{M}$ on $\epsilon$, and denote $X_{M}(\epsilon)$ by $X_{M}$ from hereon. Thus, if $x \in X_{M}$, then there is a non-zero probability that $sign\circ \{\hat{\eta}(x) - \lambda(\hat{\overline{\eta}}(x) - \overline{c}) - c\}\neq sign\circ\{\eta(x) - \lambda(\overline{\eta}(x) - \overline{c}) - c\}$, i.e., $(\hat{\overline{\eta}}(x), \hat{\eta}(x))$ and  $(\overline{\eta}(x), \eta(x))$ lie on different sides of the line, $L(\lambda, \pi, c, \overline{c})$. Due to $(32)$ and the definitions of $X_{M}$ and $X_{bad}$, we can be certain that, for any $x\in \chi: x \notin \{X_{bad} \bigcup X_{M}$\}, $sign\circ \{\hat{\eta}(x) - \lambda(\hat{\overline{\eta}}(x) - \overline{c}) - c\} = sign\circ\{\eta(x) - \lambda(\overline{\eta}(x) - \overline{c}) - c\}$. In the worst case, $\left\{X_{M} \bigcap X_{bad}\right\} = \emptyset$. Therefore, we have shown that, with probability at least $(1-\frac{\delta}{4})$: 
\begin{equation}
    \mathbbm{P}_{X}\left[sign\circ \{\hat{\eta}(x) - \lambda(\hat{\overline{\eta}}(x) - \overline{c}) - c\}\neq sign\circ\{\eta(x) - \lambda(\overline{\eta}(x) - \overline{c}) - c\}\right] \leq \delta^{'} + \mathbbm{P}_{X}(X_{M})
\end{equation}

We have thus shown that, with probability at least $(1-\frac{\delta}{4})$, $\delta^{'} + \mathbbm{P}_{X}(X_{M})$ is an upper bound on $\mathbbm{P}_{X}\left[\hat{f}(x)\neq f^{*}(x)\right]$. This is precisely the upper bound, $'B'$ that we had wanted to characterise. Using (33), we state our key result for this setting:

{\bf Theorem C.6 \hspace{0.1em} \it Let $\smash{\delta, \delta^{'}, \epsilon \in (0, \frac{1}{2})}$. Consider in the $(\overline{\eta}, \eta)$-plane, the line,  $L(\lambda, \pi, c, \overline{c}) = \{\eta -\lambda\overline{\eta} + \lambda\overline{c} - c\}$. Pick any $t > Q = 4G \{max\{c(1-\pi), (1-c)\pi, \vert\lambda\vert\overline{c}(1-\beta), \vert\lambda\vert(1-\overline{c})\beta\}\}$, where $G = max\{ \frac{B}{1-\pi}, \frac{B}{\pi \beta}, \frac{B}{\pi (1-\beta)}\}$, and $B = \delta^{'} + \mathbbm{P}_{X}(X_{M})$. Here $X_{M} := \{x \in \chi:$ the square of length $2\epsilon$ centred at $(\overline{\eta}(x), \eta(x))$ intersects the line $L(\lambda, \pi, c, \overline{c})$ in the $(\overline{\eta}, \eta)$-plane\}. Provided access to $n \geq max\{m_{\eta}((\epsilon, \frac{\delta^{'}}{2}), \frac{\delta}{8}), m_{\overline{\eta}}((\epsilon, \frac{\delta^{'}}{2}), \frac{\delta}{8})\}$ training samples drawn i.i.d. from $\mathbbm{P}$, the plug-in algorithm yields an estimator $\hat{f}$, such that, with probability at least $(1-\delta) \hspace{0.3em}: \hspace{0.3em} regret_{\mathbbm{P}}^{\psi}(\hat{f}) \leq t$
}
\\\\
Thus, by Theorem C.6, using an argument analogous to the \textbf{Subset Argument} presented following Theorem C.3, the regret can be made to decrease arbitrarily, provided a sufficient increase in the number of training samples. The precise rate of decay depends on 1) the sample complexities associated with learning the regression function and 2) the rate at which the probability measure ($\mathbbm{P}_{X}$) around the line, $L(\lambda, \pi, c, \overline{c}) = \{\eta -\lambda\overline{\eta} + \lambda\overline{c} - c\}$ (in the $(\overline{\eta}, \eta)$-plane) decays upon shrinking the regions of consideration around them (i.e., the region akin to the 'Projected $X_{M}$' region in \textit{Figure 4}). A visual display of our analysis applied to a numerical example, can be found in \textit{Figure 4}. 
\\\\
\begin{figure}[tbh]
\centering

\includegraphics[width = 10cm]{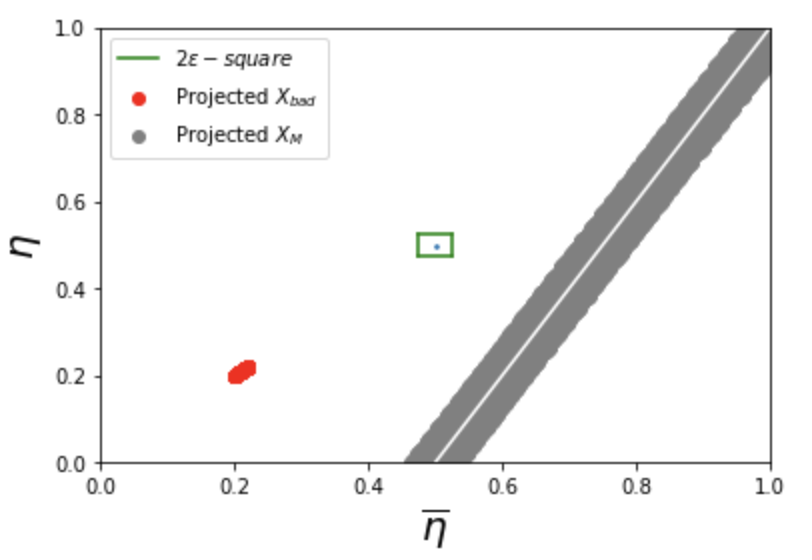}
\caption{\textit{The line $L(\lambda = 2.0, c = 0.8, \overline{c} = 0.9)$ in the restricted $(\overline{\eta}, \eta)$-plane, i.e., the 0-1 plane. The white stripe within the gray region is the line, $L$. The gray region is the set of coordinates of $X_{M}$ in the $(\overline{\eta}, \eta)$-plane. We will refer to this region as the 'projected $X_{M}$' region. The red region corresponds to the set of coordinates of $X_{bad}$ in the $(\overline{\eta}, \eta)$-plane. We will refer to this region as the 'projected $X_{bad}$' region. For any point $x \in \chi:$ \hspace{0.3em} corresponding projected coordinates in the $(\overline{\eta}, \eta)$-plane, i.e., $(\overline{\eta}(x, 1), \eta(x))$ lie outside of \{projected $X_{M} \bigcup$ projected $X_{bad}$\}, we can be certain that $sign\circ \{\hat{\eta}(x) - \lambda(\hat{\overline{\eta}}(x) - \overline{c}) - c\} = sign\circ\{\eta(x) - \lambda(\overline{\eta}(x) - \overline{c}) - c\}$ - an example of such a point is the point centred within the green square of length 2$\epsilon$. We set $\epsilon = 0.025$ in this simulation.}}
\end{figure}

\textit{Remark regarding Figure 4: The red region corresponding to the 'projected $X_{bad}$ region' is a dummy region simulated solely for demonstration purposes. In general, the area and location of these regions in the $(\overline{\eta}, \eta)$-plane is not known. Indeed such regions need not even be wholly contained within a local region in the plane, they may be the union of smaller disjointed regions scattered all across the plane.}

\subsubsection{Approximate Demographic Parity when the sensitive attribute is available at test time}
The tweaked template presented in sub-subsection C.2.2 is directly applicable in this setting. We reintroduce the basic set-up. Let $\smash{\delta, \delta^{''}, \epsilon \in (0, \frac{1}{2})}$. Suppose access to $n \geq m_{\eta}((\epsilon, \delta^{''}), \frac{\delta}{4})$ training samples drawn i.i.d. from $\mathbbm{P}$. Thus by Definition C.4, we have that with probability $\geq \left(1 - \frac{\delta}{4}\right)$, we can obtain $\hat{\eta}$ such that: 
\[
\mathbbm{P}_{X, \overline{Y}}(\vert \eta(x, \overline{y}) - \hat{\eta}(x, \overline{y}) \vert \geq \epsilon) \leq \frac{\delta^{''}}{2}
\]

Set $\delta^{'} := \max\{\frac{\delta^{''}}{\overline{\pi}}, \frac{\delta^{''}}{1-\overline{\pi}}\}$. Then we have that, with probability at least $(1-\frac{\delta}{4})$: 
\begin{equation}
    \mathbbm{P}_{X \vert \overline{Y} = 1}\left[\vert \hat{\eta}(x, 1) - \eta(x, 1) \vert \geq \epsilon \right] \leq \delta^{'} 
\end{equation}
and, 
\begin{equation}
   \mathbbm{P}_{X \vert \overline{Y} = -1}\left[\vert \hat{\eta}(x, -1) - \eta(x, -1) \vert \geq \epsilon \right] \leq \delta^{'}  
\end{equation}
Following the strategy identical to that laid out in C.2.2, we get that: Provided access to $n \geq m_{\eta}((\epsilon, \delta^{''}), \frac{\delta}{4})$ training samples drawn i.i.d. from $\mathbbm{P}$, for $t > Q = 4G \{max\{c(1-\pi), (1-c)\pi, \vert\lambda\vert\overline{c}(1-\beta), \vert\lambda\vert(1-\overline{c})\beta\}\}$, where $G = max\{\frac{B}{1-\pi}, \frac{B}{\pi \beta}, \frac{B}{\pi (1-\beta)}\}$, the plug-in algorithm yields an estimator $\hat{f}$, such that, with probability at least $(1-\delta) \hspace{0.3em}: \hspace{0.3em} regret_{\mathbbm{P}}^{\psi}(\hat{f}) \leq t$. $t$ here, depends on $G$ which is determined via the upper bound (which holds with probability $\geq (1-\frac{\delta}{4})$) on $\mathbbm{P}_{X, \overline{Y}}\left[ f^{*}(x, \overline{y}) \neq f^{*}(x, \overline{y}) \right]$, i.e., $B$. We are thus left with characterising $B$. We now characterise the upper bound, $B$, in this setting. The underlying geometry simplifies significantly in this case. Note that \\\\
\[
\mathbbm{P}_{X, \overline{Y}}\left[ f^{*}(x, \overline{y}) \neq f^{*}(x, \overline{y}) \right]
\]
\begin{equation}
 \begin{split}
 = \quad (1-\overline{\pi})\underbrace{\mathbbm{P}_{X \vert \overline{Y} = -1 }\left[ sign\circ\{\eta(x, -1) - c + \lambda\overline{c} \} \neq  sign\circ\{(\hat{\eta}(x, -1) - c + \lambda\overline{c} \} \right]}_\textit{term 1} \\ + \quad \overline{\pi}\underbrace{\mathbbm{P}_{X \vert \overline{Y} = 1 }\left[sign\circ\{\eta(x, 1) - c + \lambda\overline{c} - \lambda\} \neq sign\circ\{\hat{\eta}(x, 1) - c + \lambda\overline{c} - \lambda\} \right]}_\textit{term 2}
 \end{split}
\end{equation}
Let us focus on term 1 in $(36)$ first. Notice that, $sign\circ\{\eta(x, -1) - c + \lambda\overline{c} \} \neq  sign\circ\{(\hat{\eta}(x, -1) - c + \lambda\overline{c}\}$ whenever $x: \eta(x, -1)$ and $\hat{\eta}(x, -1)$ lie on different sides of the threshold $T_{-1}(\lambda, \pi, c, \overline{c}) = \left(c - \lambda\overline{c}\right)$ on the $\eta(\cdot, -1)$-axis. 
\\\\
Conditioning on the $(1-\frac{\delta}{4})$-event, we have by $(35)$ that:

\[\mathbbm{P}_{X \vert \overline{Y} = -1}\left[\vert \hat{\eta}(x, -1) - \eta(x, -1) \vert \geq \epsilon\right] \leq \delta^{'}\]

Denote $X_{-bad} := \{x \in \chi: \vert \hat{\eta}(x, -1) - \eta(x, -1) \vert \geq \epsilon\}$. Also, let $X_{-M}(\epsilon) := \{x \in \chi: \vert \eta(x, -1) -T_{-1}(\lambda, \pi, c, \overline{c}) \vert \leq \epsilon\}$. We drop the explicit dependence of $X_{-M}$ on $\epsilon$, and denote $X_{-M}(\epsilon)$ by $X_{-M}$ from hereon.  Thus, if $x \in X_{-M}$, then regardless of whether $x \in X_{-bad}$ or not, there is a non-zero probability, that $\hat{\eta}(x)$ lies the other side of the threshold relative to $\eta(x)$. Thus, due to $(35)$ and the definitions of $X_{-M}$ and $X_{-bad}$, we can be certain that $\eta(x, -1)$ and $\hat{\eta}(x, -1)$ lie on the same side of threshold $T_{-1}(\lambda, \pi, c, \overline{c})$ whenever $x \notin \{X_{-bad} \bigcup X_{-M}\}$. Let $\mathbbm{P}_{X\vert \overline{Y} = -1}(X_{-M}) = Q_{-M}$, and from $(35)$, we know that, $\mathbbm{P}_{X \vert Y = -1}(X_{-bad}) \leq \delta^{'}$. 
\\\\
Similarly, for term 2 in $(36)$, $sign\circ\{\eta(x, 1) - c + \lambda\overline{c} - \lambda\} \neq sign\circ\{\hat{\eta}(x, 1) - c + \lambda\overline{c} - \lambda\}$ whenever $x: \eta(x, 1)$ and $\hat{\eta}(x, 1)$ lie on different sides of the threshold $T_{1}(\lambda, \pi, c, \overline{c}) = \left(c + \lambda - \lambda\overline{c}\right)$ on the $\eta(\cdot, 1)$-axis. 
\\\\
Conditioning on the $(1-\frac{\delta}{4})$-event we have by $(34)$ that:

\[\mathbbm{P}_{X \vert \overline{Y} = -1}\left[\vert \hat{\eta}(x, 1) - \eta(x, -1) \vert \geq \epsilon\right] \leq \delta^{'}
\]

Denote $X_{bad} := \{x \in \chi: \vert \hat{\eta}(x, 1) - \eta(x, 1) \vert \geq \epsilon\}$. Also, let $X_{M}(\epsilon) := \{x \in \chi: \vert \eta(x, 1) - T_{1}(\lambda, \pi, c, \overline{c}) \vert \leq \epsilon\}$. We drop the explicit dependence of $X_{M}$ on $\epsilon$, and denote $X_{M}(\epsilon)$ by $X_{M}$ from hereon. Thus, if $x \in X_{M}$, then regardless of whether $x \in X_{bad}$ or not, there is a non-zero probability, that $\hat{\eta}(x, 1)$ lies the other side of the threshold relative to $\eta(x, 1)$. Thus, due to $(34)$, and the definitions of $X_{M}$ and $X_{bad}$, we can be certain that $\eta(x, 1)$ and $\hat{\eta}(x, 1)$ lie on the same side of threshold $T_{1}(\lambda, \pi, c, \overline{c})$ whenever $x \notin \{X_{bad} \bigcup X_{M}\}$. Let $\mathbbm{P}_{X\vert \overline{Y} = 1}(X_{M}) = Q_{M}$, and from $(34)$, we know that, $\mathbbm{P}_{X \vert Y = 1}(X_{bad}) \leq \delta^{'}$.
\\\\
In the worst case, $\left\{X_{-M}\bigcap X_{-bad} \vert Y = -1\right\} = \left\{X_{M} \bigcap X_{bad} \vert Y = 1\right\} = \emptyset$. We have thus shown, that with probability at least $(1-\frac{\delta}{4})$ that term 1 in $(36)$ $\leq Q_{-M} + \delta^{'}$ and term 2 in $(36)$ $\leq Q_{M} + \delta^{'}$. Setting $P_{M} = max\{Q_{-M}, Q_{M}\}$, we then get that: 
\[
\mathbbm{P}_{X, \overline{Y}}\left[ f^{*}(x, \overline{y}) \neq f^{*}(x, \overline{y}) \right] \leq (1-\overline{\pi})(P_{M} + \delta^{'}) + \overline{\pi}(P_{M} + \delta^{'}) = P_{M} + \delta^{'}
\]
We have thus shown that, with probability at least $(1-\frac{\delta}{4})$, $\delta^{'} + P_{M}$ is an upper bound on $\mathbbm{P}_{X}\left[f^{*}(x)\neq \hat{f}(x)\right]$. This is precisely the upper bound $'B'$, we had wanted to characterise. We can now state our main result for this setting below: 
\\\\
{\bf Theorem C.7 \hspace{0.1em} \it Let $\smash{\delta, \delta^{''}, \epsilon \in (0, \frac{1}{2})}$. Set $\delta^{'} = \max\{\frac{\delta^{''}}{\overline{\pi}}, \frac{\delta^{''}}{1-\overline{\pi}}\}$. Let $T_{-1}(\lambda, \pi, c, \overline{c}) = \left(c - \lambda\overline{c}\right)$ on the $\eta(\cdot, -1)$ and $T_{1}(\lambda, \pi, c, \overline{c}) = \left(c + \lambda - \lambda\overline{c}\right)$ denote thresholds and $\eta(\cdot, 1)$ axes respectively. Pick any $t > Q = 4G \{max\{c(1-\pi), (1-c)\pi, \vert\lambda\vert\overline{c}(1-\beta), \vert\lambda\vert(1-\overline{c})\beta\}\}$, where $G = max\{ \frac{B}{1-\pi}, \frac{B}{\pi \beta}, \frac{B}{\pi (1-\beta)}\}$, and $B = \delta^{'} + P_{M}$. Here $P_{M} :=  max\{Q_{-M}, Q_{M}\}$, where $Q_{-M} = \mathbbm{P}_{X\vert \overline{Y} = -1}(X_{-M})$ and $Q_{M} = \mathbbm{P}_{X\vert \overline{Y} = -1}(X_{M})$, and $X_{-M} := \{x \in \chi: \vert \eta(x, -1) -T_{-1}(\lambda, \pi, c, \overline{c}) \vert \leq \epsilon\}$, while $X_{M} := \{x \in \chi: \vert \eta(x, 1) -T_{1}(\lambda, \pi, c, \overline{c}) \vert \leq \epsilon\}$ . Provided access to $n \geq m_{\eta}((\epsilon, \delta^{''}), \frac{\delta}{4})$ training samples drawn i.i.d. from $\mathbbm{P}$, the plug-in algorithm yields an estimator $\hat{f}$, such that, with probability at least $(1-\delta) \hspace{0.3em}: \hspace{0.3em} regret_{\mathbbm{P}}^{\psi}(\hat{f}) \leq t$
}
\\\\
Thus, by Theorem C.7, using an argument analogous to the \textbf{Subset Argument} presented following Theorem C.3, the regret can be made to decrease arbitrarily, provided a sufficient increase in the number of training samples. The precise rate of decay depends on 1) the sample complexities associated with learning the regression function and 2) the rate at which the probability measure ($\mathbbm{P}_{X}$) around the thresholds $T_{-1}(\lambda, \pi, c, \overline{c}) = \left(c - \lambda\overline{c}\right)$ and $T_{1}(\lambda, \pi, c, \overline{c}) = \left(c + \lambda - \lambda\overline{c}\right)$ decay (in the $\eta(\cdot, -1), \eta(\cdot, 1)$-axes respectively) upon shrinking the regions of consideration around them.
\subsection{Fairness-Accuracy-Sample Size Trade-Offs}
In this subsection, we study the fairness-accuracy trade-off, provided access to a finite training sample. The first sub-subsection introduces necessary background from \cite{menon2018cost}, leading up to our motivation (presented in the latter half of the sub-subsection) for studying this trade-off under a finite training sample. The second sub-subsection details out our analysis.  
\subsubsection{Background and Discussion}
 It is well known that there is generally a cost to accuracy in training fair classifiers \cite{menon2018cost, zhao2019inherent, chen2018my}. \cite{menon2018cost} characterise this via the fairness frontier, which quantifies the fundamental limits of the accuracy achievable by any classifier for fairness aware learning problems of the kind introduced in Problem 2.3 of the main text. To understand this better, let us first revisit our fairness aware learning (FAL) problem. Recall that the FAL objective, for some trade-off parameter $\lambda \in \mathbbm{R}$, and cost parameters, $c, \overline{c} \in (0, 1)^{2}$ is given by:
\begin{equation}
    min_{f: \chi \rightarrow [0, 1]} \underbrace{CS(f; \mathcal{D}, c)}_\text{term 1} - \lambda \underbrace{CS(f; \overline{\mathcal{D}}, \overline{c})}_\text{term 2}
\end{equation}

Recall that the cost sensitive risk (CSR) is a convex combination of errors. Indeed, term 1 is a measure of weighted error for classifier $f$, w.r.t. predicting the label. Ideally, we would want our classifier to predict the label perfectly, so we would like to ensure term 1 be as close to 0 as possible. Term 2 on the other hand quantifies the classifier's ability to predict the sensitive attribute. Intuitively, we want our classifier's predictions to be independent of the sensitive attribute. We would thus require $f$ to not be predictive of the sensitive attribute, i.e., we would like term 2 to be close to 1. Of course, we can not generally hope to achieve perfect accuracy and/ or perfect fairness and thus a balance needs to be struck. Indeed, an equivalent formulation of the FAL problem, asks to minimize term 1, while ensuring term 2 be larger than some $\tau \in [0, 1]$. We can formulate this as follows: 

\begin{equation}
min_{f} {CS(f; \mathcal{D}, c)}  \quad s.t.\quad CS(f; \overline{\mathcal{D}}, \overline{c}) \geq \tau
\end{equation}

For a thorough discussion on the equivalence between problems $(37)$ and $(38)$, we refer the reader to \cite{menon2018cost}. Thus for a given $\tau \in [0, 1], \hspace{0.3em} \exists \lambda \in \mathbbm{R}$, for which the FAL problem formulated in $(38)$, reduces to its soft constrained version formulated in $(37)$. The problem without any fairness constraint, would require setting $\lambda, \tau = 0$ in $(37)/(38)$, as doing so would require minimizing term 1 alone, i.e., the cost sensitive risk associated the classifier's predictive ability with respect to the label. For a given $\tau \in [0, 1]$, \cite{menon2018cost} define the fairness frontier, via $F:[0, 1] \rightarrow \mathbbm{R}$, as: 
\[
F(\tau) := CS(f_{\tau}^{*}; \mathcal{D}, c) - CS(f_{0}^{*}; \mathcal{D}, c)
\]
where, $f_{\tau}^{*} \in Argmin_{f:\chi \rightarrow [0, 1]}\{{CS(f; \mathcal{D}, c)}  \quad s.t.\quad CS(f; \overline{\mathcal{D}}, \overline{c}) \geq \tau\}$ and $f_{0}^{*} \in Argmin_{f:\chi \rightarrow [0, 1]}\{{CS(f; \mathcal{D}, c)}\}$. Thus, $F(\tau)$ quantifies the degradation in predictive accuracy (in terms of the CSR) upon imposing a fairness constraint of the kind in $(38)$ parameterised via $\tau$, provided optimal classifiers can be retrieved. As noted in \cite{menon2018cost}, $f_{0}^{*}(x) = \mathbbm{I}\{sign \circ \{\eta(x) - c\}\}$. We present a result of \cite{menon2018cost}, which characterises $F(\tau)$ in terms of $\lambda$: 

{\bf Lemma C.8 \hspace{0.1em} \it Pick any cost parameters, $c, \overline{c} \in (0, 1)$. For any $\tau \in [0, 1]$, there is some $\lambda \in \mathbbm{R}$ and Bayes Optimal randomised classifier $f_{\lambda}^{*} \in \{CS(f; \mathcal{D}, c) - \lambda CS(f; \overline{\mathcal{D}}, \overline{c})\}$ so that the frontier is: 
\[
F(\tau) = \mathbbm{E}_{X}\left[(c-\eta(x))(f_{\lambda}^{*}(x) - \mathbbm{I}\{\eta(x) > c\})\right]
\]
}
\\\\
Recall that, $F(\tau)$ is defined with respect to the 'optimal' classifiers, $f_{0}^{*}, f_{\tau}^{*}$, corresponding to the fairness unaware and the fairness aware learning problems respectively. The optimal classifiers are characterised via distributional quantities, namely the regression functions, which we do not have access to in general. Thus, in practice, we can only hope for estimators that closely approximate the optimal classifiers. As we saw in our non-asymptotic analysis, the estimation error for the plug-in algorithm can be characterised on the basis of the sample size of the training data. In this subsection as well, we wish to understand the underlying fairness-accuracy trade-off, provided only a finite training sample.\\\\
We show that under a finite sample setting, the (provable) accuracy degradation incurred in transitioning from a fairness unaware setting to a fairness aware learning setting is in fact more than $F(\tau)$. In other words, owing to the finite training sample, there is a cost to pay in terms of the best achievable fairness-accuracy trade-off. This cost can be provably reduced with an increase in sample size. \textbf{The core message here is that there is not solely a fairness-accuracy trade-off, however, there is effectively a fairness-accuracy-sample size trade-off}, which we characterise more precisely in the next sub-subsection.
\\\\
Recall that for a given $\tau \in [0, 1]$, $\exists \lambda \in \mathbbm{R}$ that leads to an equivalence in the problems laid out in $(37)$ and $(38)$. Denote $\lambda(\tau)$ as the $\lambda$ (as in $(37)$) that leads to an equivalent FAL as $\tau$ (as in $(38)$). We will overload notation and introduce the function $G: \mathbbm{R} \rightarrow \mathbbm{R}, s.t.,  G(\lambda(\tau)) = F(\tau)$. Since we will work in the soft-constrained realm of the FAL problem, we will drop the explicit dependence of $\lambda$ on $\tau$, and simply write $G(\lambda)$. Thus, $G(\lambda)$ quantifies the same fairness frontier as $F(\tau)$. Keeping this in mind, we are now ready to proceed with our finite-sample trade-off analysis. 
\subsubsection{Analysis}
We will present our analysis only for the Approximate Equality of Opportunity case when the sensitive attribute is unavailable at test-time. Thus, $\overline{\mathcal{D}}, \overline{\eta}$ will be used to refer to $\overline{\mathcal{D}}_{EO}, \overline{\eta}_{EO}$ respectively. The analysis can be extended analogously to other settings of interest.
\\\\
Assume access to a training data set, $S = \{x_{i}, \overline{y}_{i}, y_{i}\}_{i=1}^{n}$, comprised of $n$ samples drawn i.i.d. from $\mathbbm{P}$, the joint distribution over $(X, \overline{Y}, Y)$. Let $f_{\lambda}^{*} \in Argmin_{f:\chi \rightarrow[0, 1]}\{CS(f; \mathcal{D}, c) - \lambda CS(f; \overline{\mathcal{D}}, \overline{c})\}$ and $f_{0}^{*} \in Argmin_{f:\chi \rightarrow[0, 1]}\{CS(f; \mathcal{D}, c)\}$. Denote $\hat{\eta}, \hspace{0.3em} \hat{\overline{\eta}}$ as our estimators for the true regression functions, $\eta, \hspace{0.3em} \overline{\eta}$. Let $ \hat{f}_{\lambda}$, $\hat{f}_{0}$ be the plug-in estimators corresponding to $f_{\lambda}^{*}$ and $f_{0}^{*}$ respectively. So, $\forall x \in \chi$:
\[
f_{\lambda}^{*}(x) = sign\circ\{(1 + \frac{\lambda\overline{c}}{\pi})\eta(x) - \frac{\lambda}{\pi}\overline{\eta}(x, 1)\eta(x) -c\} \quad and, \quad f_{0}^{*}(x) = sign \circ \{\eta(x) - c\}
\]

\[
\hat{f}_{\lambda} (x) = sign\circ\{(1 + \frac{\lambda\overline{c}}{\pi})\hat{\eta}(x) - \frac{\lambda}{\pi}\hat{\overline{\eta}}(x, 1)\hat{\eta}(x) -c\} \quad and, \quad \hat{f}_{0}(x) = sign \circ \{\hat{\eta}(x) - c\}
\]
Let $\smash{\delta, \delta^{'}, \epsilon \in (0, \frac{1}{2})}$. Now, suppose $n \geq max\{m_{\eta}((\epsilon, \frac{\delta^{'}}{2}), \frac{\delta}{8}), m_{\overline{\eta}}((\epsilon, \frac{\delta^{'}}{2}), \frac{\delta}{8})\}$. This implies by Definitions C.1, C.2 and an application of the union bound that we can obtain estimators $\hat{\eta}\hspace{0.3em}, \hat{\overline{\eta}}$, such that, with probability $\geq \left(1 - \frac{\delta}{4}\right)$: 
\begin{equation}
    \mathbbm{P}_{X}(\vert \eta(x) - \hat{\eta}(x) \vert \geq \epsilon) \leq \frac{\delta^{'}}{2} \quad and, \quad \mathbbm{P}_{X}(\vert \overline{\eta}(x, 1) - \hat{\overline{\eta}}(x, 1) \vert \geq \epsilon) \leq \frac{\delta^{'}}{2}
\end{equation}

Let $G(\lambda) = CS(f_{\lambda}^{*}; \mathcal{D}, c) -  CS(f_{0}^{*}; \mathcal{D}, c)$ be the fairness frontier. Now, for $t > 0$, we have that: 
\[
\mathbbm{P}_{S \sim \mathbbm{P}^{n}}\left[ \vert CS(\hat{f}_{\lambda}; \mathcal{D}, c) -  CS(\hat{f}_{0}; \mathcal{D}, c) \vert > G(\lambda) + t \right]
\]
\[
   \overset{(a)}{\leq} \mathbbm{P}_{S \sim \mathbbm{P}^{n}}\left[ \vert CS(\hat{f}_{\lambda}; \mathcal{D}, c) -  CS(f_{\lambda}^{*}; \mathcal{D}, c) \vert + \vert CS(f_{\lambda}^{*}; \mathcal{D}, c) -  CS(f_{0}^{*}; \mathcal{D}, c) \vert + \vert CS(f_{0}^{*}; \mathcal{D}, c)-  CS(\hat{f}_{0}; \mathcal{D}, c) \vert > G(\lambda) + t \right] 
\]
\[
\overset{(b)}{=} \quad \mathbbm{P}_{S \sim \mathbbm{P}^{n}}\left[ \vert CS(\hat{f}_{\lambda}; \mathcal{D}, c) -  CS(f_{\lambda}^{*}; \mathcal{D}, c) \vert + \vert CS(f_{0}^{*}; \mathcal{D}, c)-  CS(\hat{f}_{0}; \mathcal{D}, c) \vert >  t \right]
\]
\begin{equation}
\overset{(c)}{\leq} \quad \mathbbm{P}_{S \sim \mathbbm{P}^{n}}\left[ \vert CS(\hat{f}_{\lambda}; \mathcal{D}, c) -  CS(f_{\lambda}^{*}; \mathcal{D}, c) \vert > \frac{t}{2}\right] \hspace{0.3em} + \hspace{0.3em } \mathbbm{P}_{S \sim \mathbbm{P}^{n}}\left[\vert CS(f_{0}^{*}; \mathcal{D}, c)-  CS(\hat{f}_{0}; \mathcal{D}, c) \vert >  \frac{t}{2}\right]
\end{equation}
Here, (a) follows by adding and subtracting $CS(f_{\lambda}^{*}; \mathcal{D}, c)$ and by adding and subtracting  $CS(f_{0}^{*}; \mathcal{D}, c)$ and then applying the triangle inequality. (b) follows due to the definition of the fairness frontier, and (c) follows by an application of the union bound. Now, following precisely the steps laid out in the General Template of C.2, we get that: 

\begin{equation}
\begin{split}
(40) \quad {\leq} \quad \mathbbm{P}_{S\sim \mathbbm{P}^{n}} \Bigg[ \underbrace{\{c(1-\pi)\}\mathbbm{P}_{X \vert Y = -1}\Big[f_{\lambda}^{*}(x) \neq \hat{f}_{\lambda}(x)\Big]}_\text{term 1} > \frac{t}{4}\Bigg]  \\ + \quad  \mathbbm{P}_{S\sim \mathbbm{P}^{n}} \Bigg[ \underbrace{\{(1-c)\pi\}\mathbbm{P}_{X \vert Y = 1}\Big[f_{\lambda}^{*}(x) \neq \hat{f}_{\lambda}(x)\Big] }_\text{term 2} > \frac{t}{4}\Bigg] \\
+ \quad \mathbbm{P}_{S\sim \mathbbm{P}^{n}} \Bigg[ \underbrace{\{c(1-\pi)\}\mathbbm{P}_{X \vert Y = -1}\Big[f_{0}^{*}(x) \neq \hat{f}_{0}(x)\Big]}_\text{term 3} > \frac{t}{4}\Bigg]  \\ + \quad  \mathbbm{P}_{S\sim \mathbbm{P}^{n}} \Bigg[ \underbrace{\{(1-c)\pi\}\mathbbm{P}_{X \vert Y = 1}\Big[f_{0}^{*}(x) \neq \hat{f}_{0}(x)\Big] }_\text{term 4} > \frac{t}{4}\Bigg]
\end{split} 
\end{equation}
\\\\
Now, conditional on the $(1-\frac{\delta}{4})$-event of $(39)$, let $B\in\mathbbm{R}$, such that:
\begin{equation}
max\bigg\{\mathbbm{P}_{X}\left[f_{\lambda}^{*}(x)\neq \hat{f}_{\lambda}(x) \right], \mathbbm{P}_{X}\left[f_{0}^{*}(x)\neq \hat{f}_{0}(x) \right]\bigg\} \leq B
\end{equation}

Let us characterise $B$ in $(42)$. First see, to obtain an upper bound on $\mathbbm{P}_{X}\left[f_{\lambda}^{*}(x)\neq \hat{f}_{\lambda}(x) \right]$, we appeal to the analysis laid out in C.2.1, from where we get that:  $\mathbbm{P}_{X}\left[f_{\lambda}^{*}(x)\neq \hat{f}_{\lambda}(x) \right] \leq$ $B_{1} = \delta^{'} + \mathbbm{P}_{X}(X_{M})$ conditional on the $(1-\frac{\delta}{4})$-event of $(39)$. Here $X_{M} := \{x \in \chi:$ \textit{the square of length $2\epsilon$ centred at $(\overline{\eta}(x, 1), \eta(x))$ intersects the hyperbola $H(\lambda, \pi, c, \overline{c})$ in the $(\overline{\eta}(\cdot, 1), \eta)$-plane\}.} \\\\
Now, upper bounding $\mathbbm{P}_{X}\left[f_{0}^{*}(x)\neq \hat{f}_{0}(x)\right]$, conditional on the $(1-\frac{\delta}{4})$-event of $(39)$, is relatively easy. Define $X_{bad} := \{x \in \chi: \vert \eta(x) - \hat{\eta}(x) \vert  \geq \epsilon\}$. We know by $(39)$, that $P_{X}(X_{bad}) \leq \frac{\delta^{'}}{2}$. Also define $X_{M^{0}} := \{x \in \chi: \vert \eta(x) -c \vert < \epsilon\}$. Thus, if $x \in X_{M^{0}}$, there is a non-zero probability, that $f_{0}^{*}(x)\neq \hat{f}_{0}(x)$, regardless of whether $x \in X_{bad}$ or not. Thus, if $x \notin \{X_{bad}\bigcup X_{M^{0}}\}$, then conditional on the $(1-\frac{\delta}{4})$-event of $(39)$, we can be certain that $f_{0}^{*}(x)= \hat{f}_{0}(x)$. We have thus shown that, with probability at least $(1-\frac{\delta}{4})$:
\[
\mathbbm{P}_{X}\left[f_{0}^{*}(x)\neq \hat{f}_{0}(x)\right] \leq \frac{\delta^{'}}{2} + \mathbbm{P}_{X}\left[X_{M^{0}}\right] = B_{2}
\]
Now, we set $B = max\{B_{1}, B_{2}\}$ and  $G = max\{\frac{B}{1-\pi}, \frac{B}{\pi}\}$. Further, set $t$, such that, $min\{\frac{t}{4c(1-\pi)}, \frac{t}{4(1-c)\pi}\} > G \iff t > Q:= 4G \hspace{0.3em}max\{c(1-\pi), (1-c)\pi\}$. Now, just like we showed in the General Template presented in C.2, we have that with probability at least  $(1-\frac{\delta}{4})$, each of terms 1 to 4 in $(41)$ are at most $\frac{t}{4}$. Thus collecting the failure probabilities and plugging them into $(41)$, we get that $(41) \leq \delta$. We can now state our key result precisely.  

{\bf Theorem C.9 \hspace{0.1em} \it Let $\lambda \in \mathbbm{R}$ and $G(\lambda)$ be the corresponding fairness frontier.  Let $f_{\lambda}^{*} \in Argmin_{f:\chi \rightarrow[0, 1]}\{CS(f; \mathcal{D}, c) - \lambda CS(f; \overline{\mathcal{D}}, \overline{c})\}$ and $f_{0}^{*} \in Argmin_{f:\chi \rightarrow[0, 1]}\{CS(f; \mathcal{D}, c)\}$. Let $ \hat{f}_{\lambda}$, $\hat{f}_{0}$ be the plug-in estimators corresponding to $f_{\lambda}^{*}$ and $f_{0}^{*}$ respectively.  Let $\smash{\delta, \delta^{'}, \epsilon \in (0, \frac{1}{2})}$. Consider in the $(\overline{\eta}(\cdot, 1), \eta)$-plane, the hyperbola $H(\lambda, \pi, c, \overline{c}) := \{(1 + \frac{\lambda\overline{c}}{\pi})\eta - \frac{\lambda}{\pi}\overline{\eta}(\cdot, 1)\eta - c = 0$\}. Pick any $t > Q = 4G \hspace{0.3em}max\{c(1-\pi), (1-c)\pi\}$, where $G = max\{\frac{B}{1-\pi}, \frac{B}{\pi}\}$. Here $B := max\{B_{1}, B_{2}\}$, where $B_{1} = \delta^{'} + \mathbbm{P}_{X}(X_{M})$. Here $X_{M} := \{x \in \chi:$ the square of length $2\epsilon$ centred at $(\overline{\eta}(x, 1), \eta(x))$ intersects the hyperbola $H(\lambda, \pi, c, \overline{c})$ in the $(\overline{\eta}(\cdot, 1), \eta)$-plane\}. And $B_{2} = \frac{\delta^{'}}{2} + \mathbbm{P}_{X}\left[X_{M^{0}}\right]$. Here $X_{M^{0}} := \{x \in \chi: \vert \eta(x) -c \vert < \epsilon\}$. Provided access to $n \geq max\{m_{\eta}((\epsilon, \frac{\delta^{'}}{2}), \frac{\delta}{8}), m_{\overline{\eta}}((\epsilon, \frac{\delta^{'}}{2}), \frac{\delta}{8})\}$ training samples drawn i.i.d. from $\mathbbm{P}$, it holds with probability at least $(1-\delta)$:  $\vert CS(\hat{f}_{\lambda}; \mathcal{D}, c) -  CS(\hat{f}_{0}; \mathcal{D}, c) \vert \leq G(\lambda) + t $
}
\\\\
Thus, we have shown that there is a fairness-accuracy-sample size trade-off. In other words, provided a finite training sample, the degradation in predictive accuracy when transitioning from a fairness-unaware to a fairness-aware learning setting, exceeds the degradation incurred were we to have access to the true distribution, by an amount $t$. The excess degradation, is quantified via the term $t$ in Theorem C.9. Using an argument analogous to the \textbf{Subset Argument} presented following Theorem C.3, we can ensure that the excess degradation, $t$, can be made arbitrarily small, by increasing the size of the training data set sufficiently. Notice that we made use of the same General Template of C.2 used to derive finite sample guarantees for the plug-in algorithm's performance w.r.t. the performance measure, $\mathfrak{P}_{\mathbbm{P}}^{\Psi}$ (along with the upper bound characterisation in C.2.1). This finite sample trade-off result can thus be analogously extended to other settings of interest. 

\section{Fairness under Differential Privacy}
In this section, we will consider the setting wherein, in addition to achieving approximate fairness w.r.t. the sensitive attribute, there is also a requirement to ensure that the model does not leak information about the sensitive attribute corresponding to individuals in the training data. Indeed, as noted in \cite{jagielski2019differentially}, collecting information about features such as race, gender, ethnicity etc. is often restricted or prohibited. In such situations, protecting sensitive data via differential privacy \cite{dwork2006calibrating} is a possible solution. The literature on machine learning models that promise fairness and privacy simultaneously is limited and emerging \cite{jagielski2019differentially, cummings2019compatibility, mozannar2020fair}. Motivated by this, we propose a modified version of the plug-in algorithm of \cite{menon2018cost} that guarantees privacy with respect to the sensitive attribute. We will call this algorithm DP Plug-in.\\\\
In the first sub-section, we will present necessary background on differential privacy and its application to machine learning. The second sub-section presents the DP-Plug-in Algorithm. We will consider a specific instantiation of our algorithm based on the 'Output Perturbation' method of \cite{chaudhuri2011differentially}. We will derive the finite sample guarantees associated with this instantiation. In the final sub-section, we will demonstrate DP Plug-in's performance empirically on standard publicly available data sets. Our fairness criterion throughout this section will be that of Approximate Equality of Opportunity, though our algorithm and corresponding analysis extends analogously to the other fairness criteria we have studied. Throughout this section, $\overline{\eta}$ denotes $\overline{\eta}_{EO}$ and $\mathcal{D}$ denotes $\mathcal{D}_{EO}$.

\subsection{Background}
Differential privacy was first introduced in \cite{dwork2006calibrating}. Roughly speaking, a mechanism applied to a data set, is differentially private if its output does not change significantly, once applied to a data set that differs in exactly one row (i.e., a single individual's presence/ absence in the data does not have much influence on the mechanism's output).  We denote $d(D, D^{'})$ as the number of instances two data sets $D, D^{'}$ differ in. $D, D^{'}$ are called neighboring data sets if $d(D, D^{'}) = 1$. Let $\mathfrak{D}$ denote a data domain. Formally, differential privacy can be defined as follows: 

{\bf Definition D.1 \hspace{0.1em} \it A randomized algorithm $M:D^{m} \mapsto Z$ is said to be $(\epsilon,\delta)-$differentially private if for all  $D, D^{'} \in \mathfrak{D}^{m}$, such that $d(D, D^{'}) = 1$ and all $z \subseteq Z$, 

\begin{equation}
\mathbbm{P}(M(D) \in z) \leq e^{\epsilon}\mathbbm{P}(M(D') \in z) + \delta
\end{equation}

if $\delta = 0$, $M$ is said to be $\epsilon-$differentially private
}

Thus, upon observing a differentially private algorithm's output, an adversary can not infer with much certainty, whether or not a specific individual was included in the data set upon which the mechanism was applied. For some $\epsilon >0$, setting $\delta > 0$ gives us $(\epsilon, \delta)$-differential privacy, also known as approximate differential privacy. Setting $\delta = 0$ in Definition D.1 yields $\epsilon$-differential privacy or pure differential privacy. Note, in our context we are only interested in protecting the privacy of individuals with respect to their sensitive attributes. Suppose that our data domain is given by $\left(\chi \bigtimes \mathcal{Y} \bigtimes \overline{\mathcal{Y}}\right)$, i.e., features vectors $X \in \chi$, labels $Y \in \mathcal{Y} = \{-1, 1\}$ and sensitive attribute $\overline{Y} \in \overline{\mathcal{Y}} = \{-1, 1\}$. Given a data set $S = \{x_{i}, y_{i}, \overline{y}_{i}\}_{i=1}^{n}$ of size $n$, let us partition $S$ as follows $S = (S^{'}, \overline{S})$, where $S^{'} = \{x_{i}, y_{i}\}_{i=1}^{n}$ and $\overline{S} = \{\overline{y}_{i}\}_{i=1}^{n}$.

{\bf Definition D.2 \hspace{0.1em} \it A randomized algorithm $M: \left(\chi \bigtimes \mathcal{Y} \bigtimes \overline{\mathcal{Y}}\right) \rightarrow Z$ is $(\epsilon, \delta)$-differentially private with respect to the sensitive attribute if $\forall (\overline{S}, \overline{S}_{1}) \in \overline{\mathcal{Y}}^{2}: d(\overline{S}, \overline{S}_{1}) = 1$ and $\forall z \subseteq Z$:

\begin{equation}
\mathbbm{P}(M((S^{'}, \overline{S})) \in z) \leq e^{\epsilon}\mathbbm{P}(M(S^{'}, \overline{S}_{1}) \in z) + \delta
\end{equation}
if $\delta = 0$, $M$ is said to be $\epsilon-$differentially private with respect to the sensitive attribute
}

Ensuring an algorithm is differentially private typically requires adding carefully calibrated noise to the model output. There are several noise-addition mechanisms, including the Laplace mechanism \cite{dwork2006calibrating}, the Gaussian mechanism \cite{dwork2006calibrating} and the Exponential mechanism \cite{mcsherry2007mechanism}. As the names suggest, these mechanisms are defined on the basis of the type of distributional noise being added. Determining the magnitude of noise to be added, requires computing the sensitivity of the mechanism. Roughly, the sensitivity determines the maximal amount by which a mechanism's output differs between neighbouring data sets. Intuitively, the smaller the sensitivity, the more inherently robust the mechanism is to row replacements. Thus the amount of noise required to be added is roughly proportional to the sensitivity of the mechanism. We define this below: 

{\bf Definition D.3 \hspace{0.1 em} \it The $\ell_{2}$- sensitivity of a function $f: D^{m} \mapsto  \mathbb{R}^k$ is defined as:
\begin{equation}
\Delta f = \max_{D,D^{'} \in D^{m}; \ {d(D, D^{'}) = 1}} \| f(D) - f(D^{'})\|_{2}
\end{equation}}

An important property of differential privacy is that it is preserved under post-processing. This means that any data independent operations on a differentially private output cause no degradation in privacy guarantees. The post-processing property will be important in our presentation of DP Plug-in's privacy guarantees. Formally, we can define this as follows: 

{\bf Definition D.4 \hspace{0.1 em} \it Let $M:D^{m} \mapsto Z$ be a $(\epsilon,\delta)-$differentially private function, and let $f:Z \mapsto A$ be a randomized function. Then the function $f\circ M:D^{m} \mapsto A$ is also $(\epsilon,\delta)-$differentially private}

 There has been considerable interest in the realm of differentially private machine learning \cite{chaudhuri2011differentially, jayaraman2019evaluating, chaudhuri2013stability, abadi2016deep}. Differential privacy in the context of machine learning typically requires that the model so learnt is differentially private. We will provide the details of the 'Output Perturbation' approach of \cite{chaudhuri2011differentially} to differentially private machine learning. First, let us introduce the basic set-up laid out in their work. \\\\
 
 \cite{chaudhuri2011differentially} requires training data $D = \{x_{i}, y_{i}\}_{i=1}^{n}$ where $\forall i \in [n]$, $(x_{i}, y_{i})$ denote feature label pairs. It is assumed that $\forall i \in [n], \|x_{i}\|_{2} \leq 1$. Denoting $\chi$ as the feature domain, and  $\mathcal{Y} = \{-1, 1\}$ as the label domain, their goal is to learn a predictor $f: \chi \rightarrow \mathcal{Y}$, such that for a non-negative loss function $\ell: (\mathcal{Y}\bigtimes\mathcal{Y}) \rightarrow \mathbb{R}$, $f$ minimizes the following regularzied ERM objective: 
 
 \begin{equation}
     J(f, D) = \frac{1}{n}\sum_{i=1}^{n}\ell(f(x_{i}), y_{i}) + \Lambda N(f) 
 \end{equation}
 
  Now, let us describe the 'Output Perturbation' approach of \cite{chaudhuri2011differentially}. This is a simple two step procedure applicable to parametric machine learning models. The first step entails learning a non-private model via solving a regularized empirical risk minimization (ERM) problem of form $(46)$. The second step involves adding carefully calibrated noise to the model weights so learnt. The noise calibration is carried out on the basis of the $\ell_{2}$ sensitivity of the regularized ERM problem. The regularization is important. Intuitively, higher regularization implies a more constrained parameter space, implying smaller $\ell_{2}$-sensitivity of the regularized ERM. This is formalised via the following result from \cite{chaudhuri2011differentially}:
 \\\\
 {\bf Lemma D.5 \hspace{0.1 em} \it If $N(\cdot)$ is differentiable and 1-strongly convex, and $\ell$ is convex and differentiable with $\vert \ell^{'}(z) \vert \leq 1, \hspace{0.3em} \forall z$, then the $\ell_{2}$ sensitivity of $argmin J(f, D)$ is at most $\frac{2}{n\Lambda}$}
 
 Here $z$ denotes a feature label pair. For some normalizing constant $\alpha_{p}$, \cite{chaudhuri2011differentially} define the following density, which is used in their output perturbation method: 
 
 \begin{equation}
  v(b) = \frac{1}{\alpha_{p}}e^{-\Gamma \|b\|_{2} }
 \end{equation}
 
 Algorithm 7 lays out the steps involved in the output perturbation technique.
 \begin{algorithm}
   \caption{\small{ERM with Output Perturbation}}
\begin{algorithmic}

   \STATE \small{{\bfseries Input:} Data $D = \{z_{i}\}_{i=1}^{n}$, privacy parameter $\epsilon_{p}$, regularization parameter $\Lambda$}
   \vspace{2.5mm}
   \STATE {\bfseries Output:} Approximate minimizer $f_{priv}$
    \vspace{2.5mm}
   \STATE {\bfseries Estimate:} Draw a vector $b$ according to $(47)$ with $\Gamma = \frac{n\Lambda\epsilon_{p}}{2}$ 
   \STATE {\bfseries Compute:} \small{$f_{priv} = argmin J(f; D) + b$}
\end{algorithmic}
\end{algorithm}
\\\\
Under specific conditions on the loss function $\ell$ and the regularization term $N(\cdot)$, Algorithm 7 is $\epsilon_{p}$-differntially private, as formalised via the following result of \cite{chaudhuri2011differentially}:
 \\\\
 {\bf Theorem D.6 \hspace{0.1 em} \it If $N(\cdot)$ is differentiable and 1-strongly convex, and $\ell$ is convex and differentiable with $\vert \ell^{'}(z) \vert \leq 1, \hspace{0.3em} \forall z$, then Algorithm 7 is $\epsilon_{p}$-differentially private}
 
 In the empirical section of our work, we will make use of Algorithm 7. It is easy to see that $(\epsilon, \delta)$-differentially privacy (or $\epsilon$-DP) with respect to the entire data vector $\implies$ $(\epsilon, \delta)$-differentially privacy (receptively $\epsilon$-DP) with respect to sensitive attribute, since for any mechanism, if Definition $D.1$ holds then Definition $D.2$ is implied. From this observation, it is clear that Algorithm 7 is $\epsilon_{p}$-differentially private with respect to the sensitive attribute in the case where we learn a classifier via Algorithm 7, that takes as input feature-label pairs and attempts to learn the sensitive attribute (this would be a classifier corresponding to the regression function $\overline{\eta}$). Following the proof steps of \cite{chaudhuri2011differentially} for sensitivity derivation, we would actually obtain the same $\ell$-2 sensitivity for $argmin J(f, D)$ considering only replacements in the sensitive attribute (Provided conditions on the loss function and regularize laid out in Theorem D.6 hold). 
 
 \subsection{Differentially Private Fairness Aware Plug-in Algorithm}
 In this subsection, we will present the DP Plug-in Algorithm. We assume access to a training data set $S = \{(x_{i}, y_{i}, \overline{y}_{i})\}_{i=1}^{n}$ drawn i.i.d. from distribution $\mathbb{P}$. $\forall i,$ the triplet $(x_{i}, y_{i}, \overline{y}_{i})$ is a realization of the random variable triplet $(X, Y, \overline{Y})$ denoting the feature, label and sensitive attribute respectively. Let the feature domain be denoted $\chi$, label domain be denoted $\mathcal{Y} = \{-C, C\}$, for some positive constant $C$. Let the sensitive attribute domain denoted $\mathcal{\overline{Y}} = \{-1, 1\}$. Let us also assume that $\|(x_{i}, y_{i})\|_{2} \leq 2$ (this is why we use label domain $\mathcal{Y} = \{-C, C\}$ rather than the standard $\mathcal{Y} = \{-1, 1\}$). The DP Plug-in algorithm is presented in Algorithm 8. \\\\
\begin{algorithm}[H]
   \caption{\small{DP Plugin approach to FAL, EO setting}}
\begin{algorithmic}
   \STATE \small{{\bfseries Input:} Sample $S$ = $\{x_{i}, y_{i}, \overline{y}_{i}\}_{i=1}^{n}$ from distribution $\mathbbm{P}$; cost parameters $c, \overline{c}$; trade-off parameter $\lambda$}; privacy parameter $\epsilon_{p}$
   \vspace{2.5mm}
   \STATE {\bfseries Estimate:} \small{$\pi$ via \hspace{0.5 em}$\hat{\pi} = \frac{1}{n}\sum_{i=1}^{n}\mathbbm{I}\{y_{i} = 1\}$}
   \STATE {\bfseries Estimate:} \small{$\eta$:$\chi \rightarrow [0, 1]$ using appropriate CPE on $\{x_{i}, y_{i}\}_{i=1}^{n}$}
   \STATE {\bfseries Estimate:} \small{$\overline{\eta}_{EO}$: $(\chi, Y) \rightarrow [0, 1]$ using appropriate CPE on $S$}
   \STATE {\bfseries Privatise:} \small{$\hat{\overline{\eta}}_{EO}$ via appropriate privacy preserving protocol yielding, $\epsilon_{p}$-DP protected $\hat{\overline{\eta}}_{EO}^{priv}$}
   \STATE {\bfseries Compute:} $\hat{s}^{priv}\left(x\right) =$  \begin{scriptsize}$\smash{\left\{ 1 - \frac{\lambda}{\hat{\pi}}(\hat{\overline{\eta}}_{EO}^{priv}(x, 1) - \overline{c}) \right\}\hat{\eta}(x) - c}$\end{scriptsize}
   \vspace{2.5mm}
   \STATE {\bfseries Return:} \small{$\hat{f}^{priv}\left(x\right) = H_{\alpha}\left(\hat{s}^{priv}\left(x\right)\right)$ for any $\alpha \in [0, 1]$}
\end{algorithmic}
\end{algorithm}
Note that the only step in the pipeline that involves interacting with the sensitive attribute directly, is the estimation of $\overline{\eta}$. Thus, we need to ensure that the estimator so learnt, i.e., $\hat{\overline{\eta}}$ is appropriately noised so as to obtain $\hat{\overline{\eta}}^{priv}$ which preserves $\epsilon_{p}$-differential privacy with respect to $\overline{Y}$. In general, the DP Plugin protocol allows us to use any $\epsilon_{p}$-DP procedure to obtain $\hat{\overline{\eta}}^{priv}$. Post this, interaction with $\overline{Y}$ only takes place via $\hat{\overline{\eta}}^{priv}$ in the computation of $\hat{s}^{priv}$ and $\hat{f}^{priv}$. Thus by the post-processing property of differential privacy (see Definition D.4), the entire pipeline is $\epsilon_{p}$-differentially private, and is therefore $\epsilon_{p}$-differentially private w.r.t. the sensitive attribute. We state this as Lemma D.7 below

{\bf Lemma D.7 \hspace{0.1 em} \it The DP Plug-in algorithm is $\epsilon_{p}$-differentially private with respect to the sensitive attribute}
\subsubsection{DP Plug-in via Output Perturbation - Introduction and Theory}
In this sub-subsection, we will consider an instantiation of the DP Plug-in algorithm which makes use of the output perturbation approach of \cite{chaudhuri2011differentially} as its privacy protocol (see Algorithm 9). We make assumptions about the structure of the regression function, $\overline{\eta}$, and its candidate estimators, such as $\hat{\overline{\eta}}$. In particular, we will assume that, $\overline{\eta}, \hat{\overline{\eta}}$ can be parameterised by parameter vectors $\in \mathcal{\overline{W}}$, where $\mathcal{\overline{W}} \subset \mathbb{R}^{d+1}$ is a compact set. Specifically, we assume, $\exists g: (\mathcal{\overline{W}}, (\chi, \mathcal{Y})) \rightarrow [0, 1]$ and $\exists \overline{w}^{*} \in \mathcal{W}$, such that, $\forall x \in \chi$, and for any $y \in \mathcal{Y}$, \hspace{0.3em}   $\overline{\eta}(x, y) = g(\overline{w}^{*}, (x, y))$ and for any estimator we consider, $\exists \hat{\overline{w}} \in \mathcal{W}$, such that the estimator takes form, $\hat{\overline{\eta}}(x, y) = g(\hat{\overline{w}}; (x, y))$. Further, like in previous sections, Assumption 2 from the main text holds, i.e., $\hat{\eta}, \hat{\overline{\eta}}$ are $L$-1 consistent class probability estimators. Suppose further that $\hat{\overline{\eta}}$ is the class probability estimator obtained by solving a regularized ERM over a loss $\ell(\cdot, \cdot)$ and regularizer $N(\cdot)$ which satisfy conditions laid out in Theorem D.6.  Fix $\Lambda \in \mathbb{R}$. In our context, the ERM objective here becomes: 
\begin{equation}
 J(f, S) = \frac{1}{n}\sum_{i=1}^{n}\ell(f((x_{i}, y_{i}), \overline{y}_{i})) + \Lambda N(f) 
 \end{equation}
Let $\hat{\overline{w}} \in \mathcal{\overline{W}}$ denote the parameters corresponding to the class probability estimator,  $\hat{\overline{\eta}}$ obtained by solving $(48)$. Now, substituting $argmin J(f; D)$ in Algorithm 7 with $\hat{\overline{w}}$, we have by Theorem D.6 that, $\hat{\overline{w}}_{priv}$ = $\hat{\overline{w}}$ + $b$, is $\epsilon_{p}$-differentially private (where $b$ is a random vector drawn according to $(47)$). Then we have by the post-processing property of differential privacy, that $\hat{\overline{\eta}}$ is $\epsilon_{p}$-differentially private. Finally, it follows by Lemma D.7 that the DP Plug-in algorithm that uses output perturbation as its privacy preserving protocol is $\epsilon_{p}$-differentially private (and so it is $\epsilon_{p}$-differentially private w.r.t the sensitive attribute as well). Algorithm 9 details our instatiation of the DP Plug-in algorithm that uses output perturbation as its privacy-protocol.

  \begin{algorithm}[H]
   \caption{\small{DP Plugin approach to FAL via Output Perturbation; EO setting}}
\begin{algorithmic}

   \STATE \small{{\bfseries Input:} Sample $S$ = $\{x_{i}, y_{i}, \overline{y}_{i}\}_{i=1}^{n}$ from distribution $\mathbbm{P}$; cost parameters $c, \overline{c}$; trade-off parameter $\lambda$}; regularization parameter $\Lambda$; privacy parameter $\epsilon_{p}$
   \vspace{2.5mm}
   \STATE {\bfseries Estimate:} \small{$\pi$ via \hspace{0.5 em}$\hat{\pi} = \frac{1}{n}\sum_{i=1}^{n}\mathbbm{I}\{y_{i} = 1\}$}
   \STATE {\bfseries Estimate:} \small{$\eta$:$\chi \rightarrow [0, 1]$ using appropriate CPE on $\{x_{i}, y_{i}\}_{i=1}^{n}$}
   \STATE {\bfseries Estimate:} \small{$\overline{\eta}_{EO}$: $(\chi, Y) \rightarrow [0, 1]$ corresponding to regularized ERM w.r.t. loss $\ell$ over appropriate model class on $S$}
   \STATE {\bfseries Privatise:} \small{$\hat{\overline{\eta}}_{EO}$ via  $\hat{\overline{w}}_{priv}$ =  $\hat{\overline{w}} + b$; where $b$ is a random vector drawn according to $(47)$, with $\Gamma = \frac{n\Lambda\epsilon_{p}}{2}$, yielding $\epsilon_{p}$-DP protected $\hat{\overline{\eta}}_{EO}$}
   \STATE {\bfseries Compute:} $\hat{s}^{priv}\left(x\right) =$  \begin{scriptsize}$\smash{\left\{ 1 - \frac{\lambda}{\hat{\pi}}(\hat{\overline{\eta}}_{EO}^{priv}(x, 1) - \overline{c}) \right\}\hat{\eta}(x) - c}$\end{scriptsize}
   \vspace{2.5mm}
   \STATE {\bfseries Return:} \small{$\hat{f}^{priv}\left(x\right) = H_{\alpha}\left(\hat{s}^{priv}\left(x\right)\right)$ for any $\alpha \in [0, 1]$}
\end{algorithmic}
\end{algorithm}
 
Let us now analyse the finite sample properties of Algorithm 9.  Before proceeding, we summarise our assumptions: 
\\\\
{\bf Assumption 10} {\hspace{0.1em} Domain space $\chi$ is compact and  $\forall x \in \chi$ and $y \in \mathcal{Y}$, $\|(x, y)\|_{2} \leq 2$
}
\\\\
{\bf Assumption 11} {\hspace{0.1em} The regression function, $\overline{\eta}$ and estimators, $\hat{\overline{\eta}}$ are completely parameterised via parameter vectors $\overline{w}^{*} \in \mathcal{\overline{W}}$ and $\hat{\overline{w}} \in  \mathcal{\overline{W}}$, through function $g: ( \mathcal{\overline{W}}, (\chi, \mathcal{Y})) \rightarrow [0, 1]. \hspace{0.3em}$ Here $\mathcal{\overline{W}} \subset \mathbb{R}^{d+1}.$ \hspace{0.3em} $\exists \phi: (\chi, \mathcal{Y}) \rightarrow \mathbb{R},$ such that, $\forall \hspace{0.2em} (\overline{w}, \overline{w}^{'}) \in \overline{\mathcal{W}}^{2}, \hspace{0.3em}$   $\forall x \in \chi$ and for any $y \in \mathcal{Y}$:
   
     \[
     \vert g(\overline{w};(x, y)) - g(\overline{w}{'}; (x, y)) \vert \hspace{0.3em} \leq \hspace{0.3em} \phi(x, y)\vert \overline{w} - \overline{w}^{'} \vert \hspace{0.35em}
     \]
     and, 
\[
     \forall x \in \chi, \forall y \in \mathcal{Y}, \phi(x, y) \hspace{0.3em} \leq \hspace{0.3em} F
     \]

     Assume WLOG, $F = 1$. Thus, $\forall x \in \chi$ and any $y \in \mathcal{Y}$: 
     \begin{equation}
     \vert g(\overline{w};(x,y)) - g(\overline{w}{'}; (x,y)) \vert \hspace{0.3em} \leq \hspace{0.3em} \vert \overline{w} - \overline{w^{'}} \vert  \hspace{0.35em} \forall  (\overline{w}, \overline{w}^{'}) \in \overline{\mathcal{W}}^{2}
     \end{equation}
}

{{\bf Assumption 12} {\hspace{0.1 em}} $\hat{\overline{\eta}}$ is the class probability estimator obtained by solving a regularized ERM over a loss $\ell(\cdot, \cdot)$ and regularizer $N(\cdot)$ which satisfy conditions laid out in Theorem D.6.
}

Assumptions 10 and 12 are required to ensure all conditions of Theorem D.6 are met. The regularity conditions brought up in Assumption 11 are reasonable. Indeed the condition $(49)$ on $g(\cdot; (\cdot, \cdot))$  holds  true (up to constants on the right hand side of the inequalities) for regression functions differentiable with respect to the parameter vector when the parameter space is compact. This is because regression functions (in case of a binary label setting) take values in a bounded range, i.e., $[0, 1]$, implying that for any $y \in \mathcal{Y}$ and any $x \in \chi$, $g(\cdot; (x,y))$ has bounded gradients across the range of the parameter vector and is thus Lipschitz continuous  $\implies \hspace{0.3em} \phi(x, y)$ exists and is bounded. Since the bounded gradient condition holds $\forall x \in \chi$ and any $y \in \mathcal{Y}$, constant $F$ exists. We set $F = 1$ without loss of generality. A case where Assumption 11 holds, is when the true regression functions and their estimators are logistic regression models in the same parameter spaces. As we will see in our analysis, formalising these conditions allows us to isolate out the component of regret induced due to noise addition for privacy preservation. 
\\\\
Suppose $b$ is a random vector of dimensionality $d+1$, drawn from $(47)$, with $\Gamma = \frac{n\Lambda\epsilon_{p}}{2}$. Then, we define the 'tail-decay complexity' for $b$ as follows: 
\\\\
{\bf Definition D.8} {\hspace{0.1em} \it The tail-decay complexity (for fixed $\Lambda, \epsilon_{p}$) for a random vector $b$ with density given by $v(b) = \frac{1}{\alpha_{p}}e^{-\Gamma \|b\|_{2} }$ is a mapping $m_{b}: (0,1)^{2} \rightarrow \mathbbm{N}$, where $m_{b}(\epsilon, \delta) $ is the minimal (integer) number of training samples required to ensure that, $\mathbb{P}(\vert b \vert > \epsilon) \leq \delta$}
\\\\
A more stringent privacy requirement would lead to a smaller choice of $\epsilon_{p}$, implying a larger average magnitude of noise to be added, leading to an increase (in the point-wise sense) in the tail-decay complexity associated with $b$.
\\\\
Let $\smash{\delta, \delta^{'}, \epsilon \in (0, \frac{1}{2})}$. Now, suppose the sample size, $n \geq max\{m_{\eta}((\epsilon,\frac{\delta^{'}}{2}), \frac{\delta}{8}), m_{\overline{\eta}}((\frac{\epsilon}{2}, \frac{\delta^{'}}{4}), \frac{\delta}{8}), m_{b}(\frac{\epsilon}{2}, \frac{\delta^{'}}{4})\}$. Upon applying Algorithm 9, we obtain parameter estimates, $\hat{\overline{w}}$ and $\hat{\overline{w}}_{priv} = \hat{\overline{w}} + b$, where $b$ is a random vector drawn according to $(47)$. Here, $\hat{\overline{\eta}}, \hat{\overline{\eta}}_{priv}$ will denote the class probability estimators corresponding to $\hat{\overline{w}}, \hat{\overline{w}}_{priv}$. By the condition on $n$, by definitions C.1 and C.2, and by the union bound we have that with probability $\geq (1-\frac{\delta}{4})$, class probability estimators so obtained, namely $\hat{\eta}$, $\hat{\overline{\eta}}$ are such that: 

\begin{equation}
    \mathbb{P}_{X}(\vert \hat{\eta}(x) - \eta(x) \vert \geq \epsilon) \leq \frac{\delta^{'}}{2}
\end{equation}
and, 
\begin{equation}
    \mathbb{P}_{X}(\vert \hat{\overline{\eta}}(x, 1) - \overline{\eta}(x, 1) \vert \geq \frac{\epsilon}{2}) \leq \frac{\delta^{'}}{4}
\end{equation}

The randomness pertaining to equations $(50)$ and $(51)$ is due to the random draw of the training sample, $S$. Now, by the condition on $n$ and Definition D.8, we have that: 

\begin{equation}
\mathbb{P}(\vert b \vert \geq \frac{\epsilon}{2}) \leq \frac{\delta^{'}}{4}
\end{equation}

The randomness pertaining to $(52)$ is due to the fact that $b$ is a random vector drawn according to the density in $(47)$, i.e., the source of randomness here is separate from that in $(50)$ and $(51)$. Now, with probability $(1-\frac{\delta}{4})$, we get that: 

\[
\mathbb{P}_{X}\left[\vert \overline{\eta}(x, 1) - \hat{\overline{\eta}}_{priv}(x, 1) \vert \geq \epsilon \right]
\]
\[
\overset{(a)}{\leq} \mathbb{P}_{X}\left[\vert \overline{\eta}(x, 1) - \hat{\overline{\eta}}(x, 1) \vert \geq \frac{\epsilon}{2} \right] + \mathbb{P}_{X}\left[\vert \hat{\overline{\eta}}(x, 1) - \hat{\overline{\eta}}_{priv}(x, 1) \vert \geq \frac{\epsilon}{2} \right]
\]
\[
\overset{(b)}{=} \mathbb{P}_{X}\left[\vert \overline{\eta}(x, 1) - \hat{\overline{\eta}}(x, 1) \vert \geq \frac{\epsilon}{2} \right] + \mathbb{P}_{X}\left[\vert g(\hat{\overline{w}}; (x, 1)) - g(\hat{\overline{w}}_{priv}; (x, 1)) \vert \geq \frac{\epsilon}{2} \right]
\]
\[
\overset{(c)}{\leq} \mathbb{P}_{X}\left[\vert \overline{\eta}(x, 1) - \hat{\overline{\eta}}(x, 1) \vert \geq \frac{\epsilon}{2} \right] + \mathbb{P}\left[\vert \hat{\overline{w}} - w^{*} \vert \geq \frac{\epsilon}{2} \right]
\]
\[
\overset{(d)}{=} \mathbb{P}_{X}\left[\vert \overline{\eta}(x, 1) - \hat{\overline{\eta}}(x, 1) \vert \geq \frac{\epsilon}{2} \right] + \mathbb{P}\left[\vert b \vert \geq \frac{\epsilon}{2} \right] 
\]
\[
\overset{(e)}{\leq} \frac{\delta^{'}}{2}
\]
where $(a)$ follows from the union bound, $(b)$ follows from the structure assumed on $\overline{\eta}$ and its estimators, $(c)$ follows from Assumption 11, $(d)$ follows from the definition of $\hat{\overline{w}}_{priv}$ and $(e)$ follows from $(51)$ and $(52)$. Therefore, we have that with probability $\geq (1-\frac{\delta}{4})$: 

\begin{equation}
    \mathbb{P}_{X}(\vert \hat{\eta}(x) - \eta(x) \vert \geq \epsilon) \leq \frac{\delta^{'}}{2}
\end{equation}
and, 
\begin{equation}
    \mathbb{P}_{X}(\vert \hat{\overline{\eta}}_{priv}(x, 1) - \overline{\eta}(x, 1) \vert \geq \epsilon) \leq \frac{\delta^{'}}{2}
\end{equation}

Our analysis so far would have worked out identically, had we set $n \geq max\{m_{\eta}((\epsilon,\frac{\delta^{'}}{2}), \frac{\delta}{8}), m_{\overline{\eta}}((\frac{\epsilon}{2}, \frac{\nu\delta^{'}}{2}), \frac{\delta}{8}), m_{b}(\frac{\epsilon}{2}, \frac{(1-\nu)\delta^{'}}{2})\}$ for any $\nu \in (0, 1)$. Ideally, we would set $\nu$ such that the sample complexity is optimal, i.e., $\nu^{*} = argmin_{\nu \in (0, 1)}\left\{max\{m_{\eta}((\epsilon,\frac{\delta^{'}}{2}), \frac{\delta}{8}), m_{\overline{\eta}}((\frac{\epsilon}{2}, \frac{\nu\delta^{'}}{2}), \frac{\delta}{8}), m_{b}(\frac{\epsilon}{2}, \frac{(1-\nu)\delta^{'}}{2})\}\right\} $. Setting $n$ according to this argument, provides a tighter characterisation. Now, following precisely the steps laid out in the sections C.2 and C.3, we get that: 
\\\\\\
{\bf Theorem D.11 \hspace{0.1em} \it  Let $\epsilon_{p} >0$. Let $\smash{\delta, \delta^{'}, \epsilon \in (0, \frac{1}{2})}$. $\smash{\nu^{*} = argmin_{\nu \in (0, 1)}\left\{max\{m_{\eta}((\epsilon,\frac{\delta^{'}}{2}), \frac{\delta}{8}), m_{\overline{\eta}}((\frac{\epsilon}{2}, \frac{\nu\delta^{'}}{2}), \frac{\delta}{8}), m_{b}(\frac{\epsilon}{2}, \frac{(1-\nu)\delta^{'}}{2})\}\right\}}$.\\\\ Consider in the $(\overline{\eta}(\cdot, 1), \eta)$-plane, the hyperbola $H(\lambda, \pi, c, \overline{c}) := \{(1 + \frac{\lambda\overline{c}}{\pi})\eta - \frac{\lambda}{\pi}\overline{\eta}(\cdot, 1)\eta - c = 0$\}. Pick any $t > Q = 4G \{max\{c(1-\pi), (1-c)\pi, \vert\lambda\vert\overline{c}(1-\beta), \vert\lambda\vert(1-\overline{c})\beta\}\}$, where $G = max\{\frac{B}{1-\pi}, \frac{B}{\pi \beta}, \frac{B}{\pi (1-\beta)}\}$, and $B = \mathbbm{P}_{X}(X_{M}) + \delta^{'}$. Here $X_{M} := \{x \in \chi:$ the square of length $2\epsilon$ centred at $(\overline{\eta}(x, 1), \eta(x))$ intersects the hyperbola $H(\lambda, \pi, c, \overline{c})$ in the $(\overline{\eta}(\cdot, 1), \eta)$-plane\}. Provided access to $n \geq max\{m_{\eta}((\epsilon,\frac{\delta^{'}}{2}), \frac{\delta}{8}), m_{\overline{\eta}}((\frac{\epsilon}{2}, \frac{\nu^{*}\delta^{'}}{2}), \frac{\delta}{8}), m_{b}(\frac{\epsilon}{2}, \frac{(1-\nu^{*})\delta^{'}}{2})\}$ training samples drawn i.i.d. from $\mathbbm{P}$, the plug-in algorithm yields an $\epsilon_{p}$-differentially private estimator $\hat{f}$, such that, with probability at least $(1-\delta) \hspace{0.3em}: \hspace{0.3em} regret_{\mathbbm{P}}^{\psi}(\hat{f}) \leq t$
}
\\\\\\
We note that the \textbf{Subset Argument} presented following Theorem C.3, is directly applicable over here as well. This implies that the regret in this setting, can be made to decrease arbitrarily, provided a sufficient increase in the number of training samples. The precise rate of decay depends on 1) the sample complexities associated with learning the regression functions and 2) the rate at which the probability measure ($\mathbbm{P}_{X}$) around the hyperbola $H(\lambda, \pi, c, \overline{c})$ decays (in the $(\overline{\eta}(\cdot, 1), \eta)$-plane) upon shrinking the region of consideration around it.

Comparing with Theorem C.3, it is easy to see that, the sample requirement (to obtain low regret with high probability via the plug-in algorithm) in the private-setting is at least as much as the sample-requirement in the non-private setting, which we can recall from Theorem C.3 was $max\{m_{\eta}((\epsilon, \frac{\delta^{'}}{2}), \frac{\delta}{8}), m_{\overline{\eta}}((\epsilon, \frac{\delta^{'}}{2}), \frac{\delta}{8})\}$. It is worth noting, that the cost of privacy in terms of sample efficiency is also hidden in $m_{b}(\frac{\epsilon}{2}, \frac{(1-\nu^{*})\delta^{'}}{2})$, since the tail-decay complexity of $b$ depends on privacy parameter $\epsilon_{p}$. A more stringent privacy requirement would lead to a smaller choice of $\epsilon_{p}$, implying a larger average magnitude of noise to be added, leading to an increase (in the point-wise sense) in the tail-decay complexity associated with $b$. Using precisely the template laid out in this sub-subsection, we can obtain similar results for privacy under the Approximate Demographic Parity criterion.  

\subsubsection{DP Plug-in via Output Perturbation - Experimental Results}
In this sub-subsection, we present empirical evaluations for the DP Plug-in approach, wherein we make use of the 'Output Perturbation' method of \cite{chaudhuri2011differentially} as the privacy protocol. In their paper, \cite{jagielski2019differentially} propose a differentially private version of \cite{hardt2016equality}'s post-processing approach that mitigates violations in equalized odds (i.e., violations that arise due to additive disparities in TPRs and FPRs). For brevity, we will refer to \cite{jagielski2019differentially}'s private implementation of \cite{hardt2016equality}'s post-processing approach for mitigating violations in equalized odds, as the 'DP Post-Proc' approach hereon. We compare DP Plug-in's performance with the DP Post-Proc method across 2 publicly available data sets, the Adult and German data sets \cite{dheeru2017uci}. Note: Since we are interested in equality of opportunity, we simply remove the constraint on false positive rates while running experiments for 'DP Post-Proc'. 

\textbf{\underline{Data description and pre-processing:}} The German data set comprises of 1000 instances and 20 attributes, 7 numerical and 13 categorical. Each entry in the data set corresponds to an individual and their financial/ banking information along with some demographic information. Each individual is classified as 'good' or 'bad' depending on their ability to pay back a loan, i.e., their credit risk. Thus the task associated with the data set is a binary classification task. The data set comprises of two features, that may be deemed sensitive, namely, gender and age. The feature age comes with a binary encoding, where individuals older than 25 years of age are encoded with 1. Gender also comes with a binary encoding. We carry out two sets of experiments on this data set. In the first set, we assume gender is the sensitive attribute, and we do not use age as a feature.  In the second set, we assume age is the sensitive attribute, and we do not use gender as a feature. The Adult data set comprises of 48842 instances and 14 attributes. The prediction task at hand here, is to classify individuals according to their income categories, i.e., to predict whether an individual earns more, or less than 50K USD annually. The data set comprises of two features, that may be deemed sensitive, namely, gender and race. Both sensitive attributes come with binary encodings. In the first set, we assume gender is the sensitive attribute, and we do not use race as a feature.  In the second set, we assume race is the sensitive attribute, and we do not use gender as a feature. Thus we have four experimental set ups in total. We pre-process both data sets, so that the $\ell$-2 norm of each individual's joint feature-label vector is bounded above by 1. We do this pre-processing to ensure Assumption 10 introduced in the previous sub-subsection is met. Recall that Assumption 10 is necessary to obtain the privacy guarantees of \cite{chaudhuri2011differentially} upon leveraging their Output Perturbation method for privacy preservation. For each of the four experimental set ups, we obtain 20 randomized train-validation-test splits of the data sets in a 70:20:10 proportion. We run our experiments across each of the 20 batches obtained per data set. 

\textbf{\underline{Models:}} None of the models make use of the sensitive attribute as a predictor. Indeed, we assume only train time access to the sensitive attribute, and further require that the modelling pipelines be $\epsilon_{p}$ differntially private with respect to the sensitive attribute. However, for the purpose of reporting model performance in terms of fairness, we access the sensitive attributes corresponding to the test set and do not consider this as a privacy violation (even if we did, the excess loss in privacy would be identical for both methods, so this does not matter from the stand point of drawing comparisons anyway). For the DP Post-Proc approach, we train a logistic regression model as the base model, which is then subjected to post-processing for mitigating discrimination and preserving privacy. For the DP Plug-in approach, we use a logistic regression model to obtain the class probability estimator $\hat{\eta}$ for the regression function $\eta$. We train an $\ell$-2 regularized logistic regression model to obtain the class probability estimator, $\hat{\overline{\eta}}$ for the regression function $\overline{\eta}$. The loss function associated with estimating $\overline{\eta}$ is the binary cross entropy loss function which is convex and differentiable. Further, we set the regularization to $\frac{\|w\|_{2}}{2}$, (where $w$ represent model weights) implying the regularization term is differentiable and $1$-strongly convex. The pre-processing of the data sets and the choice of loss and regularization functions ensures that the conditions laid out in Theorem D.7 hold, and so the corresponding privacy guarantee holds.  

\textbf{\underline{Evaluation Metrics}} We measure a model's fairness violation with respect to a data set, by computing the absolute difference in true positive rates between the two sensitive groups, over the test data, i.e., $\vert  \frac{\sum_{i}\mathbb{I}\{\hat{y}_{i} = C, y_{i} = C, \overline{y}_{i} = 1\}}{\sum_{i} \mathbb{I}\{y_{i} = C, \overline{y}_{i} = 1\}} - \frac{\sum_{j}\mathbb{I}\{\hat{y}_{j} = C, y_{j} = C, \overline{y}_{j} = -1\}}{\sum_{j} \mathbb{I}\{y_{j} = C, \overline{y}_{j} = -1\}}\vert$, where the summation is over the test data. Note: Due to pre-processing, we use $y \in \{-C, C\}$, where $C >0$ corresponds to a positively labeled instance. For utility, we measure a model's balanced accuracy with respect to the test data. The balanced accuracy is defined as the average of the sensitivity and specificity. Our experimental goal is to evaluate the fairness-accuracy trade-offs of both methods. In other words, we evaluate the least fairness violation achievable by a model, at a given level of balanced accuracy. At a given balanced accuracy, method A outperforms method B is the fairness violation associated with method A is lower than that associated with method B. Roughly, we will say method A outperforms method B, if method A yields lower fairness violations across the majority of the balanced accuracy range being considered. We will compare DP Post-Proc and DP Plug-in the manner described. 

\textbf{\underline{Parameters and reporting methodology:}} We set our privacy parameter, $\epsilon_{p} = 1.0$ for all experiments. For the DP Post-Proc method, we vary the decision threshold associated with the base model between $0.1$ to $0.975$ in step sizes of $0.025$; and we vary the constraint parameter, denoted $\gamma$ in \cite{jagielski2019differentially} between $0.0$ to $1.0$ in  step sizes of $0.001$. We thus traverse over the 2-D grid of (threshold, $\gamma$) pairs and ensure that, each iteration over the grid satisfies differentially privacy (w.r.t the sensitive attribute) with $\epsilon_{p} = 1.0$. For the DP Plug-in approach, we vary the trade-off parameter, $\lambda$ between $-10$ to $10$ in  step sizes of $0.5$, and we vary cost parameters $c, \overline{c}$ between $0.1$ to $0.9$ in step sizes of $0.1$. We thus traverse over the 3-D grid of $\lambda, c, \overline{c}$ in the specified ranges. One advantage inherent to the plug-in approach of \cite{menon2018cost} is that we can traverse through the hyper-parameter space without having to train our models again. DP Plug-in inherits this advantage. Using steps laid out in Algorithm 9, we ensure that the estimation of $\overline{\eta}$ is $\epsilon_{p} = 1.0$-differentially private with respect to the sensitive attribute. The post-processing property of differential privacy (Definition D.4) ensures no further privacy leakage occurs upon traversing the hyper-parameter space. Traversing the respective hyper parameter spaces for each model, we are able to evaluate model fairness at different levels of balanced accuracy. We discretise the balanced accuracy scores of the model into segments of $2.5 \%$, ranging from $50 \%$ balanced accuracy up to the maximum achievable balanced accuracy for the model, and for each segment, we compute the maximum achievable fairness value, or equivalently, the lowest fairness violation. We repeat this across all 20 randomized train-validation-test splits and obtain the average minimal fairness violation for every segment. We also obtain the standard deviation values for the minimal fairness violation corresponding to each segment. Note that the randomness here is not only due to the randomized train-validation-test splits, but also because of the noise injection being carried out for privacy preservation. We plot these balanced accuracy vs. fairness violation values for each of our four experimental set ups in \textit{Figure 5}. For all four plots, the solid blue curve represents model performance in the 'balanced accuracy-fairness violation plane' corresponding to our method, i.e., the DP Plug-in approach. Whereas the solid red curve, corresponds to model performance for the DP Post-Proc approach. The dotted curves represent the $\pm 0.2$ standard-deviations in fairness violations corresponding to each segment of balanced accuracy considered.

\textbf{\underline{Conclusions:}} For both settings corresponding to the Adult data set, i.e., when gender is assumed sensitive and when race is assumed sensitive, our approach outperforms the DP Post-Proc approach. This is evidenced in the top panel of \textit{Figure 5}. For almost the entire span of the balanced accuracy range considered, the DP Plug-in approach yields a smaller fairness violation compared to the DP Post-proc method. In the case of the German data set, when gender is assumed sensitive, our approach again outperforms the DP Post-Proc method uniformly across the 'balanced accuracy-fairness violation' plane. However, the DP Post-Proc method outperforms our approach when we assume age to be the sensitive attribute in the German data set. Comparatively speaking, the difference in the area under the (balanced accuracy - fairness violation) curves is significantly higher when DP Plug-in outperforms DP Post-Proc than vice-versa. Thus, even when outperformed by DP Post-Proc, DP Plug-in is relatively competitive. In conclusion, we found that the DP Plug-in method outperforms the DP Post-Proc method in three out of four experimental scenarios considered.

\begin{figure}[tbh]
\centering
\includegraphics[width = 16cm]{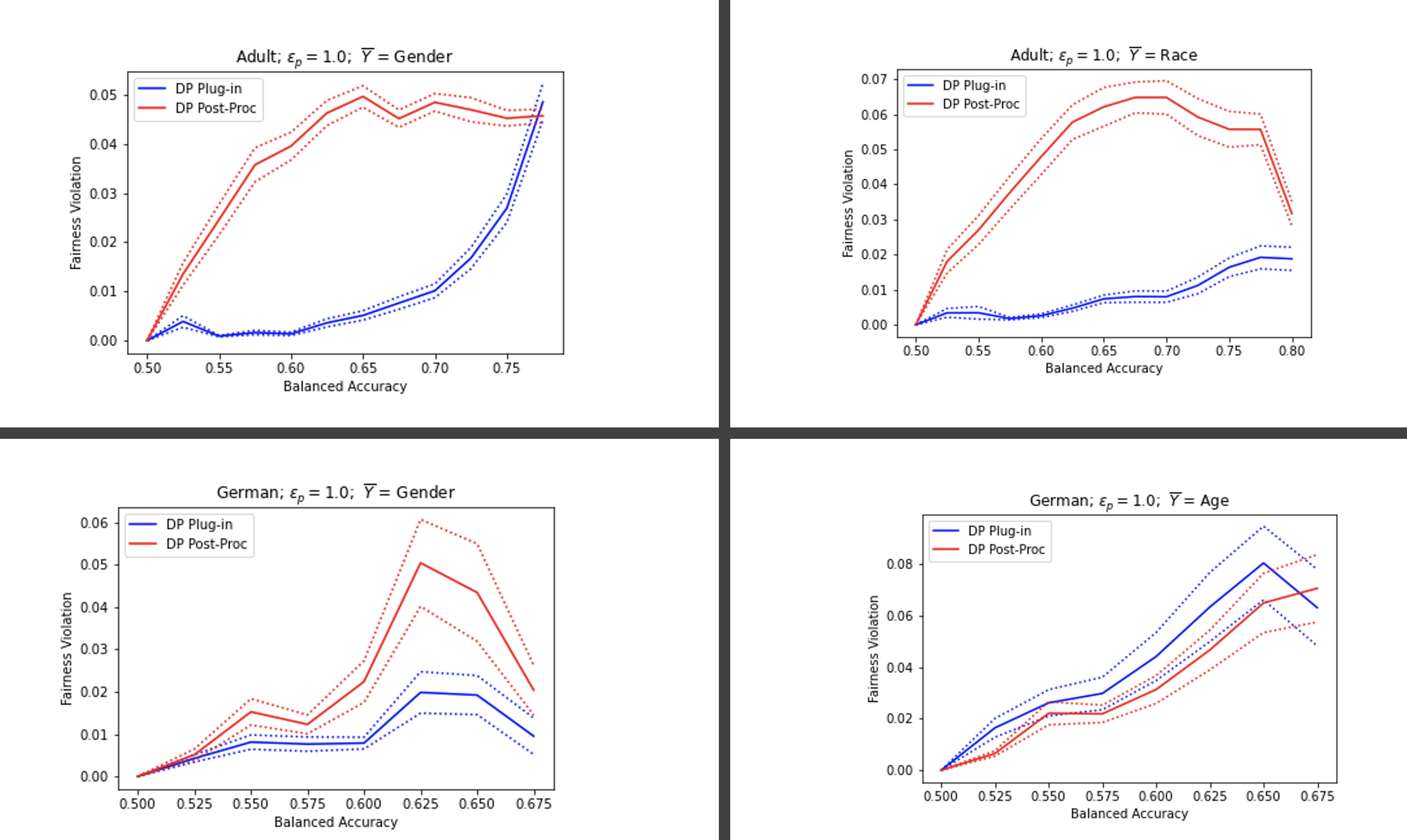}
\caption{\textit{For all four plots, the solid blue curve represents model performance in the 'balanced accuracy-fairness violation plane' corresponding to our method, i.e., the DP Plug-in approach. Whereas the solid red curve, corresponds to model performance for the DP Post-Proc approach. The dotted curves represent the $\pm 0.2$ standard-deviations in fairness violations corresponding to each segment of balanced accuracy considered. \textbf{Top Left}: Adult data set, privacy parameter $\epsilon_{p} = 1.0$, gender assumed sensitive. The DP Plug-in method obtains lower fairness violations compared to DP Post-Proc across the vast majority of the balanced accuracy range considered. \textbf{Top Right}: Adult data set, privacy parameter $\epsilon_{p} = 1.0$, race assumed sensitive. The DP Plug-in method obtains lower fairness violations compared to DP Post-Proc across the vast majority of the balanced accuracy range considered. \textbf{Bottom Left}: German data set, privacy parameter $\epsilon_{p} = 1.0$, gender assumed sensitive. The DP Plug-in method obtains lower fairness violations compared to DP Post-Proc across the vast majority of the balanced accuracy range considered. \textbf{Bottom Right}: German data set, privacy parameter $\epsilon_{p} = 1.0$, age assumed sensitive. DP Post-Proc obtains slightly lower levels of fairness violation across the vast majority of the balanced accuracy range considered.}}
\end{figure}

\end{document}